\documentclass[11pt,a4paper]{article}

\usepackage[margin=1in]{geometry}
\usepackage{times}
\usepackage{moreverb,url}
\usepackage{amsmath}
\usepackage{graphicx}
\usepackage{booktabs}
\usepackage{natbib}
\usepackage[colorlinks,bookmarksopen,bookmarksnumbered,
            citecolor=red,urlcolor=red]{hyperref}

\newcommand\BibTeX{{\rmfamily B\kern-.05em \textsc{i\kern-.025em b}\kern-.08em
T\kern-.1667em\lower.7ex\hbox{E}\kern-.125emX}}

\usepackage{xcolor}

\begin{document}

\title{From Doyle to AGM: A Survey and an Implementation Roadmap for Belief Change}

\author{%
  Yuri Almeida\thanks{NOVA Laboratory for Computer Science and Informatics (NOVA LINCS),
  Universidade da Madeira, Campus Universitário da Penteada,
  Funchal, 9020-105, Madeira, Portugal.
  Corresponding author: \texttt{yuri.almeida@staff.uma.pt}.}
  \and
  Arthur Casals\thanks{NOVA Laboratory for Computer Science and Informatics (NOVA LINCS),
  Universidade da Madeira, Campus Universitário da Penteada,
  Funchal, 9020-105, Madeira, Portugal.
  E-mail: \texttt{arthur.casals@nlincs.uma.pt}.}
}

\maketitle

\begin{center}
\textit{This is the authors' accepted manuscript of the article
published in The European Journal on Artificial Intelligence (SAGE).
The final published version is available at
\href{https://doi.org/10.1177/30504554261426170}{doi:10.1177/30504554261426170}.}
\end{center}

\begin{abstract}
This paper presents a targeted narrative review establishing the historical and theoretical foundations for computational belief change implementation. Seeded by Doyle and London's foundational 1980 taxonomy, we trace the evolution of belief revision from computational origins through the theoretical transformation of the AGM framework to contemporary approaches. Our analysis demonstrates how pre-AGM computational pragmatism relates to AGM theoretical constructs, revealing both continuities and transformations across this evolution. We analyze how each taxonomical category evolved in the post-AGM era, identifying the theoretical foundations and historical precedents that inform contemporary implementation challenges. This foundation enables subsequent research into robust computational blueprints that synthesize historical insights with formal guarantees, providing the baseline for systematic implementation analysis and engineering-focused belief change research.
\end{abstract}


\section{Introduction}\label{sec:introduction}

Belief revision (BR) is the process by which an intelligent agent modifies its beliefs when confronted with new information that contradicts its current knowledge state \cite{gardenfors1988knowledge}. This fundamental cognitive and computational process involves determining which existing beliefs to abandon, which new beliefs to accept, and how to maintain consistency and coherence in the resulting belief system. Unlike simple belief addition, which merely incorporates new information, belief revision requires sophisticated mechanisms to handle conflicts, prioritize information sources, and preserve the most reliable or important beliefs while discarding those that have become untenable.

To illustrate the problem faced by BR, we adapt an example from~\cite[pp. 1-2]{ferme2018belief}. Suppose that the following statements are all true:

\begin{itemize}
    \item John was born in Portugal ($\alpha$)
    \item Pedro was born in Massaguassú ($\beta$)
    \item John and Pedro are compatriots ($\gamma$)
\end{itemize}

From these statements, we infer that Massaguassú is in Portugal. Now, assume that we received two new pieces of information:

\begin{itemize}
    \item Massaguassú is not a country ($\delta$)
    \item Massaguassú is in Brazil ($\epsilon$)
\end{itemize}

Since the five statements are inconsistent, it follows that it is not viable to accept all five as true. There are several different approaches to dealing with this problem. Should we forget some of our previous beliefs? If so, which are the ones we choose to forget? And why? These are just a few of the questions that BR theories try to answer. 

The formal study of belief revision encompasses both normative theories that specify how rational agents should revise their beliefs and descriptive theories that model how belief revision actually occurs in natural and artificial systems. At its core, belief revision addresses three fundamental operations: \textit{expansion} (adding new beliefs without removing old ones), \textit{contraction} (removing beliefs to restore consistency), and \textit{revision} (adding new beliefs while removing conflicting old ones to maintain consistency)~\cite{alchourron1985logic}.

BR is crucial for Artificial Intelligence (AI), as real-world AI systems must operate in dynamic, uncertain environments where information is incomplete, potentially contradictory, and subject to change \cite{doyle1979truth}. Unlike static knowledge bases that assume perfect and unchanging information, intelligent agents must continuously update their understanding based on new observations, sensor data, user inputs, and changing environmental conditions.

It also plays a pivotal role in simulating human-like reasoning~\cite{harman1986change}. By modeling cognitive processes, AI systems can more effectively emulate human thought patterns, which is invaluable in developing intelligent assistants, educational tutors, and collaborative systems, to name a few examples. Furthermore, BR supports the integration of learning and reasoning in AI~\cite{mitchell1997machine}. As systems increasingly combine data-driven methods, such as machine learning, with symbolic reasoning, BR facilitates the reconciliation of learnt patterns with logical rules through approximate operators~\cite{chopra2001approximate} and anytime algorithms~\cite{wassermann1999resource} that balance computational efficiency with theoretical soundness. This integration fosters the development of robust hybrid AI systems capable of excelling in complex tasks~\cite{bengio2013representation,russell2010artificial}.

Finally, BR is indispensable for maintaining consistency in knowledge systems. Systems such as expert systems and knowledge graphs rely on consistent information to ensure accuracy and utility~\cite{davis1982knowledge}. Updating these systems without introducing logical errors is a fundamental challenge that BR effectively addresses. By enabling these diverse capabilities, BR functions as a key mechanism to advance the adaptability, reliability, and intelligence of AI systems~\cite{russell2010artificial}.

In summary, the importance of BR in AI manifests itself in multiple dimensions:

\textbf{Adaptive Reasoning:} AI systems must adapt their reasoning as new evidence emerges, requiring mechanisms to identify which beliefs are no longer supported and how to propagate changes throughout interconnected knowledge structures~\cite{doyle1979truth}.

\textbf{Consistency Maintenance:} As agents acquire new information, conflicts inevitably arise with existing beliefs. BR provides principled methods for resolving these conflicts while preserving the coherence of the overall knowledge set~\cite{alchourron1985logic}.

\textbf{Learning and Knowledge Acquisition:} Machine learning systems inherently involve BR as they update their models based on new training data, requiring sophisticated approaches to balance new evidence against established patterns~\cite{mitchell1997machine}.

\textbf{Multi-Agent Systems:} When multiple AI agents share information, BR becomes essential to incorporate the knowledge of other agents while maintaining individual epistemic autonomy~\cite{stone2000multiagent}.

\textbf{Human-AI Interaction:} AI systems that interact with humans must revise their beliefs based on user feedback, corrections, and changing preferences, necessitating robust BR mechanisms that can handle subjective and contextual information~\cite{amershi2014power}.

\textbf{Ontology Repair and Knowledge Base Maintenance:} Modern AI systems increasingly rely on formal ontologies to structure domain knowledge, yet these ontologies must evolve as understanding deepens or inconsistencies emerge~\cite{flouris2005handling,flouris2006bridging,konstantinidis2007rdf}. Recent work has established formal connections between AGM belief revision and ontology repair, demonstrating that contraction operations can be adapted to description logic ontologies through optimal repair strategies that minimize information loss while restoring consistency~\cite{flouris2006inconsistencies,flouris2006bridging,flouris2013formal,KR2024-9,souza2024bridging}. This extension of classical belief revision to expressive logical frameworks provides principled methods for maintaining large-scale knowledge bases in semantic web technologies, knowledge graphs, and enterprise systems, making BR essential for the reliability of contemporary knowledge-centric AI applications.

\subsection{Problem Statement and Motivation}\label{sec:motivation}

The field of belief revision faces a fundamental challenge in bridging the gap between theoretical foundations and practical implementation~\cite{ferme2018belief,FGR25}. While the AGM framework provides rigorous normative principles for rational belief change, these operations are computationally intractable in the worst case (co-NP-complete)~\cite{nebel1998hard}, making exact implementation infeasible for large-scale applications and often requiring approximations that may compromise theoretical guarantees~\cite{williams1997anytime,chopra2001approximate, delgrande2013approach}. In contrast, pre-AGM computational approaches demonstrated practical feasibility but lacked systematic theoretical foundations that could ensure rational behavior or provide principled evaluation criteria.
This theory-practice gap manifests itself in several critical ways. \textbf{Computational Complexity:} AGM revision operations are generally co-NP-complete, making exact implementation infeasible for large-scale applications, yet the relationship between approximation strategies and theoretical soundness remains poorly understood~\cite{liberatore1996complexity,liberatore2023mixed,nebel1998hard}. \textbf{Preference Specification:} AGM theory requires preference structures such as epistemic entrenchment orderings or selection functions, but provides limited guidance on how to acquire these preferences from users, data, or domain expertise~\cite{williams1994transmutations}. \textbf{Implementation Strategies:} Whereas AGM provides clear normative constraints, it offers little guidance on data structures, algorithms, or architectural decisions needed for practical systems~\cite{delgrande2013approach,williams1996towards}.

The motivation for this research stems from the recognition that pre-AGM computational approaches, developed before theoretical foundations were established, may contain valuable implementation insights that have been overlooked by the post-AGM focus on theoretical development. These early systems were designed with explicit attention to computational tractability, incremental processing, and practical deployment constraints, concerns that remain central to contemporary AI applications but are often secondary in theoretical research.

Additionally, the proliferation of AI systems in dynamic, uncertain environments, from autonomous vehicles to conversational agents, has renewed interest in practical belief revision implementations that can operate under real-time constraints with limited computational resources. Understanding how pre-AGM approaches addressed these challenges, and how their insights can be integrated with AGM's theoretical framework, is crucial for developing the next generation of intelligent systems.

To better understand progress in this field, it is essential to revisit the BR algorithms that preceded the influential AGM model, which has become the dominant approach. 
By comparing the state of pre-AGM algorithms with post-AGM developments, it is possible to understand how these models can be evaluated in the context of the AGM theory, and ultimately how they can be adapted for the implementation of belief revision in intelligent agents.
We argue that the theory of belief change for belief bases has now matured to a stage where the next step, implementing BR in real-world agents, can be undertaken. This is due to advances within the research field itself, and also to the increased computational power of small devices over the past decades. With this understanding, we contend that the algorithms for BR proposed by the AI community in the 1970s and 1980s likely contain valuable insights into the implementation of BR mechanisms.

In this paper, we analyze how the models proposed in the pre-AGM era can be evaluated in light of the AGM theory. We begin by analyzing pre-AGM BR algorithms, building on the classic article, 'A Selected Descriptor-Indexed Bibliography to the Literature on Belief Revision'~\cite{doyle1980selected}. Doyle and London’s work provides a detailed overview of early research and pre-AGM approaches, offering a valuable framework for our analysis. We use this bibliography as a foundation for revisiting key contributions and outlining developments in the field of BR. Subsequently, we analyze how these contributions can be assessed in light of the AGM theory and ultimately how they can be adapted when implementing belief revision mechanisms in intelligent agents. 

Rather than a systematic literature review, this paper presents a targeted narrative review, grounded in Doyle and London's work, which serves as a basis for tracing the evolution of belief revision research from its computational origins through theoretical formalization to contemporary implementations. This targeted narrative review establishes the historical and theoretical foundation for a systematic research program addressing computational belief revision implementation. By comprehensively analyzing the evolution from pre-AGM computational approaches through AGM theory to contemporary systems, we provide the baseline necessary for systematic implementation analysis and engineering-focused research. Subsequent work will build on this foundation to develop computational blueprints and implementation analysis frameworks that synthesize historical insights with formal guarantees.

\subsection{Paper organization}\label{sec:organization}

This paper is organized as follows:

\textbf{Section~\nameref{sec:related_work}} presents a comprehensive review of related work, positioning our survey approach within the broader landscape of belief revision surveys and reviews. We compare our methodology and scope with existing literature reviews and identify the unique contributions of our historical-theoretical analysis.

\textbf{Section~\nameref{sec:background}} establishes the philosophical and theoretical foundations of belief revision research, tracing its emergence from epistemological inquiries and Newell's knowledge level framework through the transition from computational practice to a theoretical foundation that sets the stage for our historical analysis.

\textbf{Section~\nameref{sec:AGM}} provides essential background on the AGM framework, including its theoretical foundations, postulates, construction methods, and criticisms. This section establishes the theoretical lens through which we analyze pre-AGM computational approaches and post-AGM developments.

\textbf{Section~\nameref{sec:preAGM}} examines pre-AGM computational approaches to belief revision, with particular emphasis on systems identified in Doyle and London's taxonomy. We analyze Truth Maintenance Systems, Assumption-Based Truth Maintenance Systems, probabilistic belief networks, and other foundational computational approaches, documenting their algorithmic innovations and implementation strategies.

\textbf{Section~\nameref{sec:methodology}} presents our methodological framework for systematic correspondence analysis between pre-AGM computational approaches and AGM theoretical constructs. We detail our taxonomical mapping strategy, postulate correspondence analysis, and systematic approach to tracing theoretical evolution across Doyle's categories, providing the analytical foundation for our historical-theoretical integration.

\textbf{Section~\nameref{sec:taxonomy_evolution}} provides a comprehensive analysis of how each category in Doyle's taxonomy evolved in the post-AGM era. For each category, we examine key theoretical developments, identify seminal papers that drove transformation, and analyze how computational insights were formalized into theoretical frameworks.

\textbf{Section~\nameref{sec:synthesis}} synthesizes our findings, examining the enduring value of Doyle's classification, the lasting impact of the AGM revolution, and opportunities for integrating pre-AGM and post-AGM approaches. We discuss the contributions of key figures, including Doyle and McCarthy, and analyze patterns of theoretical development across the field.

\textbf{Section~\nameref{sec:discussion}} concludes with a summary of the findings, research contributions, and the identification of systematic research directions enabled by our foundational analysis. We establish the theoretical and historical baseline necessary for subsequent computational blueprint development while identifying strategic directions for engineering-focused implementation research.

\section{Related Work}\label{sec:related_work}

The field of belief revision has been the subject of numerous surveys, reviews, and comparative analyses over the past four decades. These works provide different perspectives on the field's development, theoretical foundations, and practical applications. This section positions our survey within this broader landscape of literature reviews and identifies how our work extends and complements existing analyses.

\subsection{Foundational Surveys and Reviews}

In 'A Selected Descriptor-Indexed Bibliography to the Literature on Belief Revision', which is the earliest comprehensive survey of belief revision research~\cite{doyle1980selected}, the authors established the taxonomical framework that forms the foundation of our analysis. This work was primarily descriptive, cataloging existing approaches without systematic theoretical evaluation. In "Knowledge in Flux"~\cite{gardenfors1988knowledge} provided the first comprehensive theoretical treatment of belief revision, focusing on the AGM framework and its philosophical foundations rather than computational implementation.

In "How to Give It Up"~\cite{makinson1985give} offered an early systematic examination of formal aspects of belief change, providing mathematical foundations that complemented Gärdenfors' more philosophical approach. This work focused primarily on contraction operations and their formal properties, establishing important theoretical results that informed subsequent AGM development.

The first comprehensive handbook treatments of belief revision appeared in the mid-1990s, providing systematic coverage that synthesized the first decade of post-AGM research. The handbook chapter by~\cite{gardenfordsrott95} offered the first authoritative survey of AGM theory and its extensions, covering the fundamental postulates, construction methods, and representation theorems while addressing early criticisms and limitations. This work established the canonical presentation of AGM theory that would influence subsequent textbook treatments and became the standard reference for researchers entering the field. A decade later, \cite{vanharmelen2008} handbook chapter provided updated coverage that reflected the maturation of the field, including comprehensive treatment of iterated revision, belief base approaches, and computational complexity results that had emerged since the mid-1990s. These handbook chapters serve complementary roles to book-length treatments, providing authoritative surveys that capture the field's consensus understanding at specific historical moments while remaining more accessible than comprehensive monographs.
\subsection{Post-AGM Theoretical Surveys}

Following the establishment of the AGM framework, several comprehensive surveys examined its extensions and applications. In "Change, Choice and Inference"~\cite{rott2001change} provides extensive coverage of connections between belief revision and non-monotonic reasoning, examining how AGM principles relate to default logic, conditional logic, and other approaches to uncertain reasoning. This work complements our analysis by focusing on theoretical connections rather than historical development.

In "A Textbook of Belief Dynamics"~\cite{hansson1999textbook} offers comprehensive coverage of belief revision theory with emphasis on formal constructions and their properties. In their more recent work,~\cite{ferme2018belief} provide updated coverage including belief base approaches, kernel methods, and contemporary extensions. These works focus primarily on theoretical development rather than the historical evolution that is central to our analysis. The authors further extend their coverage in a later publication, but focusing only on the AGM development~\cite{FGR25}.

In his survey of belief revision implementations~\cite{williams1997applications} examined practical aspects of belief revision systems, focusing on computational approaches and their relationship to theoretical foundations. 

\subsection{Specialized Domain Surveys}

Several surveys have examined belief revision within specific application domains. A survey of ontology change by~\cite{flouris2008ontology} examines belief revision principles as applied to ontology evolution and maintenance, providing detailed coverage of description logic extensions and practical implementation strategies. This work demonstrates how AGM principles have been adapted for specific logical frameworks.

In their work on belief merging~\cite{konieczny2002merging} survey approaches to combining information from multiple sources, extending AGM principles to multi-agent settings. 

A complexity survey by \cite{liberatore1996complexity} provides comprehensive coverage of computational complexity results for belief revision operations, examining both theoretical bounds and practical approximation strategies. This work complements our analysis by providing detailed technical results. 

\subsection{Computational and Implementation Surveys}

Several works have surveyed computational approaches to belief revision implementation. In her work on resource-bounded belief revision~\cite{wassermann1999resource} examines how computational limitations affect belief revision implementation, providing important insights into practical constraints that complement our historical analysis of pre-AGM computational approaches.

In their survey of approximation methods~\cite{schaerf1995tractable} examine various approaches to approximate belief revision, including compilation methods, anytime algorithms, and heuristic approaches. 

In their work on approximate belief revision,~\cite{chopra2001approximate} provide a systematic account of approximation as an explicit response to computational constraints in belief change. Rather than treating approximation as an \textit{ad hoc} compromise, they characterize families of approximation operators and analyze their trade-offs with respect to AGM postulates and rationality criteria. This work remains primarily situated within the AGM framework, and the implementation of the proposed system is marked as future work.

\subsection{Philosophical and Cognitive Perspectives}

Several surveys have examined belief revision from philosophical and cognitive perspectives. In his work on belief revision and reasoning~\cite{harman1986change} examines connections between belief revision and human reasoning, providing insights into descriptive aspects of belief change that complement AGM's normative framework. 

A survey of defeasible reasoning by~\cite{pollock1987defeasible} examines connections between belief revision and defeasible reasoning, addressing some of the non-monotonic aspects that Doyle's taxonomy identified as adjacent to belief revision proper.

These philosophical surveys provide important context for understanding belief revision as both a computational and cognitive phenomenon. 

\subsection{Contemporary Multi-Agent and Social Perspectives}

Recent surveys have addressed belief revision in multi-agent and social contexts. In their work on dynamic epistemic logic~\cite{baltag2008qualitative} survey formal approaches to knowledge and belief change in multi-agent settings, providing sophisticated logical frameworks that extend beyond the individual focus of classical AGM theory. 

\cite{dragoni1997belief} examines early approaches to distributed belief revision with a survey of belief revision in multi-agent systems, providing important coverage of extensions beyond single-agent settings. This work complements our analysis by addressing multi-agent aspects, but does not provide the systematic historical perspective central to our approach.

\subsection{Positioning of Our Work}

Our review differs from existing surveys in several important ways:

\textbf{Historical Scope:} Unlike surveys that focus primarily on post-AGM theoretical developments~\cite{rott2001change,hansson1999textbook,ferme2018belief}, our work systematically examines pre-AGM computational approaches and traces their evolution through the AGM transformation to contemporary implementations.

\textbf{Taxonomical Framework:} We use Doyle and London's original taxonomy as an organizational framework, providing systematic coverage of all major aspects of belief revision research while maintaining historical continuity.

\textbf{Theory-Practice Integration:} Our analysis explicitly examines the relationship between theoretical development and computational implementation, identifying how practical concerns have influenced theoretical development and how theoretical insights have informed practical approaches. This distinguishes our work from purely theoretical surveys~\cite{gardenfors1988knowledge,makinson1985give} and purely computational surveys~\cite{williams1997applications}.

\textbf{Methodological Approach:} We develop a correspondence analysis that maps pre-AGM computational techniques to AGM theoretical constructs, providing a methodological framework that could be applied to other areas of AI research.

While our work builds upon and complements existing surveys, it provides a unique perspective that combines historical analysis, theoretical evaluation, and practical implementation insights within a unified framework. This approach fills an important gap in the literature by analyzing how the field has evolved from its computational origins to its current theoretical sophistication.

\section{Background and Foundations}\label{sec:background}

The philosophical roots of BR theory stem from epistemological inquiries into knowledge acquisition and belief change. The formal study of BR emerged from the intersection of AI, philosophy of science, and cognitive psychology in the 1960s and 1970s. In his influential concept of the \textit{knowledge level} \cite{newell1982knowledge} provided a crucial theoretical foundation by distinguishing among different levels of analysis in cognitive systems: the biological level, the symbol level, and the knowledge level.

At the knowledge level, Newell argued, intelligent behavior can be understood in terms of the knowledge an agent possesses and the goals it pursues, independent of the specific symbolic representations or computational mechanisms used to implement that knowledge. This perspective was revolutionary because it suggested that BR could be studied as a rational process governed by the principles of coherence, consistency, and goal-directed behavior, rather than merely as a collection of ad hoc computational procedures.

Newell's framework influenced early AI researchers to develop principled approaches to knowledge representation and reasoning that could support rational BR. This led to the development of truth maintenance systems \cite{doyle1979truth}, assumption-based reasoning frameworks~\cite{dekleer1986assumption}, and other computational architectures designed to track the justifications for beliefs and manage belief dependencies systematically~\cite{doyle1979truth}.

The knowledge level perspective also highlighted the importance of distinguishing between the \textit{competence} level (what an ideal rational agent should do) and the \textit{performance} level (what computationally bounded agents actually do). This distinction became central to BR research, leading to both normative theories that specify ideal BR behavior and computational theories that address the practical constraints of implementing BR in real systems~\cite{newell1982knowledge}.

Early contributions from philosophers such as~\cite{james1896will} and~\cite{peirce1877fixation} emphasized the dynamic character of belief systems and the necessity of rational methods for updating them. These philosophical investigations established several key principles:
\begin{itemize}
    \item \textbf{Minimal Change Principle}: When incorporating new information, one should preserve as much of the existing belief system as possible while ensuring consistency;
    \item \textbf{Coherence}: The resulting belief system should preserve internal consistency and logical soundness;
    \item \textbf{Epistemic Entrenchment}: Some beliefs are more fundamental or important than others, influencing which beliefs should be retained during revision.
\end{itemize}

Peirce and James are central to pragmatist ideas about belief and coherence, but the formal, systematic foundations of minimal change and entrenchment in belief revision were developed later by decision‑theoretic and AGM‑style authors (Levi; Alchourrón, Gärdenfors, Makinson) and by AI/logic researchers. Historical precursors such as~\cite{zimmerman2022pragmatism} and~\cite{hollingsworth2022towards} also shaped the pragmatists’ views.

From the computational perspective, BR explores how intelligent systems (or agents) can update their beliefs in response to new information. This process is essential because, in real-world contexts, AI systems often need to handle dynamic, incomplete, or contradictory information. With a BR process in place, intelligent systems can maintain a consistent information base in the presence of new input, thereby supporting more accurate decision-making~\cite{nebel1998hard,wassermann1999resource}. 

The formal study of BR can be traced to early work in AI and philosophical logic during the 1960s and 1970s. However, the field was revolutionized by the seminal work of~\cite{alchourron1985logic}, which provided the first comprehensive formal framework for belief change operations.
 
Before the AGM framework, researchers developed various algorithmic approaches to BR, often developed for specific applications. In particular, the problem of how to revise a database or knowledge base appeared in the AI community during the 1970s. The survey article by~\cite{doyle1980selected} provided a selected bibliography of about 250 papers related to this area. In its introduction, it states that “Belief Revision concentrates on the issue of revising systems of beliefs to reflect perceived changes in the environment or acquisition of new information."  

During the 1980s, several authors~\cite{alex1985language, dalal1988investigations, fagin1983semantics, katsuno1991propositional, katsuno1991difference, martins1988model, satoh1988nonmonotonic, winslett1988reasoning} proposed different models for performing belief change. However, these models were developed at the syntactic level, i.e., with a commitment to the representation language. Therefore, as~\cite{nebel1989knowledge} pointed out, 'the question remains how belief revision can be described on an abstract level independent of how beliefs are represented and manipulated inside a machine'.  

It remains unclear how to describe BR at the knowledge level, as~\cite{newell1982knowledge} introduced in his seminal article `The knowledge level. In that work, he argued that `...there exists a distinct computer system level, lying immediately above the symbol level, which is characterized by knowledge as the medium and the principle of rationality as the law of behavior'. Newell further suggested that intelligent behavior could be understood in terms of an agent’s beliefs, goals, and rational action selection, abstracting away from implementation details.

These ideas later inspired the formal development of the Belief–Desire–Intention (BDI) paradigm, in which agent behavior is explicitly modeled in terms of beliefs, desires, and intentions~\cite{rao1995agents, cohen1990intention}. While Newell did not propose a BDI architecture himself, his knowledge-level analysis provided a conceptual foundation that strongly influenced subsequent BDI models and agent-oriented reasoning frameworks. That paper exerted an enormous influence on the AI community, particularly on researchers in Knowledge Representation and Reasoning, including Brachman, Levesque, Moore, Halpern, Moses, Vardi, Fagin, Ullman, Shapiro, Borgida, and others.

\subsection{The AGM Revolution}\label{sec:intro_AGM}

The AGM framework, developed by~\cite{alchourron1985logic, gardenfors1988knowledge, makinson1985give}, represents the most influential normative theory of belief revision. Named after its creators' initials, the AGM framework provides a set of postulates that specify how rational agents should revise their beliefs when confronted with new information. This work fundamentally transformed the field by:

\begin{enumerate}
    \item Providing a unified theoretical framework for three types of belief change: expansion, contraction, and revision;
    \item Establishing rationality postulates that any reasonable belief change operation should satisfy;
    \item Introducing representation theorems that connected abstract postulates with concrete construction methods;
    \item Creating formal relationships between different types of belief change operations.
\end{enumerate}

In the AGM literature, representation theorems provide equivalent characterizations of the abstract postulates in terms of concrete constructions. For example, contraction operators satisfying the AGM postulates can be represented as \emph{partial meet contractions}, obtained by selecting preferred remainder sets via a selection function~\cite{alchourronmakinson1982,alchourron1985logic}. Other representation theorems give equivalent formulations in terms of epistemic entrenchment orderings~\cite{gardenfors1988revisions}, kernel contraction~\cite{alchourron1986maps,hansson1994kernel}, and Grove’s system of spheres for revision~\cite{grove1988modelings}.

The AGM model characterizes specific change operations that represent the modifications an agent can perform on its knowledge. Each of these operations is defined by a set of rationality postulates that determine its behavior independently of implementation details. These postulates allow us to conceive of the change operations as ``black boxes,'' that is, mechanisms whose behavior is specified separately from how they are implemented.~\cite{alchourron1985safe}

Additionally, by the end of the 1980s, four equivalent methods for constructing AGM functions were introduced, along with semantics for such operations: partial meet functions~\cite{alchourronmakinson1982, alchourron1985logic}, epistemic entrenchment~\cite{gardenfors1988revisions}, safe/kernel contraction~\cite{alchourron1986maps, hansson1994kernel}, and sphere-systems semantics~\cite{grove1988modelings}. Together, these developments elevated the AGM to the status of a standard model. The influence of Newell’s paper and the AGM model led to the abandonment of syntactic approaches.  

However, the AGM framework is defined for belief sets, i.e., sets of sentences closed under logical consequence (usually infinite sets). Hence, resource-bounded agents cannot employ the AGM model without modifications~\cite{wassermann1999resource}.

During the 1990s and 2000s, the AGM model was extended to belief bases, i.e., sets of sentences not closed under logical consequence~\cite{hansson1991belief, hansson1994kernel, falappa2006logic, ferme2008axiomatic}. Several works advocate the use of belief bases instead of belief sets to represent states of knowledge~\cite{ferme1992actualizaci, fuhrmann1991theory, hansson1991thesis, hansson1992defense, hansson1994taking, nebel1989knowledge, rott1998just, wassermann1999resource}.  

While the AGM model played a pivotal role in formalizing BR, its practical application in real-world systems has faced challenges, primarily due to computational complexity and resource limitations. Despite these obstacles, AGM-based concepts have been adapted and extended to address the requirements of dynamic decision-making systems~\cite{nebel1998hard,wassermann1999resource}.

\subsection{Transition from Practice to Theory}

Our survey serves as a bridge between the practice-oriented pre-AGM era documented in Section~\nameref{sec:preAGM} and the theory-oriented AGM framework that we examine in the subsequent Section~\nameref{sec:AGM}. Doyle's taxonomy provides the organizational structure for understanding how diverse computational approaches to belief revision were unified under the AGM theoretical umbrella, while our correspondence analysis reveals both the continuities and innovations that characterize this theoretical transformation.

The pre-AGM period was rich in concrete computational mechanisms that addressed specific implementation challenges. Consider three exemplary cases that illustrate our methodological approach:

\textbf{Operator descriptions} in planning systems like STRIPS~\cite{fikes1971strips} fall under Doyle's \emph{Representations} category and pair naturally with the operator application \emph{Procedures} to implement model change. These systems directly addressed the frame problem by making explicit what changes and, implicitly, what remains constant during action execution. From an AGM perspective, these mechanisms anticipate the Success and Inclusion postulates, though they lack the systematic theoretical foundation that AGM provides.

\textbf{Context layers} and \textbf{change-triggered procedures}, including pattern-directed invocation mechanisms, capture consequential effects and localize revisions, key innovations under Doyle's \emph{Representations} category. These approaches operationalize a weak notion of causality through precondition/effect relationships and support sophisticated backtracking strategies under the \emph{Procedures} category. AGM theory abstracts away from these implementation details, while preserving their essential insights about minimal change and preference-guided revision through selection functions and epistemic entrenchment orderings~\cite{gardenfors1988knowledge}.

\textbf{Logical data dependencies} and \textbf{Truth Maintenance Systems} explicitly record support relationships between beliefs and compute revised belief status through dependency-directed algorithms, representing perhaps the most direct pre-AGM bridge between computational practice and later AGM theory~\cite{doyle1979truth}. These systems map to dependency structures under \emph{Representations} and dependency-based revision under \emph{Procedures}, anticipating AGM's kernel contraction methods~\cite{hansson1994kernel} and belief base approaches that would emerge in the 1990s.

Non-monotonic techniques, including defaults, circumscription, and default logic, inhabit a separate taxonomical family precisely because they solve a different problem than AGM revision: how to license tentative commitments in the absence of complete proof, rather than how to minimally revise a belief corpus when definitive information arrives~\cite{reiter1980logic, mccarthy1980circumscription}. Doyle's taxonomical segmentation acknowledges the frequent co-occurrence of these techniques in implemented systems without conflating their distinct purposes, a distinction that AGM theory would later formalize through its focus on categorical belief change rather than uncertain inference.

\section{The AGM Model}\label{sec:AGM}
The AGM framework emerged in the mid-1980s as a response to the need for principled normative foundations for belief revision \cite{alchourron1985logic}. Although pre-AGM approaches had developed sophisticated computational methods, they lacked systematic theoretical foundations that could guide the development of rational belief revision systems and provide criteria for evaluating different approaches.

The framework was developed through a series of influential papers by Carlos Alchourrón, Peter Gärdenfors, and David Makinson, culminating in the comprehensive treatment in "Knowledge in Flux" by~\cite{gardenfors1988knowledge}. The AGM approach was revolutionary in applying rigorous logical and set-theoretic methods to belief revision, establishing it as a legitimate area of formal investigation.

The underlying formal logical framework used in the AGM approach has the following characteristics:

\begin{itemize}
    \item A propositional language $\mathcal{L}$, closed under the standard logical connectives
    \item A consequence operator $Cn$ that is:
    \begin{itemize}
        \item Monotonic: If $A \subseteq B$ then $Cn(A) \subseteq Cn(B)$
        \item Idempotent: $Cn(Cn(A)) = Cn(A)$
        \item Compact: If $\alpha \in Cn(A)$, then $\alpha \in Cn(A')$ for some finite $A' \subseteq A$
    \end{itemize}
    \item A set of formulas $K$ representing the belief set, where $K = Cn(K)$
\end{itemize}

The framework provides a systematic approach to belief change based on three fundamental operations:

\begin{itemize}
    \item \textbf{Expansion}: This operation involves adding a new belief to the belief set without concern for consistency. It simply incorporates the new information, regardless of potential contradictions, and may therefore result in inconsistent belief sets. This operation, denoted as $K + \alpha$, represents the addition of a new belief $\alpha$ to the belief set $K$, along with all the logical consequences. Formally:
    \[
    K + \alpha = Cn(K \cup \{\alpha\})
    \]
    \item \textbf{Contraction}: This operation removes a belief from the belief set while maintaining logical consistency. It is used when the agent needs to "unlearn" certain beliefs, often to make room for future updates. Formally, it is denoted as $K - \alpha$ and entails removing a belief $\alpha$ such that it is no longer a logical consequence of the remaining beliefs. This may imply removing other beliefs from the belief set. The operation must satisfy the principle of minimal change while maintaining logical closure.
    \item \textbf{Revision}: The most critical operation in the AGM model, revision consists of adding a new belief while preserving the consistency of the belief set. If the new belief contradicts existing beliefs, the model prescribes ways to modify the current belief set to accommodate the new information while preventing inconsistency. This operation, denoted as $K * \alpha$, involves incorporating a new belief $\alpha$ while maintaining consistency. The Levi Identity, proposed by~\cite{levi1978subjunctives}, provides a \textbf{fundamental relationship} between revision and contraction operations, describing revision as a two-step process: first, contract the belief set to remove any sentences that would conflict with the new belief; then expand the contracted set to include the new belief. This two-step approach ensures that the belief set remains consistent after revision, aligning with the AGM postulates. The relationship between revision and contraction is formalized as follows:
    \[
    K * \alpha = (K - \neg\alpha) + \alpha
    \]
\end{itemize}

It assumes that beliefs are represented as logically closed sets (belief sets) and that the underlying logic satisfies standard properties, including supraclassical monotonicity (\cite{strasser2024}), deduction, and compactness.

A belief set $K$ is a set of sentences in the language $\mathcal{L}$ that is closed under logical consequence:

\[
K = Cn(K) = \{\alpha \in \mathcal{L} : K \vdash \alpha\}
\]

This closure property embodies the idea of logical omniscience: agents are assumed to be aware of all logical consequences of their beliefs~\cite{gardenfors1988knowledge}.

\subsection{Postulates of the AGM Model}\label{sec:AGM_Postulates}
The AGM model is based on a set of rationality postulates that govern how belief sets behave when revised or contracted. These postulates ensure that belief revision is carried out logically and coherently. The basic AGM postulates for revision (K*1 – K*6) are:

\begin{enumerate}
    \item \textbf{Closure:} The revised belief set must be closed under logical consequence: if the revised set of beliefs logically implies a proposition, this proposition must also be included in the set. \[ K * \alpha = Cn(K * \alpha) \]
    \item \textbf{Success:} The new belief must be included in the revised belief set unless it leads to a contradiction. \[ \alpha \in K * \alpha \]
    \item \textbf{Inclusion:} The revised belief set should preserve as much of the original belief set as possible, with changes kept minimal and occurring only when necessary to preserve consistency. \[ K * \alpha \subseteq K + \alpha \]
    \item \textbf{Consistency:} The revised belief set must not contain contradictions. \[ K * \alpha \ \text{is consistent if} \ \alpha \ \text{is consistent} \]
    \item \textbf{Vacuity:} If the new belief is already part of the belief set, the revision must leave the set unchanged. \[ \neg\alpha \notin K \ \Rightarrow \ K + \alpha \subseteq K * \alpha \]
    \item \textbf{Extensionality:} If two pieces of information are logically equivalent, revising the belief set with either must result in the same outcome. \[ \alpha \equiv \beta \ \Rightarrow \ K * \alpha = K * \beta \]
\end{enumerate}

The supplementary postulates (K*7 – K*8) address composite revisions:

\begin{enumerate}
\setcounter{enumi}{6}
\item \textbf{Superexpansion}: Revision by conjunction is a subset of revision followed by expansion. \[ K * (\alpha \land \beta) \subseteq (K * \alpha) + \beta \]
\item \textbf{Subexpansion}: If a conjunction is consistent with beliefs, revision followed by expansion is equivalent to revision by conjunction. \[ \neg\beta \notin K * \alpha \ \Rightarrow \ (K * \alpha) + \beta \subseteq K * (\alpha \land \beta) \]
\end{enumerate}

For contraction, the model has its own set of postulates, designed to ensure that beliefs are removed logically while minimizing information loss. The AGM postulates for contraction (K-1 – K-8) are:

\begin{enumerate}
    \item \textbf{Closure}: The contracted belief set must remain closed under logical consequence.
    \[ K - \alpha = Cn(K - \alpha) \]
    
    \item \textbf{Inclusion}: The contracted belief set must not contain any new beliefs; it must be a subset of the original set.
    \[ K - \alpha \subseteq K \]
    
    \item \textbf{Vacuity}: If the belief to be removed is not currently part of the belief set, the contraction leaves the set unchanged.
    \[ \text{If } \alpha \notin K, \text{ then } K - \alpha = K \]
    
    \item \textbf{Success}: If the belief is not a logical truth (tautology), it must be successfully removed from the belief set.
    \[ \text{If } \not\vdash \alpha, \text{ then } \alpha \notin Cn(K - \alpha) \]
    
    \item \textbf{Recovery}: The original belief set must be recoverable by expanding the contracted set with the removed belief (ensuring minimal information loss).
    \[ K \subseteq (K - \alpha) + \alpha \]
    
    \item \textbf{Extensionality}: If two pieces of information are logically equivalent, contracting the belief set by either must result in the same outcome.
    \[ \alpha \equiv \beta \Rightarrow K - \alpha = K - \beta \]
    
    \item \textbf{Conjunctive Overlap}: Beliefs that remain after contracting by $\alpha$ and after contracting by $\beta$ individually must also remain when contracting by their conjunction.
    \[ K - \alpha \cap K - \beta \subseteq K - (\alpha \land \beta) \]
    
    \item \textbf{Conjunctive Inclusion}: If contracting by a conjunction succeeds in removing a specific conjunct, the result should not retain more beliefs than contracting by that conjunct alone.
    \[ \alpha \notin K - (\alpha \land \beta) \Rightarrow K - (\alpha \land \beta) \subseteq K - \alpha \]
\end{enumerate}

The AGM framework operates with two distinct but related concepts for representing beliefs: belief sets and belief bases. These representational choices have deep implications for both the theoretical development and practical implementation of belief revision systems.

Knowledge sets (or belief sets) in the AGM context represent idealized and logically closed sets of beliefs that embody the theoretical foundations of rational belief revision~\cite{gardenfors1988knowledge, alchourron1985logic}. These structures are characterized by deductive closure ($Cn$), meaning that if a knowledge base contains certain beliefs, it must also contain all logical consequences of those beliefs. This property ensures theoretical completeness and provides strong guarantees about consistency when revision operators follow the AGM postulates. However, the requirement for logical closure implies that knowledge sets are infinite in size for any non-trivial set of initial beliefs, as they must contain all possible logical derivations.

The practical limitations of logically closed knowledge sets led to the development of knowledge bases (or belief bases) as a more computationally tractable alternative~\cite{hansson1999textbook, nebel1998hard}. Belief bases are finite, non-closed sets of beliefs that provide explicit representation of only the basic or fundamental beliefs held by an agent. This approach offers several advantages for practical implementation: finite size enables computational tractability, the distinction between explicit and derived beliefs allows for more efficient reasoning procedures, and the overall structure provides more realistic modeling of how actual belief systems operate in resource-bounded agents~\cite{dalal1988investigations,ferme1992actualizaci,fuhrmann1988relevant,fuhrmann1991,hansson1989new,hansson1991belief,hansson1992defense,hansson1994taking,nebel1989knowledge,rott1998just,wassermann1999resource}.

The choice between knowledge sets and knowledge bases reflects a fundamental tension in belief revision research between theoretical elegance and computational feasibility. While knowledge sets provide the mathematical precision necessary for formal analysis and the development of rationality postulates, belief bases offer the practical advantages required for real-world implementation. Later approaches often attempt to bridge this gap by developing algorithms that operate on belief bases while approximating the rational behavior specified by AGM theory for knowledge sets~\cite{liberatore1996complexity, wassermann1999resource}.

\subsection{Constructive models}\label{sec:AGM_ConstrModels}
Constructive models in AGM theory serve as concrete mechanisms to implement belief change operations consistent with AGM postulates~\cite{alchourron1985logic}. While the AGM postulates provide normative constraints on BR, they do not specify how belief changes are to be performed. Constructive models bridge this gap by offering explicit procedures for belief change operations~\cite{gardenfors1988knowledge}.

The significance of constructive models can be highlighted in several respects:
\begin{enumerate}
    \item They provide explicit procedures for implementing belief change operations~\cite{gardenfors1988knowledge};
    \item They demonstrate that the AGM postulates are consistently satisfiable~\cite{alchourron1985logic};
    \item They offer insights into the relationship between different types of belief change~\cite{levi1991fixation};
    \item They serve as blueprints for practical implementations~\cite{nebel1998hard}.
\end{enumerate}

The development of constructive models begins with the concept of remainder sets~\cite{alchourronmakinson1981,alchourronmakinson1982}. For a belief set $K$ and a sentence $\alpha$, the remainder set $K \bot \alpha$ is defined as the collection of maximal subsets of $K$ that do not imply $\alpha$. Formally:
\[
K' \in K \bot \alpha \ \text{iff:}
\]
\begin{itemize}
    \item $K' \subseteq K$
    \item $\alpha \notin Cn(K')$
    \item For any $K''$ such that $K' \subset K'' \subseteq K$, $\alpha \in Cn(K'')$
\end{itemize}

\vspace{1em}

\subsubsection{Maxichoice Contraction}

The maximachoice contraction selects exactly one maximal subset of $K$ that does not imply $\phi$:

$$
    K -_{\text{max}} \phi = \gamma(K \perp \phi)
$$

where $K \perp \phi$ is the set of all maximal subsets of $K$ that do not imply $\phi$, and $\gamma$ is a selection function that chooses an element from this set.

\subsubsection{Partial Meet Contraction}

Partial Meet contraction takes the intersection of some maximal subsets:

$$
    K -_{\text{pm}} \phi = \bigcap \gamma(K \perp \phi)
$$

where $\gamma$ selects a non-empty subset of $K \perp \phi$.

\subsubsection{Full Meet Contraction}

Full meet contraction takes the intersection of all maximal subsets:

$$
    K -_{\text{fm}} \phi = \bigcap (K \perp \phi)\    
$$

\subsubsection{Selection Functions and Preferences}

The choice of the selection function $\gamma$ encodes preferences for different ways of contracting the belief set. These preferences can be represented using various structures:

\textbf{Epistemic Entrenchment:} A pre-order on sentences representing their relative importance or reliability \cite{gardenfors1988knowledge}.

\textbf{Grove Systems:} Collections of propositions ordered by plausibility \cite{grove1988modelings}.

\textbf{Ordinal Conditional Functions:} Numerical measures of disbelief that can guide BR \cite{spohn1988ordinal}.

\subsection{Revision}\label{sec:AGM_Revision}
The core of the AGM model lies in the way the revision function is constructed. Revision in AGM theory addresses the fundamental problem of incorporating new information into a belief set while preserving consistency~\cite{gardenfors1988knowledge}. Unlike expansion, which might lead to inconsistencies, revision ensures that the resulting belief set remains consistent whenever possible. The operation must balance two competing principles:
\begin{enumerate}
    \item Minimal change: Preserve as much of the original belief set as possible;
    \item Consistency maintenance: Ensure that the resulting belief set is consistent if the epistemic input is consistent.
\end{enumerate}

As introduced before, the Levi Identity formalizes revision as a two-step process in the context of revision operations. This identity demonstrates that revision can be reduced to a sequence of:

\begin{enumerate}
    \item Contracting by the negation of the new belief; and
    \item Expanding with the new belief.
\end{enumerate}

The Harper Identity provides the converse relationship~\cite{harper1976rational}:
$$
K - \alpha = K \cap (K * \neg\alpha)
$$
This identity shows that contraction can be defined in terms of revision, establishing a deep connection between these operations.

Partial meet revision extends the partial meet contraction framework to revision operations~\cite{alchourron1985logic}. The process involves:
\begin{enumerate}
    \item Computing $K \bot \neg\alpha$ (the remainder set for $\neg\alpha$);
    \item Applying a selection function $\gamma$;
    \item Taking the intersection of selected remainders;
    \item Adding $\alpha$ and its logical consequences.
\end{enumerate}
Formally:
$$
K * \alpha = \left(\bigcap \gamma(K \bot \neg\alpha)\right) + \alpha
$$

With his system of spheres, \cite{grove1988modelings} provides an alternative semantic model for revision based on possible worlds. The framework consists of:
\begin{itemize}
    \item A set of possible worlds;
    \item A system of nested spheres centered on $[K]$ (the worlds where $K$ is true);
    \item A mechanism for selecting the closest $\alpha$-worlds.
\end{itemize}
The revision operation selects the most plausible worlds where the new belief holds:
$$
[K * \alpha] = \min_{\preceq}([\alpha])
$$

There are also specific types of revision that can be used with the AGM model. Selective revision~\cite{ferme1999selective} allows partial acceptance of new information:
\begin{itemize}
    \item Transformation: New information undergoes a transformation before incorporation;
    \item Screening: Only certain aspects of new information are accepted;
    \item Modification: New information is modified during the revision process.
\end{itemize}

Multiple revision~\cite{zhang2001infinitary} handles the simultaneous incorporation of multiple pieces of information:
\begin{enumerate}
    \item Package revision: Simultaneous revision with multiple beliefs;
    \item Iterative revision: Sequential application of revision operations;
    \item Prioritized revision: Revision with ordered sets of beliefs.
\end{enumerate}

The classical AGM framework assumes that new information always takes priority over existing beliefs, as embodied in the Success postulate, which requires that newly received information must be incorporated into the revised belief set. However, this assumption has been challenged by developments in non-prioritized revision, which recognizes that not all new information deserves equal treatment or automatic acceptance~\cite{hansson1997semi}.

Non-prioritized revision fundamentally challenges the \textit{Success} postulate by introducing a more nuanced approach to information integration. Rather than mandating that new information must always be incorporated, this framework allows partial acceptance or complete rejection of new information based on its evaluated credibility and reliability. The system may determine that existing beliefs, particularly those with strong evidential support or high confidence levels, should be preferred over incoming information that appears dubious or contradicts well-established knowledge. This approach reflects a more realistic model of how rational agents actually process information in uncertain environments where the quality and trustworthiness of sources can vary significantly.

Parallel to these developments in information prioritization, researchers have also focused on adapting belief revision to work directly with belief bases rather than the idealized belief sets of classical AGM theory~\cite{nebel1998hard}. This shift involves fundamental changes in the way revision operations are conceptualized and implemented. When working with finite, non-closed sets of beliefs, revision procedures must explicitly distinguish between beliefs that are explicitly represented in the base and those that are merely derivable through logical inference. This distinction becomes crucial because different minimal change criteria may apply to explicit versus derived beliefs, leading to revision outcomes that can differ significantly from those predicted by classical AGM theory.

The combination of non-prioritized revision with belief bases approaches represents a significant evolution in belief revision theory, moving toward more realistic and implementable frameworks. These developments acknowledge the practical constraints faced by real agents: limited computational resources that make belief set closure infeasible, varying information quality that makes blanket acceptance policies inappropriate, and the need for revision procedures that can operate efficiently on finite representations while still maintaining principled approaches to belief change~\cite{wassermann1999resource, liberatore1996complexity}.

\subsection{Importance and Criticisms}\label{sec:AGM_ImpLimi}

The AGM model was the first widely accepted formalization of how rational BR should be handled. It provided a clear, logical framework that could be applied in multiple domains, including AI, philosophy, and logic~\cite{alchourron1985logic}.

One of the central ideas of the AGM model is the principle of minimal change. When revising beliefs, an agent should alter as little as possible to accommodate new information. This principle is crucial for maintaining the coherence of the agent's beliefs over time and for preventing unnecessary disruptions in reasoning~\cite{gardenfors1988knowledge}.

The AGM model applies to numerous fields within AI, such as knowledge representation, non-monotonic reasoning, ontologies, databases, and multi-agent systems. It is particularly useful for dynamic systems that must constantly update their knowledge as they interact with their environment~\cite{allen1983}.

By providing a set of logical postulates, the AGM model ensures that BR is rational and consistent, making it a critical tool for enabling AI systems to reason effectively in the face of new or conflicting information~\cite{alchourron1985logic}.

The AGM model has inspired numerous extensions and refinements over the years, addressing limitations such as resource constraints in real-time systems or the need for more flexible revision methods. These works have further cemented their role in the ongoing development of BR theory~\cite{hansson1999survey}.

It has also had an impact on practical implementations, such as:
\begin{itemize}
    \item \textbf{Systematic evaluation}: Providing the ability to formally assess BR systems~\cite{hansson1996test};
    \item \textbf{Design guidance}: Offering clear criteria for developing new algorithms~\cite{gardenfors1988knowledge};
    \item \textbf{Unified framework}: Providing a common language for discussing and comparing different approaches~\cite{makinson2002ways}.
\end{itemize}

Despite its widespread acceptance as the predominant framework for belief change, the AGM framework has been subject to considerable critique. This section offers an overview of such critiques, which can be viewed as limitations of the model.

Theoretically, logical omniscience and deductive closure create a mismatch with bounded agents. AGM treats derived and non-derived beliefs as indistinguishable once in the belief set, offering no internal mechanism to mark provenance or status \cite{rott1998just, levi1978subjunctives, levi1991fixation}. Deductive closure renders belief sets infinite for any nontrivial base; representing or computing with such sets is infeasible for real agents. \cite{hansson1993theory, hansson2013outcome} argues for finiteness of bases and outcomes, but standard AGM operations on finite inputs still yield infinite closures \cite{hansson1997semi}, forcing approximations that weaken guarantees~\cite{chopra2001approximate}. Maintaining consistency while maximizing information retention is also computationally demanding in general.

The \textit{Recovery} postulate claims that contracting by $\varphi$ and then re-adding $\varphi$ restores the original belief set \cite{gardenfors1982rules}. Numerous counterexamples undermine this principle \cite{hansson1991belief, hansson1996knowledge, hansson1996hidden, hansson1999recovery, niederee1991multiple}. Consequently, contraction schemes without \textit{Recovery}, called \textit{withdrawals}, were proposed~\cite{makinson1997screened, nebel1998hard, ferme1998logic, ferme1998semi, levi1997contraction, levi2004mild, meyer2002systematic, rott1999severe}. No widely accepted weaker replacement for \textit{Recovery} has yet emerged \cite{ferme2018belief}.

AGM revision and contraction assume full epistemic deference to new inputs \cite{cross1992conditionals}. Even in preference-ordered settings such as\cite{katsuno1991propositional}, fresh information overrides entrenched beliefs, but immediately loses any privileged status once incorporated. This rigidity motivated operators that relax or violate \textit{Success}, grouped as non-prioritized revision \cite{makinson1997screened, ferme1999selective, hansson1992defenseb, hansson1997semi, olsson1997coherence, hansson1997what, hansson2001credibility, ferme2003credibility, booth2012credibility, hansson1991belief, hansson1997semi, fuhrmann1996essay, hansson2002local, konieczny2008improvement} and non-prioritized contraction \cite{ferme2001shielded, ferme2003credibility}. \textit{Remainder} sets likewise treat all survivors as epistemically equivalent to the agent’s aims; \cite{levi1991fixation, levi1997contraction} argues instead for preserving beliefs with the highest instrumental or epistemic value relative to goals.

Iterated change, i.e., consecutive revisions or contractions, is underspecified in the AGM model: the original postulates give no constraints on how revision or contraction should depend on history (memory), enabling pathological sequences and instability \cite{boutilier1996iterated}. Fermé and Hansson proposed an organization of iterated review according to their memory:

\begin{itemize}
    \item \textit{Without memory:} In this scenario, each belief set undergoes revision according to a predefined method that remains uninfluenced by the manner in which the belief set was initially acquired. \textit{Full meet revision} is a classic example of this category. An analysis of these operators was conducted by\cite{areces2001iterable};
    \item \textit{Full memory:} opposed to the previous case, a full history of changes is maintained. Some operators of this class are \cite{brewka1991belief,lehmann1995belief,konieczny2000framework,falappa2013stratified};
    \item \textit{Partial memory:} These operators store information on how it arrived at the belief set, which is used for future operations, but the stored information is not enough to do a rollback. Most of the proposed operators are of this category. According to~\cite{rott2009shifting}, it is possible to further organize this category:

    \textbf{Conservative revision:} This operation conservatively minimizes changes to the pre-order to accept the input. When revising by $p$, the maximal $p$-worlds are repositioned to the bottom, leaving the rest of the pre-order unchanged. \cite{boutilier1993revision, boutilier1996iterated, rott2003coherence};

    \textbf{Moderate revision:} Revising by p rearranges the pre-order by placing $p$-worlds at the bottom and $\neg p$-worlds at the top, maintaining their relative orders \cite{makinson2009propositional, makinson2011conditional};

    \textbf{Radical revision:} similar to moderate revision, but it makes the new belief permanent. In radical revision, based on a pre-order, the $p$-worlds maintain their order while the $\neg p$-worlds are excluded, becoming inaccessible \cite{segerberg1998irrevocable, ferme2000irrevocable}.
\end{itemize}

Epistemic entrenchment orderings face modeling issues: transitivity and totality can misrepresent context-dependent importance and may fail to capture realistic, non-linear hierarchies \cite{rott1992preferential}.

AGM is single-agent by design. Multi-agent settings require aggregation, conflict resolution, and source-sensitive trust handling that exceed the original scope \cite{dragoni1997belief}. Some studies in this area include the work by~\cite{konieczny2011logic, konieczny2002merging} on Merging , \cite{toulmin2003uses},\cite{alej2009belief, falappa2002belief}, \cite{castelfranchi1997representatio, paglieri2006belief, paglieri2004argumentation, paglieri2006toulmin} on Argumentation, and \cite{booth2006admissible, booth2010double}, \cite{zhang2010logic}, and \cite{arlocosta2007knowing} on Game Theory.

The binary acceptance stance is also too coarse for many applications; probabilistic beliefs, partial acceptance, and dynamic uncertainty demand graded representations and updates~\cite{spohn1988ordinal,jeffrey1983logic, wassermann1999resource}.

In knowledge-base maintenance, AGM-style global consistency checks are costly \cite{eiter1992complexity}. Local updates can produce hidden global conflicts, and preserving semantic dependencies during change is non-trivial~\cite{hansson2002local}. Real-time systems add further constraints: strict latency bounds, resource limits, and continuous, possibly conflicting streams strain classical operators~\cite{wassermann1999resource,alechina2006resource}.

With the AGM theoretical framework established, we can now examine the computational approaches that preceded and informed its development. These pre-AGM systems, while lacking systematic theoretical foundations, embodied algorithmic insights and implementation strategies that would later find formal expression in AGM theory and continue to influence contemporary belief revision implementations.

\section{Pre-AGM Belief Revision}\label{sec:preAGM}

Before the AGM model, particularly in the late 1970s and early 1980s\footnote{While our analysis is anchored in Doyle and London's 1980 bibliography, we include several systems published shortly thereafter (ATMS 1986, Pearl's networks 1986) because: (1) they represent direct extensions of pre-AGM approaches documented in the bibliography (ATMS extends TMS; Pearl's work builds on earlier probabilistic approaches catalogued by Doyle); (2) they were developed independently of AGM theory and embody pre-AGM computational pragmatism; (3) they significantly influenced subsequent belief revision implementations.}, several computational approaches to BR emerged from artificial intelligence research, especially in the context of reasoning systems and knowledge representation. These algorithms were developed as part of efforts to create automated reasoning systems capable of handling uncertain and dynamic information. As a result, they focused primarily on practical implementation rather than theoretical foundations~\cite{doyle1979truth, dekleer1986assumption, mccarthy1980circumscription}.

Central to understanding this pre-AGM landscape is the bibliography by~\cite{doyle1980selected}, which provided the first systematic organization of the emerging field. The authors' survey took on a task that, at the time, had not yet been formalized by a consensus logic of belief change: to make sense of a rapidly growing \emph{practice} of belief maintenance and model update across AI subfields by carving the space into an interpretable taxonomy~\cite{doyle1980model}. Their method was deliberately empirical and cross–disciplinary. Rather than propose a monolithic theory, they curated an indexed bibliography and attached to each item a bundle of \emph{descriptors} that reflected what the artifact \emph{does} (e.g. maintains dependencies), \emph{how} it is \emph{represented} (e.g. operator schemas; context layers), \emph{what kind} of \emph{inference} it licenses (e.g. default rules), and \emph{where} it is \emph{used} (e.g. planning, explanation). The result is a pragmatic and multi-axis classification: eight top-level families with fine-grained subcategories that span implementation and theory.

Doyle and London’s indexing process proceeds in three steps:
\begin{enumerate}
  \item Identify recurring \textbf{tasks and pressures} that force belief change (e.g., model update after action; reconciling conflicting hypotheses; absorbing new evidence);
  \item Identify \textbf{representational devices} that systems rely on to make updates feasible (e.g., operator add/delete lists, context layers, dependency graphs);
  \item Identify \textbf{inference regimes} and \textbf{meta–theoretical commitments} involved (e.g., default/non–monotonic rules versus purely monotonic reasoning; probabilistic/fuzzy calculi; modal, temporal, and situation logics).
\end{enumerate}

Then each bibliographic entry is \emph{multi–labeled} along these axes. Crucially, the categories are \emph{not mutually exclusive}: a single system might introduce a novel representation (context layers), rely on a specific procedure (dependency-directed backtracking) and presuppose an inference style (defaults), while being deployed in a particular application (explanation).

At the top level, the taxonomy comprises:
\begin{enumerate}
    \item \textbf{Global Issues:} The frame problem, choosing between competing belief revisions, change of belief or mind, and revising beliefs for new information;
    \item \textbf{Representations for Revising Beliefs:} Manual updates, operator descriptions, context-layered databases, change-triggered procedures, logical data-dependencies, causal dependencies, and dependency-like representations;
    \item \textbf{Procedures for Revising Beliefs:} Chronological and non-chronological backtracking, operator application, and dependency-based revision;
    \item \textbf{Non-monotonic Inference Techniques:} Default reasoning, non-monotonic justifications, negation in databases, minimal model circumscription, and partial matching;
    \item \textbf{Inexact Inferential Techniques:} Probabilistic inference, decision theory, conclusion theory, multi-valued logics, and fuzzy logics;
    \item \textbf{Theoretical Issues:} Logics of rational belief, temporal logics, situational calculus, evolution theory, proof theory, non-monotonic logics, meta theory, theory of incomplete databases, modal logics, and reasoning about uncertainty;
    \item \textbf{Applications:} Explanation, modeling and execution, replanning from failures, transfer of expertise, learning, debugging computer programs, evolutionary computer systems, causal and perturbation analysis, and control of reasoning;
    \item \textbf{Related Issues:} Human belief and memory, epistemology, inductive reference, hypothetical reasoning and counterfactual conditionals, and scientific theory change;
\end{enumerate}

The prehistory of BR, before an agreed-upon logic of change, already identifies core tasks: \emph{(i)} \textbf{update} world models after actions (the frame problem); \emph{(ii)} \textbf{reconcile} inconsistent hypotheses; \emph{(iii)} \textbf{absorb} new information with minimal disruption. Doyle's categories exactly mirror these pressures. For example, ``Global Issues'' enumerates the update/choice/assimilation triad; ``Representations'' and ``Procedures'' identify the data structures and algorithms that made these tasks tractable at scale. In other words, the taxonomy is a codification of engineering practice that responds to the motivational cases introduced historically. This classification system not only captured the state of the field circa 1980 but also provided a framework that would prove durable across subsequent decades of research.

\begin{figure}[htbp]
    \centering
    \includegraphics[width=\linewidth]{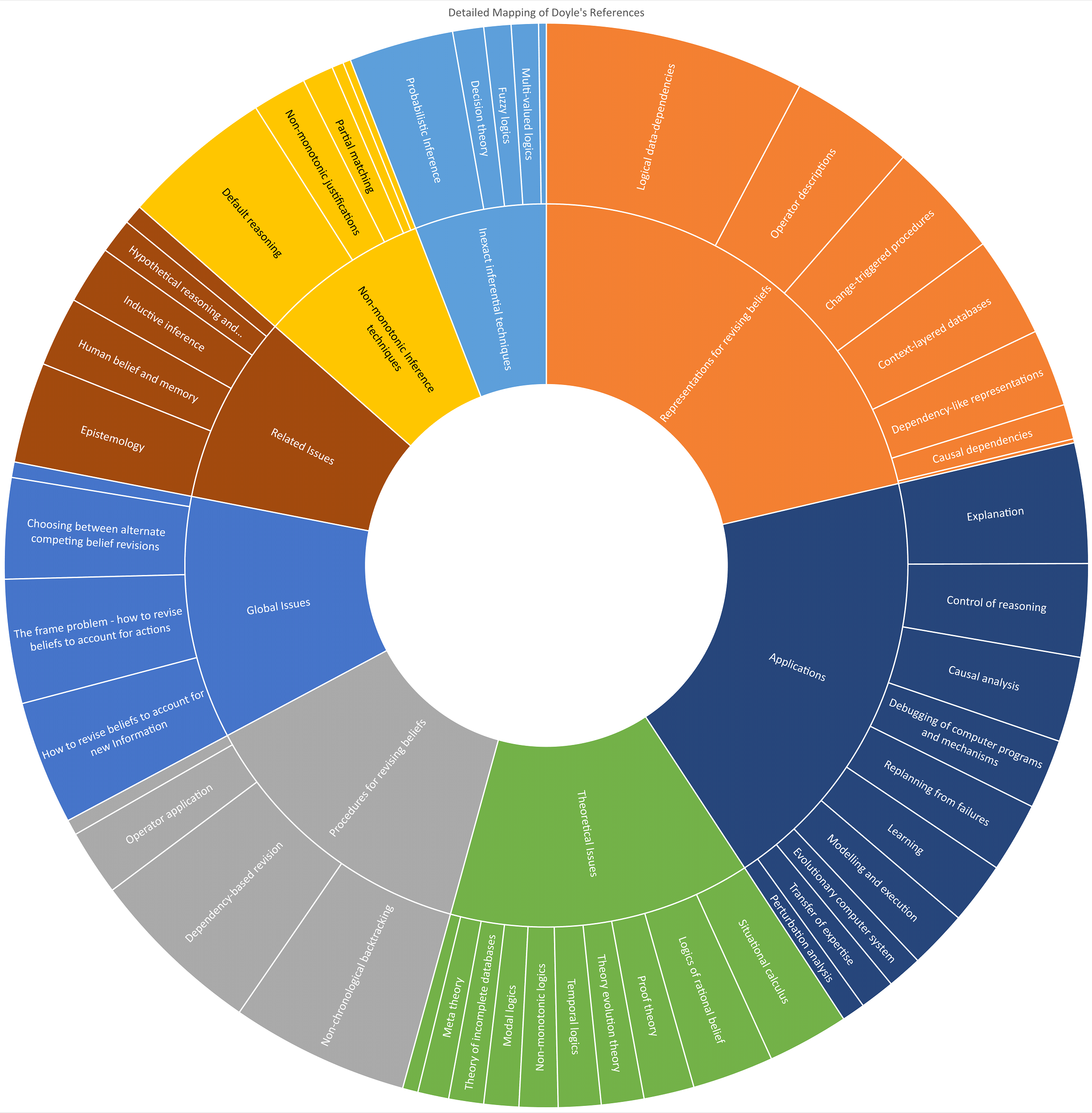}
    \caption{Detailed mapping of Doyle's references.} 
    \label{fig:doyle-wheel-small}
\end{figure}

\begin{figure*}[p]
    \centering
    \includegraphics[width=\textwidth]{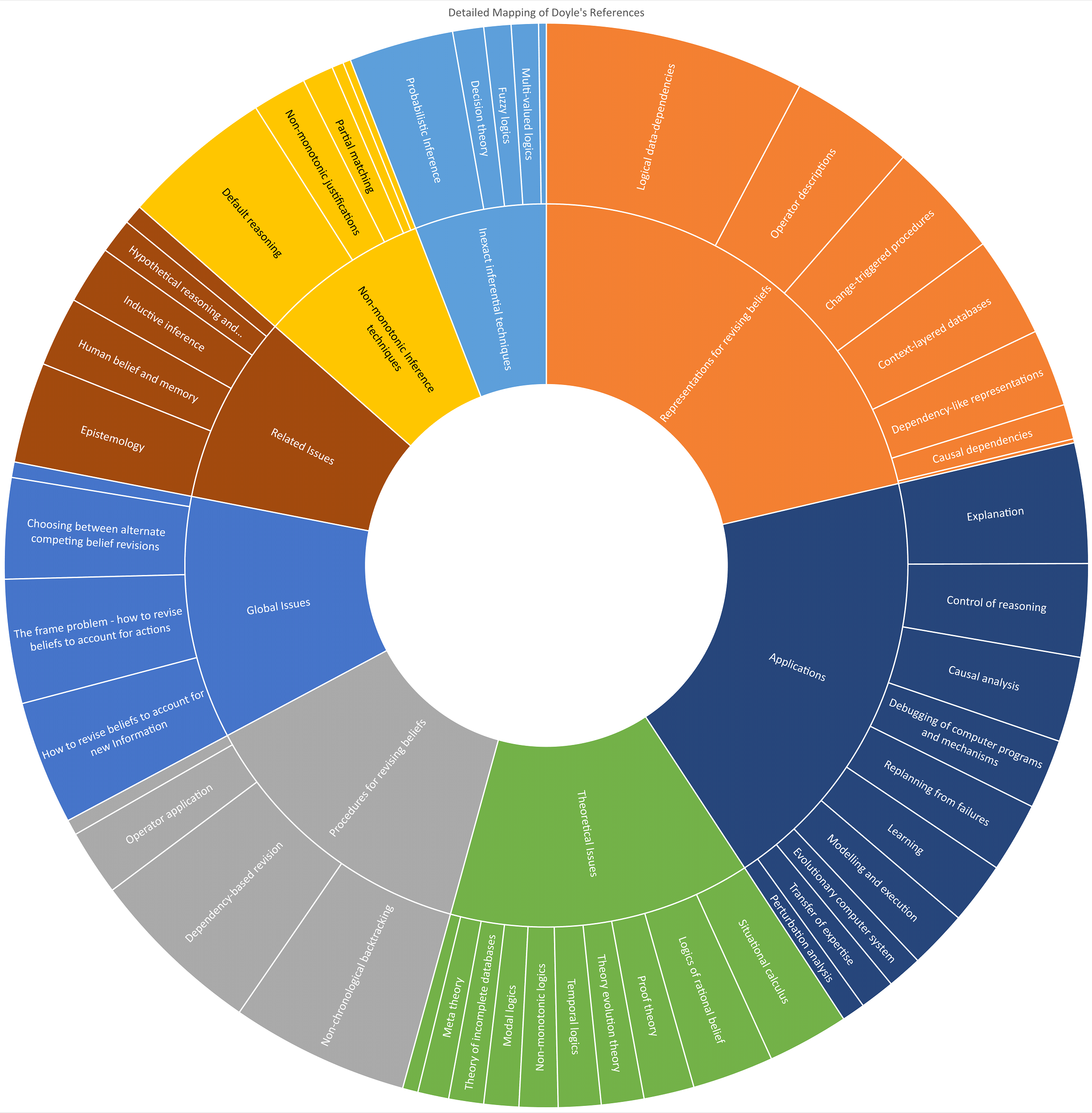}
    \caption{Detailed mapping of Doyle's references (scaled).}
    \label{fig:doyle-wheel}
\end{figure*}

To understand the relative emphasis and development within each category of Doyle and London's taxonomy, we conducted a quantitative analysis of the literature distribution across the eight classification areas. This analysis reveals not only which research directions received the most attention during the pre-AGM period but also provides insight into the computational priorities and theoretical interests that shaped early belief revision research. The distribution patterns illuminate the field's evolution from predominantly procedural and representational concerns toward the more theoretical foundations that would eventually culminate in the AGM framework. Tables~\ref{tab:doyle_category1} to~\ref{tab:doyle_category8} present the numerical breakdown of the articles categorized by Doyle and London within each category, while Figure~\ref{fig:doyle-wheel-small} provides a detailed visual representation of these research concentration patterns. For better visualization, a scaled version of this same figure is shown at the end of the paper (Figure~\ref{fig:doyle-wheel}).

\begin{table}[!t]
\centering
\small
\caption{Global Issues.}
\label{tab:doyle_category1}
\begin{tabular}{@{}p{0.75\linewidth}r@{}}
\toprule
\textbf{Subcategory} & \textbf{\#Refs}\\
\midrule
1. The frame problem& 32\\
2. Alternate competing belief revisions & 26\\
3. Change of Belief or Mind & 4\\
4. Revise beliefs accounting for new information & 32\\
\midrule
\textit{Total} & \textit{94}\\
\bottomrule
\end{tabular}
\end{table}

\begin{table}[!t]
\centering
\small
\caption{Representations for Revising Beliefs.}
\label{tab:doyle_category2}
\begin{tabular}{@{}p{0.75\linewidth}r@{}}
\toprule
\textbf{Subcategory} & \textbf{\#Refs}\\
\midrule
1. Manual updates & 1\\
2. Operator descriptions & 32\\
3. Context-layered databases & 26\\
4. Change-triggered procedures & 30\\
5. Logical data-dependencies & 67\\
6. Causal dependencies & 9\\
7. Dependency-like representations & 20\\
\midrule
\textit{Total} & \textit{185}\\
\bottomrule
\end{tabular}
\end{table}

\begin{table}[!t]
\centering
\small
\caption{Procedures for Revising Beliefs.}
\label{tab:doyle_category3}
\begin{tabular}{@{}p{0.75\linewidth}r@{}}
\toprule
\textbf{Subcategory} & \textbf{\#Refs}\\
\midrule
1. Chronological backtracking & 4\\
2. Operator application & 17\\
3. Dependency-based revision & 45\\
4. Non-chronological backtracking & 46\\
\midrule
\textit{Total} & \textit{112}\\
\bottomrule
\end{tabular}
\end{table}

\begin{table}[!t]
\centering
\small
\caption{Non-Monotonic Inference Techniques.}
\label{tab:doyle_category4}
\begin{tabular}{@{}p{0.75\linewidth}r@{}}
\toprule
\textbf{Subcategory} & \textbf{\#Refs}\\
\midrule
1. Default reasoning & 39\\
2. Non-monotonic justifications & 14\\
3. Negation in databases & 3\\
4. Partial matching & 8\\
5. Minimal model circumscription & 2\\
\midrule
\textit{Total} & \textit{66}\\
\bottomrule
\end{tabular}
\end{table}

\begin{table}[!t]
\centering
\small
\caption{Inexact Inferential Techniques.}
\label{tab:doyle_category5}
\begin{tabular}{@{}p{0.75\linewidth}r@{}}
\toprule
\textbf{Subcategory} & \textbf{\#Refs}\\
\midrule
1. Probabilistic Inference & 27\\
2. Decision theory & 8\\
3. Conclusion theory & 2\\
4. Multi-valued logics & 7\\
5. Fuzzy logics & 7\\
\midrule
\textit{Total} & \textit{51}\\
\bottomrule
\end{tabular}
\end{table}

\begin{table}[!t]
\centering
\small
\caption{Theoretical Issues.}
\label{tab:doyle_category6}
\begin{tabular}{@{}p{0.75\linewidth}r@{}}
\toprule
\textbf{Subcategory} & \textbf{\#Refs}\\
\midrule
1. Logics of rational belief & 21\\
2. Temporal logics & 11\\
3. Situational calculus & 21\\
4. Theory evolution theory & 11\\
5. Proof theory & 13\\
6. Non-monotonic logics & 10\\
7. Meta theory & 8\\
8. Theory of incomplete databases & 9\\
9. Modal logics & 9\\
10. Reasoning about uncertainty & 4\\
\midrule
\textit{Total} & \textit{117}\\
\bottomrule
\end{tabular}
\end{table}

\begin{table}[!t]
\centering
\small
\caption{Applications.}
\label{tab:doyle_category7}
\begin{tabular}{@{}p{0.75\linewidth}r@{}}
\toprule
\textbf{Subcategory} & \textbf{\#Refs}\\
\midrule
1. Explanation & 31\\
2. modeling and execution & 15\\
3. Replanning from failures & 18\\
4. Transfer of expertise & 9\\
5. Learning & 16\\
6. Debugging of computer programs and mechanisms & 18\\
7. Evolutionary computer systems & 9\\
8. Causal analysis & 22\\
9. Perturbation analysis & 6\\
10. Control of reasoning & 24\\
\midrule
\textit{Total} & \textit{168}\\
\bottomrule
\end{tabular}
\end{table}

\begin{table}[!t]
\centering
\small
\caption{Related Issues.}
\label{tab:doyle_category8}
\begin{tabular}{@{}p{0.75\linewidth}r@{}}
\toprule
\textbf{Subcategory} & \textbf{\#Refs}\\
\midrule
1. Human belief and memory & 18\\
2. Epistemology & 26\\
3. Inductive inference & 15\\
4. Hypothetical reasoning and counterfactual conditionals & 9\\
5. Scientific theory change & 5\\
\midrule
\textit{Total} & \textit{73}\\
\bottomrule
\end{tabular}
\end{table}

For the purposes of this review, we focus particularly on extracting and analyzing the computationally representative systems that emerged from categories 2 and 3 of Doyle and London's taxonomy, which deal with representations and procedures for belief revision. These systems embody the algorithmic innovations that would later inform both the development of AGM theory and contemporary implementations in intelligent agents. The computational approaches we examine include the Truth Maintenance Systems (TMS) by~\cite{doyle1979truth}, with their sophisticated dependency tracking mechanisms, the Assumption-Based Truth Maintenance Systems (ATMS) by~\cite{dekleer1986assumption}, that enabled reasoning with multiple contexts simultaneously, the probabilistic belief networks by~\cite{pearl1986fusion}, that introduced uncertainty quantification into belief revision, and various dependency-based reasoning systems that pioneered incremental belief update strategies~\cite{mcallester1982reasoning}. We will detail this approach in Section~\nameref{sec:methodology}.

\subsection{Truth Maintenance Systems: The Doyle Legacy}\label{sec:TMS}

The development of Truth Maintenance Systems (TMS) by~\cite{doyle1979truth} represents one of the most significant pre-AGM contributions to computational belief revision. TMS addresses the fundamental problem of maintaining consistency in knowledge bases by tracking the justifications for beliefs and automatically retracting beliefs when their supporting assumptions are withdrawn.

\subsubsection{Core TMS Architecture}

The TMS architecture consists of several key components:

\textbf{Belief Nodes:} Represent individual beliefs or propositions, each associated with a status (IN or OUT) indicating whether the belief is currently accepted;

\textbf{Justification Structures:} Encode the inferential relationships between beliefs, specifying which combinations of beliefs support or contradict other beliefs;

\textbf{Dependency Networks:} Maintain explicit records of how beliefs depend on each other, enabling efficient propagation of belief changes.

\textbf{Contradiction Detection:} Mechanisms for identifying inconsistencies and triggering belief revision processes.

The TMS operates through a cycle of assumption making, inference, contradiction detection, and dependency-directed backtracking. When a contradiction is detected, the system identifies the minimal set of assumptions that must be retracted to restore consistency, using the recorded dependency structure to guide this process efficiently.

\subsubsection{Algorithmic Innovations}

Doyle's TMS introduced several algorithmic innovations that remain influential:

\textbf{Dependency-Directed Backtracking:} Instead of chronological backtracking that undoes the most recent assumptions first, TMS uses dependency information to identify and retract only those assumptions that are causally responsible for contradictions;

\textbf{Incremental Consistency Maintenance:} Rather than rebuilding the entire belief set after each change, TMS propagates the changes incrementally through the dependency network;

\textbf{Assumption-Based Reasoning:} TMS explicitly represents and manages assumptions, enabling systematic exploration of alternative assumption sets;

\textbf{Justification-Based Inference:} All beliefs are associated with explicit justifications, providing transparency and enabling sophisticated belief revision strategies;

\subsubsection{Implementation Details}

~\cite{doyle1978truth} provided detailed implementation descriptions and annotated code for TMS algorithms. Key implementation features include:

\textbf{Node Data Structures:} Each belief node maintains pointers to its justifications, consequences, and current status, enabling efficient traversal of the dependency network;

\textbf{Justification Records:} Encode support relationships using lists of antecedent nodes and subsequent nodes, with special handling for conditional and default rules;

\textbf{Propagation Algorithms:} Breadth-first and depth-first search procedures for propagating belief changes through the network while maintaining consistency;

\textbf{Conflict Resolution:} Algorithms for identifying minimal conflict sets and selecting which assumptions to retract when multiple resolution strategies are possible.

\subsection{Assumption-Based Truth Maintenance Systems (ATMS)}

\cite{dekleer1986assumption}'s Assumption-Based Truth Maintenance System (ATMS)  extends TMS by maintaining multiple consistent belief bases simultaneously, each corresponding to different assumption combinations. This approach avoids the computational overhead of repeatedly computing belief revisions by precomputing the consequences of different assumption sets.

\subsubsection{ATMS Architecture}

The ATMS maintains:

\textbf{Assumption Sets:} Collections of assumptions that can be made simultaneously without contradiction;

\textbf{Labels:} Each belief node is associated with a label indicating which assumption sets support that belief;

\textbf{Nogood Sets:} Collections of assumptions that lead to contradictions and must be avoided;

\textbf{Justification Networks:} Similar to TMS but extended to handle multiple contexts simultaneously.

\subsubsection{Computational Advantages}

ATMS provides several computational advantages over single-context TMS:

\textbf{Context Switching:} Rapid switching between different assumption sets without recomputation;

\textbf{What-If Analysis:} Efficient exploration of hypothetical scenarios by examining different assumption combinations;

\textbf{Parallel Reasoning:} Multiple reasoning contexts can be maintained and explored simultaneously;

\textbf{Assumption Minimality:} Automatic identification of minimal assumption sets sufficient to support particular conclusions.

\subsection{Probabilistic Belief Networks}

\cite{pearl1986fusion}'s work on probabilistic belief networks (later known as Bayesian networks) provided a fundamentally different approach to belief revision based on probability theory rather than logical inference. These networks represent probabilistic dependencies between variables and support belief revision through Bayesian conditioning.

\subsubsection{Network Structure}

Probabilistic belief networks consist of:

\textbf{Nodes:} Represent random variables corresponding to propositions or events;

\textbf{Directed Edges:} Encode probabilistic dependencies between variables;

\textbf{Conditional Probability Tables:} Specify the probability distributions for each variable given its parents in the network;

\textbf{Evidence Nodes:} Represent observed information that constrains the probability distributions.

\subsubsection{Belief Revision Algorithms}

Pearl developed several algorithms for belief revision in probabilistic networks:

\textbf{Message Passing:} For singly-connected networks, belief revision can be performed efficiently using local message-passing algorithms that propagate probability updates through the network;

\textbf{Clustering Methods:} For multiply-connected networks, clustering algorithms group variables to reduce computational complexity while maintaining exact inference;

\textbf{Approximation Methods:} When exact inference is intractable, various approximation methods provide efficient approximate belief revision.

\subsubsection{Computational Complexity}

The computational complexity of belief revision in probabilistic networks depends on network structure~\cite{pearl1986fusion}:

\textbf{Singly-Connected Networks:} Linear time complexity in network size using message-passing algorithms.

\textbf{Multiply-Connected Networks:} Exponential worst-case complexity, but often tractable for networks with appropriate structure.

\textbf{Approximation Trade-offs:} Various approximation methods~\cite{chopra2001approximate} provide different trade-offs between accuracy and computational efficiency.

\subsection{Default Reasoning Systems}

\cite{reiter1980logic}'s default logic and related non-monotonic reasoning systems provided another pre-AGM approach to belief revision. These systems handle incomplete information by making default assumptions that can be retracted when contradictory information is discovered.

\subsubsection{Default Logic Framework}

Default logic extends classical logic with default rules of the form:
$$\frac{\alpha : \beta_1, \ldots, \beta_n}{\gamma}$$

This rule states that if $\alpha$ is believed and it is consistent to believe $\beta_1, \ldots, \beta_n$, then $\gamma$ can be concluded by default.

\subsubsection{Computational Approaches}

Several computational approaches to default reasoning were developed:

\textbf{Fixed-Point Algorithms:} Compute default extensions by iteratively applying default rules until a fixed point is reached;

\textbf{Proof-Theoretic Methods:} Develop proof procedures that can handle default rules directly without computing complete extensions;

\textbf{Compilation Approaches:} Transform default theories into classical logic theories that can be processed using standard inference engines.

\subsection{Dependency-Based Approaches}\label{sec:dependencyBased}

Several pre-AGM systems focused explicitly on dependency tracking and management.

\subsubsection{Reason Maintenance Systems}

In his reason maintenance system \cite{mcallester1982reasoning} provides a different approach to dependency tracking that emphasizes computational efficiency and incremental reasoning.

\subsubsection{Causal Networks}
Various researchers developed causal network approaches that represent causal relationships explicitly and use them to guide belief revision processes~\cite{pearl1990causal, pearl2009causality, spirtes1993causation, dean1989model, halpern2005causes}.

While many of these algorithms lacked the formalism introduced by the AGM model, they represented important efforts to address belief revision within the AI domain. These algorithms often adopted pragmatic approaches to updating belief bases, which included:

\begin{itemize}
    \item \textbf{Updating heuristics}: Many pre-AGM algorithms were based on simple heuristics for adding or removing beliefs from the knowledge base, without considering the logical consequences of these actions. Although these techniques were computationally efficient in terms of resources, they frequently resulted in mechanisms lacking rigor, leading systems into inconsistent states~\cite{doyle1980selected,fikes1971strips,hewitt1972planner}.
    \item \textbf{Probability-based models}: Some pre-AGM approaches used probabilistic methods for adjusting beliefs upon receiving new information. These approaches often attempted to incorporate probabilistic techniques such as Bayesian statistics. However, the lack of rigor also led to difficulties, particularly regarding scalability and applicability to systems based on pure logic~\cite{pearl1986fusion,pearl1990causal,dean1989model}.
    \item \textbf{Rule-based systems}: Another common approach in the 1970s involved rule-based systems, in which new beliefs were inserted into or removed from the knowledge base according to a set of predefined rules. While such systems were useful in certain scenarios, they were unable to handle contradictions or inconsistencies in the information received, as the rules were not adaptive~\cite{fikes1971strips,hewitt1972planner,genesereth1981dart}.
\end{itemize}

\subsection{Additional Systems from Doyle's Bibliography}\label{sec:additional_preAGM}

Beyond the major systems detailed above, Doyle and London's bibliography documents several other influential pre-AGM approaches that contributed to the computational foundations of belief revision:

\textbf{STRIPS and Operator-Based Planning:}~\cite{fikes1971strips} introduced operator descriptions with explicit add and delete lists, providing early mechanisms for representing state changes. While primarily an action planning system, STRIPS's approach to minimal change (only specified facts change) anticipated AGM's inclusion postulate and informed the later distinction between belief revision and belief update~\cite{katsuno1991propositional}.

\textbf{CONNIVER and Context Mechanisms:}~\cite{mcdermott1972conniver} introduced context layers and non-chronological backtracking, enabling efficient exploration of alternative belief states. These mechanisms anticipated ATMS's multi-context reasoning and informed theoretical work on modularity and independence in belief revision~\cite{parikh1999beliefs}.

\textbf{PLANNER and Pattern-Directed Invocation:}~\cite{hewitt1972planner} provided early mechanisms for change-triggered procedures, where belief changes automatically invoke relevant inference rules. This approach influenced TMS's justification propagation and anticipated active database triggers used in contemporary belief base implementations.

\textbf{Limitations of Pre-AGM Implementations:} Most pre-AGM systems were demonstrated primarily on toy examples and small-scale domains due to 1980s computational constraints. Notable exceptions that achieved real-world deployment include the use of a TMS-style reasoning architecture in the DART expert system~\cite{genesereth1981dart} and ATMS in the GDE diagnostic system~\cite{dekleer1987diagnosing}. Contemporary implementations benefit from orders-of-magnitude improvements in processing power and memory, potentially making pre-AGM algorithmic insights more practically viable at scale.

Despite the lack of formalism, the work undertaken during this period contributed significantly to the initial development of belief revision within AI, establishing an important foundation for subsequent, more formal approaches.

\subsection{From Computational Practice to Theoretical Foundation}

The pre-AGM period established computational belief revision as a legitimate and necessary component of intelligent systems, with each approach contributing distinct algorithmic innovations that would later inform theoretical developments. The TMS pioneered dependency tracking and non-chronological backtracking~\cite{doyle1979truth}, ATMS introduced multi-context reasoning capabilities~\cite{dekleer1986assumption}, probabilistic networks brought uncertainty quantification to belief revision~\cite{pearl1986fusion}, and default reasoning systems provided frameworks for handling incomplete information~\cite{reiter1980logic}. Our mapping (Figure \ref{fig:doyle-wheel-small} and tables \ref{tab:doyle_category1} to \ref{tab:doyle_category8}) shows that 47\% of papers focused on procedural approaches (Category 3), while 77\% concentrated on representational innovations (Category 2), 70\% focused on applications (Category 7), and 48\% on theoretical issues - most of them related to practical aspects of belief revision (Category 6), demonstrating the field's primary emphasis on computational tractability and practical implementation during this formative period. 

Doyle's contribution extends beyond mere cataloging; his work with London established a taxonomical framework that anticipated the theoretical developments that would follow with AGM while capturing the diversity of computational approaches being explored at the time~\cite{doyle1980selected}. This taxonomy proved remarkably prescient, identifying key dimensions of belief revision research that remain relevant today and providing a structured lens through which to understand the relationship between early computational work and later theoretical developments~\cite{doyle1979truth, doyle1988knowledge}. The distribution patterns in our analysis show that out of the eight categories in the taxonomy, categories 2, 3, 6, and 7 received the most attention, reflecting the pragmatic priorities of AI researchers who needed working systems before formal theories were available. It is also important to mention that while there were many papers that fit into multiple categories, the papers in categories 2, 3, 6, and 7 represent 85\% of the works analyzed by Doyle and London.

This move naturally follows the historical thread presented in the origins of belief revision research. The foundational concerns of maintaining a model of the world, reasoning under changing information, and coping with inconsistency emerged from early AI domains such as planning, vision, and language processing, as well as from adjacent philosophical work on theory change~\cite{mccarthy1981some, mccarthy1980circumscription}. Doyle's taxonomy codifies these concerns as \emph{named axes of practice}, transforming ad hoc engineering solutions into systematic categories of investigation. It represents a pivot point between the \emph{pre-AGM era} of algorithmic devices, dependence tracking, non-chronological backtracking, and defaulting machinery, and the later \emph{AGM paradigm} of postulates and representation theorems that would provide normative foundations for belief revision~\cite{alchourron1985logic, gardenfors1988knowledge}. In essence, the taxonomy captures what engineers were already doing to keep beliefs coherent while the formal postulate-based account was still on the horizon.

The computational approaches examined in this section represent more than historical curiosities; they embody algorithmic insights and implementation strategies that remain relevant for contemporary belief revision systems. However, understanding their true significance requires evaluating them through the lens of the theoretical framework that emerged in the mid-1980s with the AGM model. The question that naturally arises is: how do these pre-AGM computational innovations relate to the rationality postulates and construction methods that AGM theory would later establish? To address this question systematically, we must develop a methodological framework that allows us to re-examine Doyle and London's taxonomy and the computational systems it encompasses from a post-AGM theoretical perspective, assessing both their contributions to and limitations within the broader evolution of belief revision research.

\section{Re-examining Pre-AGM Systems Through Post-AGM Theory} \label{sec:methodology}

The emergence of the AGM framework in the mid-1980s fundamentally transformed the theoretical landscape of belief revision, providing rigorous axiomatic foundations that were absent during the pre-AGM era~\cite{alchourron1985logic, gardenfors1988knowledge}. This transformation creates both an opportunity and a methodological challenge: how can we systematically re-evaluate the rich computational heritage of pre-AGM systems through the clarifying lens of AGM theory? Our approach addresses this challenge by mapping Doyle and London's empirically-derived taxonomy onto the theoretical framework established by AGM, thereby bridging the gap between historical computational practice and contemporary theoretical understanding.

\subsection{Literature Selection and Scope}\label{sec:lit_selection}

Our literature selection strategy is grounded in Doyle and London's foundational 1980 bibliography while extending systematically to encompass key post-AGM developments. The selection process operates through three complementary approaches:

\textbf{Core Historical Foundation:} We begin with the approximately 250 papers cited in~\cite{doyle1980selected}'s original bibliography, which provides comprehensive coverage of pre-AGM computational approaches to belief revision. This foundation ensures historical completeness and maintains continuity with the taxonomical framework that organizes our analysis.

\textbf{Theoretical Development Tracking:} We systematically include key papers that established and developed the AGM framework, beginning with the foundational work by Alchourrón, Gärdenfors, and Makinson~\cite{alchourron1985logic} and extending through major theoretical developments including epistemic entrenchment~\cite{gardenfors1988knowledge}, kernel methods~\cite{hansson1994kernel}, iterated revision~\cite{darwiche1997logic}, and belief base approaches~\cite{nebel1989knowledge}. Although the AGM framework consolidated these ideas in a unified post-1985 formulation, key elements of belief change theory had already been developed earlier, including Gärdenfors’ initial proposal of rationality postulates (\citeyear{gardenfors1982rules}) and the first meet and partial meet constructions introduced by Alchourrón and Makinson (\citeyear{alchourronmakinson1982}).

\textbf{Contemporary Implementation Analysis:} We include representative contemporary work that demonstrates how AGM principles have been implemented in practical systems, with particular attention to papers that explicitly connect theoretical foundations with computational approaches.

Our temporal scope extends from 1940 to 2024, with particular emphasis on the critical period from 1970-1990, which encompasses both the pre-AGM computational era and the emergence of AGM theory. We include work from artificial intelligence, philosophy, logic, computer science, and related fields, maintaining interdisciplinary coverage that reflects the field's diverse intellectual foundations.

Selection criteria prioritize works that: (1) establish foundational concepts or methods; (2) provide computational implementations or algorithmic contributions; (3) bridge pre-AGM and post-AGM approaches; (4) demonstrate practical applications of belief revision principles; (5) explicitly address the relationship between different theoretical or computational approaches. The detailed methodology is presented in Appendix~\nameref{app1}.

\subsection{Analytical Framework}\label{sec:analytical_framework}

Our analytical framework combines historical analysis with systematic theoretical evaluation through several complementary methodological components:

\textbf{Taxonomical Mapping:} We systematically map computational approaches identified in Doyle and London's bibliography to their taxonomical categories, analyzing how these categorizations reflect the empirical structure of pre-AGM research. This mapping reveals patterns of research emphasis and identifies computational innovations that were most influential in subsequent theoretical development.

\textbf{Postulate Correspondence Analysis:} For each pre-AGM computational approach, we analyze its relationship to specific AGM postulates, examining both direct correspondences and areas where computational practice anticipated theoretical development. This analysis reveals how empirical insights about rational belief change were later formalized through AGM theory.

\textbf{Evolution Tracing:} We systematically trace how each category in Doyle's taxonomy evolved in the post-AGM era, identifying key theoretical developments, seminal papers, and transformation patterns. This tracing reveals how computational insights were formalized, extended, and sometimes superseded by theoretical advances.

\textbf{Synthesis Identification:} Throughout our analysis, we identify opportunities for synthesizing insights from different eras of research, examining how pre-AGM computational strategies can inform contemporary implementations of AGM-based systems.

\subsection{Methodological Implementation}

To systematically implement this correspondence analysis, we reconstruct the distribution of entries across Doyle's original taxonomy and analyze how each subcategory has evolved in the post-AGM literature. This reconstruction process involves several methodological steps. \textbf{First}, we identify the computational systems and approaches that Doyle and London classified under each taxonomical category. \textbf{Second}, we analyze how these systems relate to specific AGM postulates, construction methods, or theoretical insights. \textbf{Third}, we trace the evolution of each subcategory through subsequent theoretical developments, identifying key papers and theoretical advances that formalized, extended, or refined the original computational insights.

While Doyle’s descriptor-based approach captures the complexity of early systems, its resulting lack of mutually exclusive categories (multi-labeling) and heterogeneous granularity pose challenges for a structured classification. However, these characteristics do not undermine its utility for our analysis. Rather, these limitations justify our AGM-aware refinements and highlight the theoretical progress made since 1980. Our methodology acknowledges these limitations while leveraging the taxonomy's enduring insights about the fundamental dimensions of belief revision research.

It is important to note that while our analysis is comprehensive within its scope, certain aspects of contemporary belief revision research lie beyond the boundaries of both Doyle's original taxonomy and our AGM-focused extension. Social dimensions of belief revision, including how social contexts and multi-agent interactions influence belief change, represent important areas of current research that were not anticipated in the pre-AGM era~\cite{baltag2008qualitative}. Similarly, emotional and psychological factors that affect human belief revision, while relevant for cognitive modeling applications, fall outside the computational and logical focus of both the original taxonomy and our theoretical analysis.

Our methodological framework provides the analytical tools necessary to examine how Doyle's taxonomy evolved in the post-AGM era. We now apply this framework systematically to trace the theoretical transformation of each taxonomical category, demonstrating how empirical observations about computational practice were formalized into rigorous theoretical constructs while identifying both continuities and innovations in this evolution.

\section{Taxonomical Evolution in the Post-AGM Era} \label{sec:taxonomy_evolution}

Doyle's descriptor-indexed taxonomy captured the \emph{practice} of belief maintenance in the pre-AGM era through empirical observation and systematic categorization of existing computational approaches. The AGM paradigm fundamentally reframed this landscape by introducing a \emph{normative}, postulate-based view of rational belief change supported by rigorous representation theorems~\cite{alchourron1985logic, gardenfors1988knowledge}. In this section, we systematically revisit each of Doyle's taxonomic categories and trace their evolution in the post-AGM literature, highlighting new theoretical derivations, conceptual unifications, and practical ramifications. Our aim extends beyond mere reference compilation to demonstrate \emph{how} each theoretical thread matured into components of the modern belief revision toolkit and how AGM's normative framework both validated and transformed pre-AGM computational insights.

The relationship between Doyle's empirical taxonomy and AGM's theoretical framework exhibits both continuity and transformation. AGM reframes belief revision as a normative problem about rational change subject to specific postulates, including closure, success, inclusion, consistency preservation, vacuity, extensionality, and additional constraints for iterated revision, while providing representation theorems that connect algebraic choice functions to belief change operators through constructions like partial-meet contraction and Levi and Harper identities~\cite{gardenfors1988knowledge}. This theoretical foundation validates many pre-AGM computational intuitions while providing systematic justification for design choices that were previously ad hoc.

\subsection{Global Issues: From Computational Problems to Theoretical Postulates}\label{sec:tax_GI}

\subsubsection{The Frame Problem and Action-Driven Change}

Doyle's original framing merged \emph{state change by action} with \emph{belief change by information}, reflecting the practical reality that early AI systems needed to handle both types of change within unified computational frameworks. Post-AGM theoretical work systematically separated these concerns, leading to the fundamental distinction between \emph{belief revision} and \emph{belief update}~\cite{katsuno1991propositional}. This separation proved theoretically crucial: AGM-style \emph{revision} integrates new information with minimal change to existing beliefs, while \emph{update} handles world change due to actions or events by modeling how the world itself has changed rather than how our information about a static world has improved.

This theoretical distinction led to dual semantic frameworks that formalize the intuitive difference between learning new facts and accommodating world changes. Belief revision employs semantic approaches, including faithful assignments~\cite{katsuno1991propositional}, \cite{grove1988modelings}'s system of spheres, and distance-based minimization~\cite{dalal1988investigations} that prioritize minimal change to belief content. Belief update, conversely, employs possible-model approaches and persistence principles~\cite{winslett1991updating} to model how the world evolves over time. In contemporary planning and knowledge representation systems, the frame problem is typically addressed through action formalisms such as successor-state axioms or update operators that incorporate inertia principles, while AGM-style revision handles sensing, communication, and information-gathering activities.

\subsubsection{Choosing Among Competing Revisions}

Doyle's theme of \emph{choice among alternatives} becomes explicit and systematic in post-AGM representation theorems through the machinery of selection functions and preference orderings. Partial-meet contraction formalizes this choice process by selecting among maximal consistent subsets (remainders) through selection functions that encode preferences over different ways of restoring consistency~\cite{alchourron1985logic}. Epistemic entrenchment provides an alternative formalization where choices are guided by resistance to contraction, with more entrenched beliefs being more difficult to give up~\cite{gardenfors1988knowledge}.

The distance-based and sphere semantics developed by~\cite{grove1988modelings} and others~\cite{dalal1988investigations} reify choice as minimization problems or proximity relationships in logical space, providing computational methods for implementing preference-guided revision. This theoretical line matured into sophisticated frameworks for \emph{rational preferences over change}, including \cite{parikh1999beliefs}'s work on relevance and splitting that enables modular belief revision, and various approaches to priority-based and relevance-sensitive change that maintain computational tractability while respecting normative constraints.

\subsubsection{Change of Belief or Mind: Iterated Revision}

The original AGM framework addressed single episodes of belief change, but Doyle's subcategory of "change of mind" explicitly identified belief change over time as a distinct problem class, anticipating the need for principled approaches to sequences of belief revisions. Post-AGM work systematically addressed this limitation by studying iterated belief revision, asking how sequences of inputs should interact and what rationality constraints should govern belief evolution over time.

The Darwiche-Pearl (DP) postulates(~\citeyear{darwiche1997logic}) for iterated revision sparked extensive research into iteration policies, leading to a sophisticated taxonomy of approaches including \emph{natural} and \emph{lexicographic} revision~\cite{boutilier1996iterated}, \emph{restrained} and \emph{priority}-based iterations that balance stability with responsiveness to new information~\cite{booth2006admissible}, and ordinal-ranking approaches based on ~\cite{spohn1988ordinal}'s ranking functions that connect belief revision to conditional logic and default reasoning. These theoretical advances provide precise answers to Doyle's question about "how to change one's mind again" by specifying rational dynamics for belief evolution.

\subsubsection{Absorbing New Information with Minimal Disruption}

Doyle's informal notion of "minimal disruption" receives precise theoretical treatment through AGM's central theoretical results. The \emph{Levi identity} ($K * \phi = (K - \neg\phi) + \phi$) and \emph{Harper identity} ($K - \phi = K \cap (K * \neg\phi)$) establish fundamental relationships between revision, contraction, and expansion operations~\cite{gardenfors1988knowledge}. Partial-meet construction methods provide systematic procedures for implementing these operations through choice functions that select among maximal consistent subsets according to specified preference criteria.

The distance-based and sphere semantics provide equivalent minimization perspectives that make the notion of "minimal change" mathematically precise through metrics on possible worlds or logical theories~\cite{grove1988modelings, dalal1988investigations}. Subsequent complexity and compilability results~\cite{eiter1992complexity, liberatore1996complexity} clarify the computational feasibility of these theoretical constructions, providing guidance for practical implementation while maintaining theoretical soundness.

\subsection{Representations: From Data Structures to Theoretical Constructs}\label{sec:tax_Rep}

\subsubsection{Manual Updates and Operator Descriptions}

STRIPS-style add/delete lists, which Doyle classified as operator descriptions for manual updates, became the canonical vocabulary for \emph{action update} in planning systems and were formally distinguished from \emph{informational revision}~\cite{fikes1971strips}. Post-AGM theoretical work clarified that revision should be represented intensionally through belief sets, belief bases, and entrenchment orderings, while operator descriptions belong to the update side of the revision/update distinction~\cite{winslett1991updating}. This clarification resolved the pre-AGM conflation that Doyle noted between different types of belief change by providing distinct theoretical frameworks for different change scenarios.

\subsubsection{Context Layers and Modularity}

Doyle's notion of layered contexts anticipated important theoretical developments in \emph{independence} and \emph{modularity} for belief revision. Post-AGM work made these intuitions precise through formal results like Parikh's splitting theorem, which demonstrates that when a logical language decomposes into independent components, belief revision can be performed component-wise without undesirable interactions between modules~\citeyear{parikh1999beliefs}. Relevance and independence constraints became design criteria for practical belief revision operators and informed the development of belief base policies that respect modular structure~\cite{hansson1999textbook}.

\subsubsection{Change-Triggered Procedures and Dependency Structures}

Doyle's emphasis on logical data dependencies evolved into the general theory of \emph{belief bases} with \emph{kernel} constructions and \emph{incision} functions that provide systematic methods for base-level belief revision~\cite{hansson1994kernel}. To contract a belief base by a formula $\phi$, kernel methods identify all minimal subsets that imply $\phi$ (the $\phi$-kernels) and remove at least one formula from each kernel through incision functions that encode preferences over which beliefs to sacrifice.

This base-level perspective supports the fine-grained control over belief change—including traceability, minimal deletion, and explicit justification management—that Doyle sought in dependency graph approaches, while maintaining compatibility with AGM's postulate-based framework. The kernel approach bridges the gap between AGM's abstract theoretical requirements and the concrete implementation needs of practical belief revision systems~\cite{ferme2018belief}.

\subsubsection{Causal and Justificatory Encodings}

Pre-AGM approaches to non-monotonic dependencies and causal relationships matured along two complementary theoretical directions in the post-AGM era. \textbf{AGM-compatible approaches} encode support relationships through entrenchment orderings or kernel-based methods, where causal knowledge guides incision functions to produce domain-aware contraction that respects causal structure. \textbf{Conditional and ordinal approaches} employ \cite{spohn1988ordinal}'s ranking functions and~\cite{pearl1990system}'s System Z to map justificatory relationships into numerical or ordinal structures that drive both belief revision and default inference, providing unified frameworks for handling both definite information and defeasible patterns.

Dynamic Epistemic Logic (DEL)~\cite{baltag1998logic1, ditmarsch2008dynamic} later provided event-driven \emph{action models} with explicit preconditions and postconditions for updating epistemic states, offering systematic approaches to the causal and procedural relationships that pre-AGM systems handled through ad hoc mechanisms. These developments formalize Doyle's intuitions about change-triggered procedures while providing rigorous semantic foundations.

\subsection{Procedures: From Algorithms to Semantic Characterizations}\label{sec:tax_Proc}

\subsubsection{Operator Application and Backtracking Strategies}

Doyle's operational catalogue of procedural approaches—including chronological and non-chronological backtracking, operator application, and dependency-based revision—evolved into \emph{semantic characterizations} that subsume many specific algorithmic approaches under unified theoretical frameworks. Partial-meet construction corresponds to remainder selection guided by preference criteria; sphere-based semantics implements locality through distance relationships; ranking-based approaches organize preferences as priority queues over possible worlds or belief states.

Non-chronological backtracking, pioneered in systems like TMS~\cite{doyle1979truth}, corresponds to \emph{dependency-directed} change at the belief base level through kernel and incision methods~\cite{hansson1994kernel}. Iterated revision policies lift these base-level procedures to handle streams of inputs while maintaining rationality constraints across multiple revision episodes~\cite{darwiche1997logic, boutilier1996iterated}.

\subsubsection{Dependency-Based Revision}

Dependency-based revision became the canonical \emph{belief base revision} pipeline in post-AGM theory: compute kernels to identify minimal inconsistent subsets; choose incisions according to minimality criteria or priority orderings; reconstruct a revised base from the remaining beliefs; apply logical closure if needed for specific applications~\cite{hansson1994kernel, ferme2018belief}. This systematic approach formalizes the dependency-tracking intuitions of pre-AGM systems while providing theoretical guarantees about the rationality of the resulting belief changes.

Specialized procedures for restricted domains—including Horn clauses, description logics, and other logical fragments—require adapted methods with appropriate definability and complexity guarantees~\cite{eiter1992complexity, flouris2008ontology}. These domain-specific approaches maintain the essential insights of dependency-based revision while exploiting structural properties of particular logical languages to achieve better computational performance.

\subsection{Non-Monotonic Inference: Orthogonality and Integration}\label{sec:tax_NonMono}

\subsubsection{Defaults and Justifications versus Revision}

Post-AGM theoretical work systematically clarified the relationship between belief revision and non-monotonic reasoning that Doyle's taxonomy implicitly recognized, while also motivating accounts in which the two are seen as complementary perspectives on the same underlying dynamics of belief change. This relationship has been characterized both as \emph{orthogonality} (where the two address fundamentally different problems) and as \emph{duality} (where they represent complementary perspectives on rational belief change~\cite{gardenfors1991nonmonotonic,rott1991two}).

\cite{gardenfors1991nonmonotonic} demonstrates that belief revision and non-monotonic reasoning can be viewed as "two sides of the same coin," with non-monotonic consequence relations corresponding to families of revision operations and vice versa.~\cite{rott1991two} further develops this connection by identifying two fundamental dimensions of belief change: the \emph{informational dimension} (captured by AGM revision) and the \emph{motivational dimension} (captured by non-monotonic inference), showing how these dimensions interact in rational belief dynamics.

Belief revision addresses changing commitments in light of new factual information, while default logics and circumscription govern tentative inference patterns in the absence of complete information~\cite{reiter1980logic, mccarthy1980circumscription}. The KLM framework for rational closure~\cite{kraus1990nonmonotonic} and Spohn's ranking functions~\cite{spohn1988ordinal} provide rigorous semantics for default entailment that complement rather than compete with AGM-style belief revision.

\subsubsection{Interaction Principles}

Despite their theoretical orthogonality, AGM revision and non-monotonic reasoning systems exhibit systematic relationships at the semantic level through ranking functions and faithful assignments. The Darwiche-Pearl postulates (~\citeyear{darwiche1997logic}) for iterated revision constrain how conditional beliefs evolve across revision episodes, while natural versus lexicographic revision policies correspond to different approaches for re-prioritizing conditional information after receiving new inputs~\cite{boutilier1996iterated}. These connections bridge Doyle's category of "justifications" with modern conditional semantics and iterated belief change.

\subsection{Inexact Inferential Techniques: Qualitative and Quantitative Bridges}\label{sec:tax_Inex}

\subsubsection{Probabilistic and Graded Belief Change}

Doyle's category of inexact inferential techniques anticipated important connections between AGM's qualitative approach and probabilistic methods for belief under uncertainty. While Bayesian conditioning differs fundamentally from AGM revision, principled theoretical bridges exist through limiting cases and qualitative probability theory. 

Axiomatic connections between qualitative and quantitative belief have been extensively explored through multiple frameworks~\cite{goldzsmidt1993maximum,darwiche1997logic,boutilier1994conditional,friedman1999belief}: Spohn's ranking functions correspond to limit cases of probability measures~\cite{spohn1988ordinal}, where ordinal conditional functions emerge as qualitative abstractions of probabilistic belief; Jeffrey-style probability updates relate to specific iterated revision policies~\cite{jeffrey1983logic}, with connections formalized through imaging operators; and representation theorems establish conditions under which qualitative preference orderings correspond to underlying probability measures~\cite{krantz1971foundations}.

Possibility and necessity measures produce possibility-based revision and update operators with AGM-like properties under appropriate qualitative axioms~\cite{dubois1993belief}, providing alternative approaches to uncertain belief change that maintain AGM's insights about minimal change while accommodating numerical uncertainty representations.

\subsubsection{Decision-Theoretic Choice Functions}

The AGM framework's selection step in partial-meet contraction can be enhanced with utility-theoretic considerations by choosing remainders that minimize expected loss or maximize expected value under relevant integrity constraints~\cite{hansson1999textbook}. This approach links \emph{belief change} with \emph{value-guided} decision making in applications where the costs and benefits of different beliefs can be meaningfully quantified, extending AGM's normative framework to accommodate practical decision-making contexts.

\subsection{Theoretical Issues: Formalization and Unification}\label{sec:tax_TI}

\subsubsection{Representation Theorems and Semantic Equivalences}

AGM's partial-meet and entrenchment representations~\cite{alchourron1985logic, gardenfors1988knowledge} were systematically unified with Grove's sphere semantics~\cite{grove1988modelings} and distance-based characterizations~\cite{dalal1988investigations, katsuno1991propositional} through a series of representation theorems that demonstrate the equivalence of different approaches to formalizing minimal change. These theoretical equivalences instantiate Doyle's concern with "proof theory" and "meta theory" through precise mathematical derivations that connect syntactic and semantic approaches to belief revision.

\subsubsection{Belief Bases versus Belief Sets}

The distinction between belief \emph{bases} (finite, non-closed sets of explicit beliefs) and belief \emph{sets} (infinite, logically closed sets of all believed consequences) emerged as a central theoretical issue in post-AGM work~\cite{hansson1994kernel, nebel1998hard}. Belief bases enable non-closed repositories with explicit traceability and fine-grained control over belief change, while belief sets provide the mathematical precision necessary for theoretical analysis. Kernel contraction and related methods provide constructive, modular approaches to base-level change with postulate soundness guarantees~\cite{ferme2018belief}.

For restricted logical fragments including Horn clauses and description logics, definability preservation and interpolation properties become central theoretical concerns~\cite{eiter1992complexity,flouris2006thesis,flouris2008ontology}, leading to specialized approaches that maintain the essential insights of AGM theory while exploiting the structural properties of particular logical languages.

\subsubsection{Ontology Repair and Description Logic Extensions}

The extension of AGM belief revision principles to description logic ontologies represents an important theoretical development that addresses practical needs in semantic web and knowledge graph applications. Description logics provide the formal foundations for ontology languages like OWL, which are widely deployed in enterprise knowledge management, biomedical informatics, and linked data systems~\cite{baader2003description}.

Recent work has established formal connections between classical AGM contraction operations and ontology repair mechanisms. \cite{KR2024-9} demonstrate that AGM-style contractions can be defined for description logic ontologies through optimal repair strategies that minimize the removal of axioms while restoring consistency. Their approach adapts partial meet contraction to the description logic setting by identifying minimal inconsistent subsets (analogous to kernels in belief base contraction) and applying incision functions that respect the structural properties of description logic axioms.

Souza's comprehensive analysis bridges belief revision theory with ontology repair implementations, demonstrating how AGM postulates can be satisfied in description logic contexts while maintaining computational tractability through approximation strategies~\cite{souza2024bridging}. This work reveals that many pre-AGM computational insights, including and particularly dependency tracking and modular repair strategies, remain relevant when adapted to the more expressive logical frameworks required for contemporary ontology applications.

Contemporary ontology repair implementations demonstrate the practical synthesis of AGM theory with computational efficiency concerns. Several systems exemplify this integration:

\textbf{Pellet and HermiT Reasoners:} These description logic reasoners incorporate justification-based repair mechanisms, following the justification-based explanation framework introduced by Horridge~\cite{horridge2011justification}, and implement kernel-style contraction for OWL ontologies, providing interactive debugging tools that help ontology engineers identify and resolve inconsistencies~\cite{sirin2007pellet,shearer2008hermit};

\textbf{OntoClean and OOPS!:} Ontology validation and repair tools that combine automated consistency checking with user-guided repair strategies, addressing the preference elicitation challenge by incorporating domain expert knowledge through interactive interfaces~\cite{guarino2009ontoclean,poveda2010oops};

\textbf{LogMap and AgreementMaker:} Ontology alignment systems that handle belief revision during ontology matching and merging, implementing AGM-inspired operators for resolving conflicts between independently developed ontologies~\cite{jimenez2011logmap,cruz2009agreement}.

These implementations validate the synthesis approach advocated throughout our survey: they combine AGM's normative framework, through formal repair semantics, with pre-AGM computational strategies (dependency tracking, incremental processing, modular repair) to achieve both theoretical soundness and practical scalability in real-world knowledge management applications.

The ontology repair domain illustrates how AGM principles generalize beyond propositional logic, while requiring careful adaptation to respect the semantic and computational properties of more expressive logical languages. This extension validates the enduring relevance of AGM's normative framework while demonstrating the continued importance of computational pragmatism in practical implementations.

\subsubsection{Iteration and Dynamics}

The combination of iterated revision theory (through the Darwiche-Pearl postulates) and Dynamic Epistemic Logic provides \emph{operational} models of information change through public announcements, private communications, and complex action models~\cite{baltag1998logic1, ditmarsch2008dynamic}. These frameworks remain consistent with AGM-style minimality principles when restricted to hard factual inputs, thus addressing Doyle's concerns about "change of mind" and "temporal reasoning" through precise temporalized dynamics that maintain normative rationality constraints.

\subsubsection{Complexity and Compilability}

Post-AGM theoretical work systematically mapped the computational complexity landscape for belief revision and update operations, establishing tight bounds including NP-completeness and polynomial hierarchy results, along with parameterized complexity analyses~\cite{eiter1992complexity, liberatore1996complexity}. These results replace Doyle's informal feasibility observations with precise computational characterizations that guide practical implementation strategies while maintaining theoretical soundness.

\subsection{Applications: From Use Cases to Systematic Frameworks}\label{sec:tax_App}

\subsubsection{Explanation, Debugging, and Diagnosis}

Model-based diagnosis and explanation systems align naturally with \emph{abductive} reasoning combined with \emph{belief revision}: generate hypothetical explanations through abduction, revise beliefs to restore consistency with observations, and prefer minimal changes according to distance or sphere-based semantics~\cite{hansson1999textbook}. Kernel methods operationalize the "blame assignment" problems in debugging applications by providing systematic approaches to targeted incision that preserve important beliefs while removing sources of inconsistency.

\subsubsection{Planning, Execution, and Replanning}

The AGM/update distinction proved decisive for planning applications: \emph{execution} and \emph{replanning} from failure employ \emph{update} operations to model world changes, while \emph{sensing} and \emph{communication} trigger \emph{revision} operations to incorporate new information about existing world states. Formal action update approaches, including Possible Models Approach (PMA) and Winslett's updates (\citeyear{winslett1991updating}), along with DEL-style event models~\cite{ditmarsch2008dynamic}, provide unified theoretical accounts that systematically address the planning concerns that motivated much pre-AGM research.

\subsubsection{Knowledge Acquisition and Learning}

Viewing learning through the lens of belief change naturally extends to belief merging frameworks, which offer systematic postulates and operators for handling multiple information sources~\cite{konieczny2002merging}. In the context of knowledge acquisition, however, the focus shifts to base-level operations. Unlike full belief set revision, base contraction preserves the structure of explicit information~\cite{hansson1999textbook}. When combined with provenance tracking~\cite{buneman2001and}, this approach enables transparent editing and conflict resolution mechanisms—such as axiom pinpointing—that allow users to maintain understanding and control over the system's evolution~\cite{kalyanpur2005debugging}.

\subsubsection{Control of Reasoning}

Doyle's notion of "control of reasoning" finds systematic realization through priority and entrenchment policies, relevance constraints (exemplified by Parikh's splitting results (\citeyear{parikh1999beliefs})), and modular approaches that govern \emph{which} beliefs receive protection during change operations and \emph{where} changes should be localized within complex knowledge structures~\cite{hansson1999textbook}.

\subsection{Related Issues: Philosophical Foundations and Logical Extensions}

\subsubsection{Epistemology and Dynamics of Belief}

Post-AGM work systematically developed connections between belief states, conditional logic, and belief dynamics through the integration of conditional logic (KLM framework)~\cite{kraus1990nonmonotonic}, ranking functions~\cite{spohn1988ordinal}, and Dynamic Epistemic Logic~\cite{ditmarsch2008dynamic}. These theoretical frameworks jointly explain how \emph{accepted conditionals} evolve under streams of information, providing rigorous foundations for the epistemological concerns that Doyle identified as related to but distinct from computational belief revision.

Comprehensive treatments by ~\cite{rott2001change} and ~\cite{hansson1999textbook, ferme2018belief} provide book-length accounts that systematically connect philosophical intuitions about rational belief change to postulate-based theoretical frameworks and computational implementation strategies.

\subsubsection{Theory Change Across Logical Systems}

Beyond propositional logic, belief revision has been systematically adapted to \emph{description logics}, \emph{Horn clause} systems, and \emph{ontology} frameworks with appropriate preservation properties and specialized algorithmic procedures including tableau-based and graph-based methods~\cite{flouris2008ontology,flouris2013formal}. These extensions fulfill Doyle's call for "theory evolution" across different logical systems, while maintaining the essential insights of AGM theory through careful adaptation to the structural properties of particular logical languages.

\section{Synthesis and Analysis}\label{sec:synthesis}

We traced the evolution of belief revision research from its computational origins in the 1970s through the theoretical revolution of the AGM framework and the follow-up work. Our analysis reveals several key influences and findings about the relationship between theory and practice in belief revision.

The AGM framework marked a watershed in belief revision. It legitimized the area as a subject of formal inquiry and set normative standards that still shape theory and practice. Its central contribution was theoretical unification. AGM connected heterogeneous pre-AGM proposals under common rationality postulates, which enabled systematic comparison and evaluation across methods~\cite{alchourron1985logic}. It also raised methodological rigor. The framework demanded mathematical precision and proof-level analysis, which lifted baseline quality and supported advanced developments in representation and inference~\cite{gardenfors1988knowledge}. Despite its abstractions, the AGM offered practical guidance. Postulates, construction methods, and representation theorems supplied design templates that link abstract criteria to implementable operators~\cite{alchourron1985logic}. The AGM framework also generated research programs. Work on limitations, extensions, and computable realizations shows the fertility and durability of the approach~\cite{hansson1999survey,ferme2018belief}.

We observed that Pre-AGM and post-AGM practices display complementary virtues that motivate integration. Pre-AGM work foregrounds computational pragmatism: it targets implementable algorithms with explicit time and space complexity guarantees~\cite{nebel1994base}, privileges incremental processing with efficient data structures for streaming or online updates~\cite{doyle1979truth}, and is attentive to practical constraints such as bounded memory and real-time responsiveness in deployed systems~\cite{wassermann1999resource}. This line also favors modular software architectures that separate parsing, representation, update, and evaluation to support substitutability and testing~\cite{doyle1979truth}.

AGM theory contributes to the formal discipline. It supplies explicit rationality postulates that norm revision behavior and make assumptions auditable~\cite{alchourron1985logic}, mathematical precision that enables proof-level analysis and cross-method comparison~\cite{gardenfors1988knowledge}, and a unifying conceptual scheme that relates disparate update procedures under common principles~\cite{alchourron1985logic}. Crucially, representation theorems connect abstract postulates to concrete operators, enabling correctness arguments and systematic design choices~\cite{gardenfors1988knowledge}.

Persistent challenges from the pre-AGM work remain salient. The frame problem endures: efficiently determining inert facts during revision is still costly in dynamic, partially observed settings, where naively recomputing closure is untenable~\cite{mccarthy1981some}. Preference elicitation is underspecified: AGM presupposes entrenchment orderings or selection functions, yet offers little operational guidance for inferring these structures from users, logs, or latent models, especially under noise and strategic behavior~\cite{rott2001revealed,hunter2022brl}. Computational hardness results for AGM-style operators constrain scalability; practical deployments therefore rely on approximations, but the accuracy–efficiency frontier and its robustness under distributional shift are not well characterized~\cite{eiter1992complexity}. Finally, iterated revision remains problematic: postulates adequate for single-shot updates can yield counterintuitive dynamics across sequences, exposing path dependence, non-commutativity, and instability that current axioms only partially address~\cite{boutilier1996iterated}.

\subsection{Theoretical Gaps and Extensions}

Our correspondence analysis reveals important categories in Doyle's taxonomy that do not map directly onto AGM postulates, highlighting areas where the theoretical framework may need extension or where alternative theoretical approaches are required.

The \textbf{Non-Monotonic Inference} category represents a fundamental theoretical gap~\cite{alchourronmakinson1982}. Although systems like Reiter's default logic~\cite{reiter1980logic} and McCarthy's circumscription (\cite{mccarthy1980circumscription}) were central to pre-AGM computational practice, they address problems different from those of AGM-style belief revision. Non-monotonic inference concerns the derivation of tentative conclusions in the absence of complete information, while AGM revision addresses the incorporation of new definitive information into existing belief sets. This distinction has led to parallel theoretical developments, with non-monotonic reasoning finding formalization through approaches like preferential models~\cite{kraus1990nonmonotonic}, expectation-based inference~\cite{gardenfors1991nonmonotonic}, and ranking functions~\cite{spohn1988ordinal} that complement rather than compete with AGM theory.

Similarly, the \textbf{Inexact Inferential Techniques} category highlights the relationship between AGM's categorical approach to belief and probabilistic or fuzzy approaches to uncertainty. While Pearl's probabilistic networks (\citeyear{pearl1986fusion}) provided sophisticated machinery for belief revision under uncertainty, AGM theory deliberately abstracts away from numerical uncertainty to focus on categorical belief change. Subsequent work has explored the connections between these approaches through limiting cases and qualitative probability theory~\cite{boutilier1994conditional}, but the fundamental tension between categorical and graded approaches to belief remains an active area of theoretical investigation.

The \textbf{Applications} category reveals practical concerns that extend beyond AGM's theoretical scope. Real-world applications often require consideration of computational complexity, real-time constraints, user interaction, and domain-specific knowledge that AGM's abstract framework does not directly address. This has led to the development of approximate and anytime algorithms~\cite{chopra2001approximate, wassermann1999resource, williams1997anytime}, user-guided revision procedures, and domain-specific adaptations that maintain AGM's normative insights while accommodating practical constraints.

\subsection{Synthesis: Theoretical Transformation of Doyle's Taxonomy}

The post-AGM evolution of each category in Doyle's taxonomy reveals systematic patterns of theoretical development that transform empirical observations about computational practice into rigorous normative frameworks: 

\textbf{Global Issues} evolved from informal concerns about minimal disruption and competing alternatives into precise theoretical constructs, including the distinction between revision and update, systematic derivations through the Levi and Harper identities, and rigorous approaches to iterated change that transform "minimal disruption" into mathematically precise theorems about rational belief change (Section~\nameref{sec:tax_GI});

\textbf{Representations} progressed from ad hoc data structures and procedural mechanisms to systematic theoretical constructs including belief sets and bases with entrenchment orderings, sphere-based and distance-based semantics, kernel and incision methods, and independence and splitting results that formalize modularity intuitions (see Section \nameref{sec:tax_Rep});

\textbf{Procedures} evolved from operational algorithms and backtracking strategies to \emph{equivalent} semantic constructors, including partial-meet contraction, faithful assignments, and implementable base-level algorithms through kernels and incisions that maintain both theoretical soundness and computational tractability (see Section \nameref{sec:tax_Proc});

\textbf{Non-Monotonic and Inexact Techniques} developed clean theoretical interfaces to default reasoning systems and graded uncertainty measures, with Darwiche-Pearl postulates and KLM frameworks providing systematic unification of conditional dynamics and iterated belief change (see Sections \nameref{sec:tax_NonMono} and \nameref{sec:tax_Inex});

\textbf{Theoretical Issues} achieved comprehensive formalization through representation and complexity theorems, iteration and dynamic logic frameworks, and fragment-aware approaches to belief change that maintain AGM insights while accommodating the structural properties of restricted logical languages (see Section \nameref{sec:tax_TI});

\textbf{Applications} evolved from informal use cases into principled change policies for planning, diagnosis, debugging, ontology evolution, and multi-source information merging that maintain both theoretical soundness and practical effectiveness (see Section \nameref{sec:tax_App});

This systematic theoretical transformation validates Doyle's original taxonomical insights while demonstrating how empirical observation of computational practice can anticipate and guide subsequent theoretical development. The post-AGM evolution reveals that Doyle's taxonomy captured enduring structural features of belief revision research that remain relevant across different levels of theoretical sophistication and computational implementation.

The analysis of taxonomical evolution reveals broader patterns about the relationship between computational practice and theoretical development in belief revision research. These patterns provide the foundation for synthesizing our findings about the field's historical development and identifying opportunities for integrating insights from different eras of research.

\subsection{Methodological Limitations}\label{sec:limitations}

Although our analysis is comprehensive within its defined scope, several important limitations should be acknowledged:

\textbf{Taxonomical Constraints:} Our analysis is organized around Doyle and London's 1980 taxonomy, which may not capture all important aspects of contemporary belief revision research. Areas such as social epistemology, emotional factors in belief change, and certain aspects of multi-agent belief revision may be underrepresented due to the historical focus of the original taxonomy.

\textbf{Implementation Coverage:} While we examine key computational approaches and their theoretical relationships, we do not provide comprehensive coverage of all belief revision implementations or systematic empirical evaluation of their performance characteristics. Our analysis is focused on the works presented in Section \nameref{sec:preAGM}, and Subsections from \nameref{sec:TMS} to \nameref{sec:dependencyBased}, and their direct follow-up works.

\textbf{Theoretical Scope:} Our focus on AGM theory and its extensions means that alternative theoretical frameworks for belief revision may receive less systematic treatment, though we acknowledge their importance and identify connections where relevant.

\textbf{Contemporary Developments:} The rapid pace of development in AI and machine learning means that some contemporary approaches to belief revision may not be fully represented, particularly those emerging from deep learning and neural-symbolic integration.

These limitations are inherent to any survey of a large and active research field, but they do not undermine the validity of our core contributions regarding the historical development and theoretical evolution of belief revision research.

\subsection{Contemporary AI Challenges and Belief Revision}\label{sec:contemporary_challenges}

The rapid advancement of artificial intelligence in the 2020s has introduced new challenges and opportunities for belief revision that extend beyond the traditional scope captured by Doyle's taxonomy and AGM theory. These contemporary developments both validate the historical foundations we have established and reveal new research directions that build upon them.

\subsubsection{Neural-Symbolic Integration}

The emergence of large language models (LLMs) and neural-symbolic AI systems creates novel requirements for belief revision mechanisms that can operate across both symbolic and subsymbolic representations~\cite{garcez2019neural,hamilton2022neural,kautz2020third}. Traditional AGM operators assume symbolic belief sets, but contemporary AI systems must handle:

\textbf{Distributed Representations:} Beliefs encoded in neural network weights require new approaches to belief identification, extraction, and modification that maintain both symbolic interpretability and neural network functionality~\cite{locatello2020object,chen2020concept};

\textbf{Probabilistic Beliefs:} Neural systems naturally represent uncertainty through probability distributions, requiring integration between AGM's categorical approach and probabilistic belief revision frameworks~\cite{belle2015probabilistic,manhaeve2018deepproblog};

\textbf{Learned Preferences:} Unlike AGM's assumption of given entrenchment orderings, neural systems can learn preference structures from data, requiring dynamic adaptation of belief revision policies based on experience~\cite{christiano2017deep,ouyang2022training}.

The progression from Pearl’s probabilistic networks (\citeyear{pearl1986fusion}) to modern neural architectures highlights a persistent tension between statistical inference and logical consistency. This suggests that today's integration challenges are not entirely novel, but rather extensions of the structural concerns that motivated the AGM era, thereby validating the relevance of our historical analysis.

\subsubsection{Distributed and Multi-Agent Systems}

Contemporary AI applications increasingly operate in distributed environments where belief revision must handle~\cite{stone2000multiagent,tampitsikas2019theoretical}:

\textbf{Asynchronous Updates:} Unlike the sequential revision assumed by classical AGM theory, distributed systems must handle concurrent belief changes from multiple sources with potential conflicts and dependencies~\cite{dragoni2003distributed};

\textbf{Trust and Source Reliability:} Modern systems must incorporate dynamic trust models and source credibility assessments that extend beyond AGM's assumption of equally reliable information sources~\cite{marsh1994formalising,sabater2005review};

\textbf{Scalability Constraints:} Cloud-based AI systems must perform belief revision across massive knowledge bases with strict latency requirements, requiring approximation strategies that maintain theoretical soundness~\cite{dean1995mapreduce,zaharia2016apache}.

These challenges extend the multi-context reasoning pioneered by de Kleer's ATMS (\citeyear{dekleer1986assumption}) to contemporary distributed architectures, demonstrating how pre-AGM computational insights remain relevant for modern implementation challenges and highlighting the need for belief revision operators that reconcile concurrency, trust, and scalability while preserving the normative guarantees articulated by AGM theory.

\subsubsection{Human-AI Collaboration}

The integration of AI systems with human decision-making introduces belief revision challenges that bridge computational and cognitive considerations~\cite{amershi2019guidelines,wang2019human}:

\textbf{Explainable Revision:} AI systems must provide transparent explanations of belief changes that humans can understand and validate, requiring revision procedures that maintain audit trails and justification structures reminiscent of Doyle's TMS dependency tracking~\cite{doyle1979truth,miller2019explanation,gunning2019xai};

\textbf{Interactive Preference Elicitation:} Systems must learn human preferences for belief revision through interaction rather than a priori specification, requiring adaptive approaches to entrenchment and selection function learning~\cite{boutilier2002preference};

\textbf{Collaborative Belief Maintenance:} Human-AI teams must maintain shared belief states while accommodating different reasoning styles and confidence levels, requiring extensions to traditional single-agent revision frameworks~\cite{klein2004common,endsley2015autonomous}.

\subsubsection{Implications for Implementation Research}

These contemporary challenges validate several key insights from our historical analysis:

\textbf{Computational Pragmatism Remains Essential:} The efficiency and modularity concerns that motivated pre-AGM systems are amplified in contemporary applications that must operate at scale with real-time constraints~\cite{russell2019human,bahdanau2014neural};

\textbf{Theoretical Foundations Provide Stability:} AGM's normative framework remains relevant for ensuring rational behavior even as implementation contexts evolve dramatically~\cite{pearl2018book,marcus2020next};

\textbf{Synthesis Approaches Are Necessary:} The combination of theoretical rigor with computational efficiency that we identified as a key research direction becomes even more critical for contemporary applications~\cite{lecun2015deep,bengio2021deep}.

Future implementation research must therefore build upon the historical foundations we have established while addressing these contemporary extensions. The systematic approach developed in this survey provides the baseline necessary for principled investigation of these emerging challenges.

\subsection{Concrete Implementation Examples: Pre-AGM Techniques Implementing AGM Operators}

While the correspondence between pre-AGM systems and AGM theory has been established theoretically, concrete examples demonstrate how classical computational techniques can directly implement AGM operators. These examples illustrate the practical synthesis of algorithmic efficiency with theoretical rigor.

\subsubsection{Example 1: TMS-Based Kernel Contraction}

Doyle's Truth Maintenance System can implement AGM kernel contraction through its dependency tracking mechanisms:

\textbf{AGM Kernel Contraction Specification:}
Given belief base $K$ and sentence $\alpha$ to remove, kernel contraction $K \div \alpha$ requires:
\begin{enumerate}
\item Identify all minimal subsets $K'$ of $K$ such that $K' \vdash \alpha$
\item Apply an incision function $\sigma$ to select beliefs to remove from each kernel
\item Return $K \setminus \bigcup \sigma(K \amalg \alpha)$
\end{enumerate}

\textbf{TMS Implementation Strategy:}
\begin{enumerate}
\item \textbf{Kernel Identification:} Use TMS dependency graphs to identify minimal inconsistent subsets
\begin{itemize}
\item Traverse justification networks to find all derivation paths for $\alpha$
\item \textbf{Note:} TMS derivation paths are not necessarily minimal in the kernel sense. Additional processing is required: (1) identify all TMS derivation paths for $\alpha$; (2) extract the sets of assumptions used in each path; (3) apply minimality filtering to obtain actual kernels.
\item This additional step adds computational overhead but leverages TMS's existing dependency structure
\end{itemize}

\item \textbf{Incision Function Implementation:} Adapt TMS mechanisms to implement incision functions
\begin{itemize}
\item \textbf{Important distinction:} TMS justifications represent inferential relationships, not belief preferences per se
\item To implement AGM incision functions, we must: (1) derive belief preferences from justification structures (e.g., beliefs supported by more/stronger justifications are more entrenched); (2) use explicit user-provided or learned preference orderings over beliefs; or (3) employ domain-specific heuristics
\item The TMS literature~\cite{doyle1979truth} did not systematically address preference elicitation. This remains an integration challenge when adapting TMS for AGM-compliant kernel contraction
\item TMS provides \emph{mechanisms} that can be adapted to implement incision functions, including assumption retraction strategies and dependency-directed conflict resolution, though the principles governing these were largely heuristic in original implementations
\end{itemize}
\item \textbf{Efficient Update:} Leverage TMS incremental update mechanisms
\begin{itemize}
\item Remove selected beliefs using existing TMS node deletion procedures
\item Propagate changes through the dependency network using standard TMS algorithms
\item Maintain consistency through established truth maintenance protocols
\end{itemize}
\end{enumerate}
\textbf{AGM Compliance Note:} The construction described here targets kernel \emph{contraction}. When embedded in a revision operator via the Levi identity, AGM Success is guaranteed provided (i) the input sentence is consistent, and (ii) the incision function is non-empty; otherwise, the construction corresponds to semi-revision or external revision variants.

\textbf{Algorithmic Advantages:}
\begin{itemize}
\item \textit{Incremental Processing:} TMS naturally handles incremental belief changes
\item \textit{Dependency Tracking:} Explicit representation enables efficient kernel identification  
\item \textit{Conflict Resolution:} Existing TMS mechanisms provide mechanistic support \emph{for implementing} incision functions
\item \textit{Computational Efficiency:} Avoids expensive theorem proving through cached dependencies
\end{itemize}

\subsubsection{Example 2: ATMS-Based Partial Meet Revision}

The Assumption-based Truth Maintenance System naturally implements partial meet revision through its context management capabilities:

\textbf{AGM Partial Meet Revision:}
For belief set $K$ and new information $\alpha$:
\begin{enumerate}
\item Compute maximal subsets of $K \cup \{\alpha\}$ that remain consistent
\item Apply selection function to choose preferred maximal subsets
\item Return intersection of selected subsets
\end{enumerate}

\textbf{ATMS Implementation:}
\begin{enumerate}
\item \textbf{Context Enumeration:} Use ATMS assumption management for maximal subset computation
\begin{itemize}
\item Treat original beliefs and new information $\alpha$ as assumptions
\item ATMS automatically computes all consistent assumption combinations
\item Each consistent context corresponds to a maximal consistent subset
\end{itemize}

\item \textbf{Selection Function:} Implement preference-based context selection
\begin{itemize}
\item Rank contexts by assumption costs or preference weights
\item Apply domain-specific selection criteria (e.g., minimize change, maximize informativeness)
\item Use ATMS focus-of-attention mechanisms to manage computational complexity
\end{itemize}

\item \textbf{Intersection Computation:} Derive beliefs common to selected contexts
\begin{itemize}
\item Identify nodes that are IN across all selected contexts
\item Use ATMS label intersection operations for efficient computation
\item Result forms the revised belief base satisfying AGM postulates
\end{itemize}
\end{enumerate}

\textbf{Implementation Benefits:}
\begin{itemize}
\item \textit{Multiple Context Management:} ATMS naturally handles alternative belief states
\item \textit{Efficient Consistency Checking:} Built-in nogood management prevents inconsistent contexts
\item \textit{Scalable Context Selection:} Focus mechanisms manage exponential context spaces
\item \textit{Parallel Processing:} Multiple contexts can be evaluated simultaneously
\end{itemize}

\subsubsection{Example 3: Probabilistic Network Belief Update as AGM Revision}

Pearl's belief networks can implement AGM revision when beliefs are represented as high-confidence probabilistic assertions:

\textbf{Probabilistic-to-Logical Mapping:}
\begin{itemize}
\item Represent beliefs as propositions with probability $> \theta$ (e.g., $\theta = 0.9$)
\item New evidence corresponds to setting specific node probabilities to 1.0
\item Belief revision corresponds to probabilistic updating followed by thresholding
\end{itemize}

\textbf{Implementation Process:}
\begin{enumerate}
\item \textbf{Evidence Incorporation:} Set evidence nodes to reflect new information
\item \textbf{Probabilistic Update:} Use belief propagation algorithms to update network
\item \textbf{Belief Extraction:} Extract beliefs exceeding confidence threshold
\item \textbf{Consistency Checking:} Verify result satisfies AGM postulates
\end{enumerate}

\textbf{AGM Postulate Satisfaction:}
\begin{itemize}
\item \textit{Success:} New evidence always included (probability = 1.0)
\item \textit{Inclusion:} Probabilistic update never adds unrelated beliefs
\item \textit{Consistency:} Network structure prevents inconsistent high-confidence beliefs
\item \textit{Minimal Change:} Probabilistic updating minimizes information-theoretic distance
\end{itemize}

\subsubsection{Empirical Performance Considerations}

\textbf{Important caveat:} Our examples demonstrate \emph{structural correspondence} between pre-AGM techniques and AGM operators, not empirical superiority over contemporary approaches. Whether TMS/ATMS dependency tracking outperforms modern SAT-based approaches depends on multiple factors:

\begin{itemize}
\item \textbf{Problem structure:} TMS/ATMS may excel in domains with sparse, hierarchical dependencies; SAT solvers may dominate in densely connected problems.
\item \textbf{Update patterns:} Incremental scenarios with frequent small updates may favor TMS/ATMS; batch scenarios requiring complete consistency checking may favor SAT compilation.
\item \textbf{Implementation quality:} Modern SAT solvers benefit from decades of optimization; comparable effort has not been invested in TMS/ATMS implementations.
\item \textbf{Parallelization:} We are not aware of well-documented parallel implementations of TMS/ATMS, though the multi-context architecture suggests natural parallelization opportunities.
\end{itemize}

Systematic empirical comparison—including benchmarking TMS/ATMS against contemporary SAT solvers, evaluating parallel implementations, and testing on real-world problem instances—represents important future work enabled by our foundational analysis but beyond the scope of this survey. Such empirical investigation is explicitly identified as a priority research direction in Section~\ref{sec:discussion}.

\subsubsection{Synthesis Observations}

These examples demonstrate several key principles for implementing AGM operators using pre-AGM techniques:

\textbf{Structural Correspondence:} Pre-AGM data structures (dependency graphs, assumption sets, probability networks) naturally map to AGM constructs (kernels, maximal subsets, preference orderings).

\textbf{Algorithmic Efficiency:} Pre-AGM algorithms provide computational shortcuts for operations that would be intractable if implemented naively from AGM definitions.

\textbf{Incremental Processing:} All three examples leverage incremental update mechanisms that avoid recomputation from scratch.

\textbf{Preference Integration:} Each approach provides natural mechanisms for incorporating domain-specific preferences required by AGM theory.

\textbf{Practical Scalability:} The examples show how theoretical AGM operators can be implemented efficiently enough for real-world applications.

These concrete implementations bridge the gap between AGM's theoretical elegance and the computational pragmatism of pre-AGM systems, demonstrating that the historical evolution of belief revision represents not just theoretical progress but also practical synthesis.

\section{Discussion}\label{sec:discussion}

The field of belief revision has evolved dramatically since Doyle and London's foundational bibliography in 1980. What began as a collection of ad hoc computational techniques has developed into a sophisticated field with rigorous theoretical foundations and practical applications across numerous domains.

The journey from pre-AGM computational approaches through the theoretical revolution of AGM to contemporary implementations reveals both the power of formal analysis and the importance of computational pragmatism. Neither pure theory nor pure implementation is sufficient; the most successful approaches combine theoretical insights with computational innovation. Examples of such approaches include:

\begin{enumerate}
\item \textbf{Ontology debugging and repair systems:} Tools for debugging unsatisfiable classes in OWL ontologies compute minimal justifications (MUPS) and support semi-automatic repair, effectively instantiating kernel-style contraction at the ontology level while remaining usable on large real-world ontologies. \cite{kalyanpur2005debugging,kalyanpur2006repairing}

\item \textbf{Belief merging systems:} Implementations that combine AGM-inspired postulates with distance-based optimization and SAT compilation demonstrate practical scalability for multi-source information integration, showing how abstract rationality constraints can guide concrete algorithm design.~\cite{konieczny2002merging}

\item \textbf{Ontology repair via kernel contraction:} Recent prototypes compute ontology repairs using kernel (and partial meet) pseudo-contraction operators, explicitly realizing AGM-style belief base change in description logics and exposing it through Protégé plug-ins.~\cite{matos2022repairing}

\end{enumerate}

These systems demonstrate that the synthesis of pre-AGM computational efficiency with AGM theoretical rigor is not merely conceptual but has been achieved in deployed applications.

This review is deliberately bound. It traces the line from pre-AGM computational practice, through the AGM framework, to contemporary implementations of AGM-style operators. It does not survey the entire landscape of belief revision.

Within this scope, three findings hold. First, pre-AGM work delivered implementable, modular methods with explicit computational costs and effective dependency tracking. Second, AGM provided unifying rationality postulates, representation theorems, and a comparative language that raised methodological standards. Third, contemporary implementations operationalize selected AGM-style operators under practical constraints: they rely on incremental algorithms and approximations to address intractability; preference elicitation remains largely ad hoc; iterated revision and the frame problem continue to be open engineering concerns.

A feasible synthesis follows from these findings: use AGM as specification and pre-AGM as implementation, with learned preferences and resource-aware execution. This is a design hypothesis, not a claim of field-wide convergence.

\subsection{Research Contributions}

This targeted narrative review advances belief revision research through several interconnected contributions that establish a foundation for systematic implementation analysis:

\textbf{Comprehensive Historical Analysis:} We have provided the first systematic examination of belief revision evolution using Doyle and London's taxonomy as an organizing framework, revealing how computational and theoretical approaches have influenced each other across four decades of research. This analysis establishes a comprehensive baseline that connects early algorithmic insights with contemporary theoretical understanding.

\textbf{Systematic Theory-Practice Integration:} Our approach systematically identifies correspondence patterns between pre-AGM computational methods and AGM theoretical constructs, providing concrete foundations for implementations that combine efficiency with theoretical soundness. This integration goes beyond previous surveys by establishing specific mappings between algorithmic approaches and formal requirements.

\textbf{Implementation Research Baseline:} By comprehensively analyzing both historical precedents and theoretical developments, we have established the foundation necessary for systematic computational implementation research. This baseline enables systematic gap analysis, architectural design, and empirical validation of belief revision systems against both efficiency and correctness criteria.

Contemporary implementations that demonstrate a synthesis of belief revision principles with computational pragmatism are still relatively rare. Among notable examples are general-purpose belief change libraries such as Tweety, which implement AGM-style belief (base) revision and contraction operators, and ontology debugging/repair tools built on top of OWL reasoners~\cite{thimm2014tweety,kalyanpur2005debugging,matos2022repairing}. In contrast, widely used OWL reasoners (Pellet, HermiT) and ontology alignment systems (LogMap, AgreementMaker) focus on efficient reasoning and matching, and are only indirectly connected to AGM-style belief change through their use in debugging, repair, and integration workflows. \cite{sirin2007pellet,shearer2008hermit,jimenez2011logmap,cruz2009agreement}

\textbf{Strategic Research Framework:} The comprehensive foundation provided here enables a systematic research program addressing computational belief revision implementation. This framework supports systematic investigation of engineering challenges while maintaining theoretical rigor and historical perspective.

Our emphasis on computational implementations and algorithmic details provides practical guidance for implementers while connecting implementation challenges to theoretical insights. This focus addresses a gap in the literature between abstract theory and concrete implementation by providing the comprehensive foundation necessary for systematic engineering research.

\subsection{Toward Computational Implementation: Foundation and Research Directions}

The historical analysis presented in this review establishes a comprehensive foundation for systematic computational implementation research. By tracing the evolution from Doyle and London's taxonomy through AGM theory to contemporary approaches, we have identified key theoretical foundations, persistent implementation challenges, and synthesis opportunities that inform future engineering efforts.

\subsubsection{Foundational Requirements for Implementation}

Our analysis reveals several foundational requirements that any robust computational belief revision system must address:

\textbf{Theoretical Baseline:} The AGM framework provides essential normative criteria that implementations should approximate or satisfy under specified conditions~\cite{alchourron1985logic}. However, our historical analysis shows that pre-AGM computational insights regarding efficiency, incrementality, and modularity remain crucial for practical systems~\cite{doyle1979truth}.

\textbf{Representational Adequacy:} The evolution from simple dependency networks~\cite{doyle1979truth} to sophisticated graph-based structures~\cite{williams1986doing} and contemporary probabilistic representations~\cite{darwiche2009modeling} demonstrates the need for representation schemes that support both efficient computation and theoretically sound revision operations.

\textbf{Algorithmic Foundations:} The progression from chronological backtracking~\cite{stallman1977forward} through AGM construction methods \citep{gardenfors1988knowledge} to modern SAT-based approaches~\cite{eiter2009answer} establishes algorithmic patterns that balance computational tractability with theoretical guarantees.

\textbf{Integration Architecture:} The historical development from isolated TMS components~\cite{doyle1979truth} to integrated agent architectures~\cite{forbus1993building} reveals architectural principles essential for embedding belief revision within broader intelligent systems.

\subsubsection{Research Directions Enabled by Historical Foundation}

The comprehensive baseline established by this survey enables several systematic research directions:

\textbf{Computational Blueprint Development:} Future research can systematically address the engineering challenges of translating the theoretical foundations identified here into robust computational frameworks. This includes investigating knowledge representation choices, reasoning service requirements, and operator implementation strategies that preserve theoretical guarantees while achieving computational efficiency comparable to pre-AGM systems.

\textbf{Implementation Analysis Framework:} The historical taxonomy and theoretical baseline provided here support systematic evaluation methodologies for belief revision implementations. Research can develop assessment frameworks that measure implementations against both the efficiency benchmarks established by pre-AGM systems and the correctness criteria provided by AGM theory.

\textbf{Synthesis Architecture Investigation:} The correspondence patterns identified between pre-AGM computational approaches and AGM theoretical constructs suggest concrete opportunities for hybrid implementations. Systematic investigation of these synthesis opportunities can yield architectures that combine AGM rigor with pre-AGM computational efficiency.

\textbf{Gap Analysis and Validation:} The comprehensive coverage provided by this survey enables systematic gap analysis between theoretical requirements and current implementation capabilities. This foundation supports the development of validation protocols and empirical studies that can guide implementation priorities.

\subsubsection{Implementation Research Methodology}

The historical perspective developed here suggests a systematic methodology for implementation research:

\textbf{Historical Precedent Analysis:} Each implementation challenge should be analyzed against the historical precedents identified in this survey, leveraging both successful pre-AGM computational strategies and post-AGM theoretical insights.

\textbf{Theoretical Constraint Identification:} Implementation research should systematically identify which AGM postulates and theoretical requirements are essential for specific applications and which can be relaxed for computational efficiency.

\textbf{Synthesis Pattern Recognition:} The correspondence analysis developed here provides templates for identifying how theoretical constructs can be realized through computational mechanisms derived from historical precedents.

\textbf{Empirical Validation Design:} Implementation research should leverage the comprehensive baseline provided here to design empirical studies that assess both correctness (against theoretical standards) and efficiency (against historical benchmarks).

This foundation enables a systematic research program that can address the engineering challenges of computational belief revision while maintaining theoretical rigor and historical perspective. The next phase of research can build on this comprehensive baseline to develop practical, theoretically sound, and computationally efficient belief revision systems for contemporary AI applications.

\subsection{Future Research Directions}

The comprehensive foundation established in this survey opens several strategic research directions that build systematically on the historical and theoretical baseline we have developed.

\subsubsection{Engineering-Focused Implementation Research}

\textbf{Empirical Validation Design:} Implementation research should leverage the comprehensive baseline provided here to design empirical studies that assess both correctness (against theoretical standards) and efficiency (against historical benchmarks).

A critical gap identified by this survey is the lack of systematic empirical comparison between AGM implementations, pre-AGM techniques, and hybrid approaches. Implementation research should leverage the comprehensive baseline provided here to design empirical studies that assess:

\begin{itemize}
\item \textbf{Correctness:} Compliance with AGM postulates under various problem conditions
\item \textbf{Efficiency:} Runtime and memory performance against historical benchmarks (TMS, ATMS) and contemporary baselines (SAT solvers, ASP systems)
\item \textbf{Scalability:} Performance degradation patterns as problem size increases
\item \textbf{Robustness:} Behavior under noisy inputs, incomplete information, and adversarial scenarios
\end{itemize}

Such empirical work would validate (or refute) our hypothesis that pre-AGM computational insights can improve contemporary AGM implementations, transforming our foundational analysis into actionable engineering guidance.

\textbf{Computational Blueprint Development:} Building on the theoretical foundations and historical precedents identified here, future work can systematically address the engineering challenges of developing robust computational belief revision frameworks. This includes investigating optimal knowledge representation choices, designing efficient reasoning services, and developing operator implementation strategies that maintain theoretical soundness while achieving practical performance.

\textbf{Architecture and Integration Studies:} The architectural insights derived from our historical analysis can inform systematic investigation of belief revision integration patterns within broader AI systems. Research can examine how different architectural approaches affect both computational efficiency and theoretical compliance across various application domains.

\textbf{Performance Analysis and Optimization:} The baseline established by our survey enables systematic performance analysis that evaluates implementations against both historical efficiency benchmarks and contemporary scalability requirements. This research direction can develop principled approaches to trading off theoretical guarantees for computational performance.

\subsubsection{Theoretical Extensions and Validation}

\textbf{Synthesis Framework Development:} The correspondence patterns identified between pre-AGM and AGM approaches suggest opportunities for developing unified theoretical frameworks that formalize the integration of computational efficiency with theoretical rigor. Future research can systematically explore these opportunities.

\textbf{Empirical Validation Methodologies:} The comprehensive coverage provided here enables the development of systematic empirical validation approaches for belief revision systems. Research can establish standardized evaluation frameworks that assess both correctness and practical performance across diverse application contexts.

\textbf{Contemporary Challenge Analysis:} Future work can leverage the historical perspective developed here to analyze how contemporary challenges—including machine learning integration, distributed systems, and human-AI collaboration—relate to historical precedents and theoretical requirements.

\subsubsection{Methodological Innovations}

\textbf{Systematic Implementation Analysis:} The foundation provided by this survey supports the development of systematic methodologies for analyzing and comparing belief revision implementations. Future research can establish frameworks for gap analysis, requirement specification, and design validation.

\textbf{Historical-Theoretical Integration:} The approach demonstrated here—systematically connecting historical computational insights with theoretical developments—can be extended to other areas of AI research where similar theory-practice integration challenges exist.

\textbf{Interdisciplinary Synthesis:} Future work can build on the interdisciplinary foundation established here to develop more sophisticated connections between computational belief revision, cognitive science research, and philosophical epistemology.

\subsubsection{Broader Research Program}

To generalize beyond AGM-adjacent algorithms, the program should: (i) conduct a comprehensive systematic review that includes non-AGM paradigms (e.g., ranking-based and probabilistic models, dynamic-epistemic and database-update logics, belief merging/fusion, argumentation-based and logic-programming updates, multi-agent and social-choice approaches); (ii) define a cross-paradigm taxonomy of operators, inputs, and outputs with common test oracles; (iii) establish reproducible benchmarks, datasets, and evaluation protocols that test accuracy, stability under iterated revision, preference-elicitation fidelity, and cost–quality trade-offs; (iv) map complexity and approximation frontiers across paradigms; and (v) develop integration criteria that specify when heterogeneous operators can be composed while preserving stated invariants. This extension would permit the proposed synthesis to be validated and refined on the scale of the entire belief-revision area.

The research directions identified here leverage the comprehensive baseline established by this survey while maintaining strategic focus on implementation challenges that require systematic investigation. These directions collectively support the development of belief revision systems that are both theoretically principled and practically deployable in contemporary AI applications.

This survey has established a comprehensive historical and theoretical foundation that enables systematic research into the implementation of computational belief revision. The baseline provided here—connecting pre-AGM computational insights with AGM theoretical requirements—supports the next phase of research: developing engineering-focused computational blueprints, implementation analysis frameworks, and empirical validation methodologies. Future work can build systematically on this foundation to create belief revision systems that are both theoretically principled and practically deployable in contemporary AI applications.

\section{Acknowledgments}
This work was partially supported by FCT - Fundação para a Ciência e a Tecnologia, Portugal, through the project PRODY: PTDC/CCI-COM/4464/2020; by NOVA LINCS ref. UIDB/04516/2020 (\href{https://doi.org/10.54499/UIDB/04516/2020}{https://doi.org/10.54499/UIDB/04516/2020}) and ref. UIDP/04516/2020 (\href{https://doi.org/10.54499/UIDP/04516/2020}{https://doi.org/10.54499/UIDP/04516/2020}) with the financial support of FCT. IP. YA was supported by FCT through SFRH/BD/138911/2018. The authors gratefully acknowledge Eduardo Fermé for his guidance in structuring and organizing the manuscript, as well as for his revisions.


\clearpage 
\bibliographystyle{SageH}
\bibliography{references.bib}

@incollection{alchourronmakinson1981,
  author    = {Alchourr{\'o}n, Carlos E. and Makinson, David},
  title     = {Hierarchies of regulations and their logic},
  booktitle = {New Studies in Deontic Logic: Norms, Actions, and the Foundations of Ethics},
  editor    = {Hilpinen, Risto},
  pages     = {125--148},
  year      = {1981},
  publisher = {D. Reidel},
  address   = {Dordrecht},
}

@article{alchourronmakinson1982,
  author  = {Alchourr{\'o}n, Carlos E. and Makinson, David},
  title   = {On the logic of theory change: Contraction functions and their associated revision functions},
  journal = {Theoria},
  volume  = {48},
  number  = {1},
  pages   = {14--37},
  year    = {1982},
}

@article{alchourron1985logic,
  author           = {Alchourr{\'o}n, Carlos E. and G{\"a}rdenfors, Peter and Makinson, David},
  journal          = {Journal of Symbolic Logic},
  number           = {2},
  pages            = {510--530},
  title            = {On the Logic of Theory Change: Partial Meet Contraction and Revision Functions},
  volume           = {50},
  year             = {1985}
}

@article{alchourron1985safe,
  title={On the logic of theory change: Safe contraction},
  author={Alchourr{\'o}n, Carlos E and Makinson, David},
  journal={Studia logica},
  volume={44},
  number={4},
  pages={405--422},
  year={1985},
  publisher={Springer}
}

@article{alchourron1986maps,
  author           = {Alchourr{\'o}n, Carlos E. and Makinson, David},
  journal          = {Studia Logica},
  pages            = {187--198},
  title            = {Maps Between Some Differents Kinds of Contraction Functions: The finite Case},
  volume           = {45},
  year             = {1986}
}

@inproceedings{alechina2006resource,
  author           = {Alechina, Natasha and Jago, Mark and Logan, Brian},
  booktitle        = {Declarative Agent Languages and Technologies III, Third
InternationalWorkshop, DALT 2005, Utrecht, The Netherlands, July 25,
2005, Selected and Revised Papers},
  editor           = {Matteo Baldoni and Ulle Endriss and Andrea Omicini and Paolo Torroni},
  pages            = {141-154},
  publisher        = {Springer},
  series           = {Lecture Notes in Computer Science},
  title            = {Resource-Bounded Belief Revision and Contraction},
  volume           = {3904},
  year             = {2006}
}

@incollection{alej2009belief,
  author           = {Falappa, Marcelo Alejandro and Kern-Isberner, Gabriele and Simari, Guillermo Ricardo},
  booktitle        = {Argumentation in Artificial Intelligence},
  editor           = {Simari, Guillermo and Rahwan, Iyad},
  pages            = {341-360},
  publisher        = {Springer US},
  title            = {Belief Revision and Argumentation Theory},
  year             = {2009}
}

@article{alex1985language,
  author           = {Borgida, Alexander},
  journal          = {ACM Trans. Database Syst.},
  number           = {4},
  pages            = {565-603},
  title            = {Language Features for Flexible Handling of Exceptions in
Information Systems},
  volume           = {10},
  year             = {1985}
}

@article{allen1983,
  author  = {Allen, James F.},
  title   = {Maintaining knowledge about temporal intervals},
  journal = {Communications of the ACM},
  year    = {1983},
  volume  = {26},
  number  = {11},
  pages   = {832--843},
  doi     = {10.1145/182.358434},
}

@article{amershi2014power,
  author           = {Amershi, Saleema and Cakmak, Maya and Knox, W Bradley and Kulesza, Todd},
  journal          = {AI Magazine},
  number           = {4},
  pages            = {105--120},
  publisher        = {AAAI},
  title            = {Power to the people: The role of humans in interactive machine learning},
  volume           = {35},
  year             = {2014}
}

@article{amershi2019guidelines,
  author           = {Amershi, Saleema and Weld, Dan and Vorvoreanu, Mihaela and Fourney, Adam and Nushi, Besmira and Collisson, Penny and Suh, Jina and Iqbal, Shamsi and Bennett, Paul N and Inkpen, Kori and others},
  journal          = {Proceedings of the 2019 CHI Conference on Human Factors in Computing Systems},
  pages            = {1--13},
  title            = {Guidelines for human-AI interaction},
  year             = {2019}
}

@incollection{areces2001iterable,
  author           = {Areces, Carlos and Becher, Ver\'onica},
  booktitle        = {Frontiers in Belief Revision},
  editor           = {H.Rott and M-A Williams},
  pages            = {261--277},
  publisher        = {Kluwer Academic Publishers},
  series           = {Applied Logic Series},
  title            = {Iterable {A}{G}{M} functions},
  year             = {2001}
}

@article{arlocosta2007knowing,
  author           = {Arl{\'o}-Costa, Horacio and Bicchieri, Cristina},
  journal          = {Studia logica},
  number           = {3},
  pages            = {353--373},
  publisher        = {Springer},
  title            = {Knowing and supposing in games of perfect information},
  volume           = {86},
  year             = {2007}
}

@book{baader2003description,
  author           = {Baader, Franz and Calvanese, Diego and McGuinness, Deborah and Nardi, Daniele and Patel-Schneider, Peter F},
  publisher        = {Cambridge University Press},
  title            = {The description logic handbook: Theory, implementation and applications},
  year             = {2003}
}

@article{bahdanau2014neural,
  author           = {Bahdanau, Dzmitry and Cho, Kyunghyun and Bengio, Yoshua},
  journal          = {arXiv preprint arXiv:1409.0473},
  title            = {Neural machine translation by jointly learning to align and translate},
  year             = {2014}
}

@article{baltag1998logic1,
  author           = {Baltag, Alexandru and Moss, Lawrence S and Solecki, Sławomir},
  journal          = {Proceedings of the 7th Conference on Theoretical Aspects of Rationality and Knowledge},
  pages            = {43--56},
  publisher        = {Morgan Kaufmann},
  title            = {The logic of public announcements, common knowledge, and private suspicions},
  year             = {1998}
}

@article{baltag2008qualitative,
  author           = {Baltag, Alexandru and Smets, Sonja},
  journal          = {Logic and the Foundations of Game and Decision Theory},
  pages            = {9--58},
  publisher        = {Amsterdam University Press},
  title            = {A qualitative theory of dynamic interactive belief revision},
  volume           = {3},
  year             = {2008}
}

@inproceedings{belle2015probabilistic,
  author           = {Belle, Vaishak and Passerini, Andrea and Van Den Broeck, Guy},
  booktitle        = {Proceedings of the 24th International Conference on Artificial Intelligence},
  isbn             = {9781577357384},
  location         = {Buenos Aires, Argentina},
  numpages         = {7},
  pages            = {2770–2776},
  publisher        = {AAAI Press},
  series           = {IJCAI'15},
  title            = {Probabilistic inference in hybrid domains by weighted model integration},
  year             = {2015}
}

@article{bengio2013representation,
  author           = {Bengio, Yoshua and Courville, Aaron and Vincent, Pascal},
  journal          = {IEEE transactions on pattern analysis and machine intelligence},
  number           = {8},
  pages            = {1798--1828},
  publisher        = {IEEE},
  title            = {Representation learning: A review and new perspectives},
  volume           = {35},
  year             = {2013}
}

@article{bengio2021deep,
  author           = {Bengio, Yoshua and Lecun, Yann and Hinton, Geoffrey},
  journal          = {Communications of the ACM},
  number           = {7},
  pages            = {58--65},
  publisher        = {ACM},
  title            = {Deep learning for AI},
  volume           = {64},
  year             = {2021}
}

@article{booth2006admissible,
  address          = {El Segundo, CA, USA},
  author           = {Booth, Richard and Meyer, Thomas},
  issn             = {1076-9757},
  issue_date       = {May 2006},
  journal          = {Journal of Artificial Intelligence Research},
  month            = {June},
  number           = {1},
  numpages         = {25},
  pages            = {127-151},
  publisher        = {AI Access Foundation},
  title            = {Admissible and Restrained Revision},
  volume           = {26},
  year             = {2006}
}

@article{booth2010double,
  author           = {Booth, Richard and Chopra, Samir and Meyer, Thomas and Ghose, Aditya},
  journal          = {Artificial Intelligence},
  pages            = {1339--1368},
  title            = {Double preference relations for generalised belief change},
  volume           = {174:16-17},
  year             = {2010}
}

@inproceedings{booth2012credibility,
  author           = {Booth, Richard and Ferm\'e, Eduardo and Konieczny, S\'ebastien and Pino P\'erez, Ram\'on},
  booktitle        = {13th International Conference on the Principles of Knowledge Representation and Reasoning, KR 2012},
  pages            = {116-125},
  title            = {Credibility-Limited Revision Operators in Propositional Logic},
  year             = {2012}
}

@inproceedings{boutilier1993revision,
  author           = {Boutilier, Craig},
  booktitle        = {Proc. 13th Int. Joint Conf. on Artificial Intelligence (IJCAI'93)},
  pages            = {519--525},
  title            = {Revision sequences and nested conditionals},
  year             = {1993}
}

@article{boutilier1994conditional,
  author           = {Boutilier, Craig},
  journal          = {Artificial Intelligence},
  number           = {1},
  pages            = {87--154},
  publisher        = {Elsevier},
  title            = {Conditional logics of normality: a modal approach},
  volume           = {68},
  year             = {1994}
}

@article{boutilier1996iterated,
  author           = {Boutilier, Craig},
  doi              = {10.1007/BF00248150},
  journal          = {Journal of Philosophical Logic},
  number           = {3},
  pages            = {263--305},
  publisher        = {Springer},
  title            = {Iterated revision and minimal change of conditional beliefs},
  volume           = {25},
  year             = {1996}
}

@article{boutilier2002preference,
  author           = {Boutilier, Craig},
  journal          = {AAAI/IAAI},
  pages            = {239--246},
  title            = {A POMDP formulation of preference elicitation problems},
  year             = {2002}
}

@inproceedings{brewka1991belief,
  address          = {Berlin},
  author           = {Brewka, Gerhard},
  booktitle        = {The Logic of Theory Change},
  editor           = {Fuhrmann and Morreau},
  pages            = {206--222},
  publisher        = {Springer-Verlag},
  title            = {Belief revision in a framework for default reasoning},
  year             = {1991}
}

@inproceedings{buneman2001and,
  title={Why and where: A characterization of data provenance},
  author={Buneman, Peter and Khanna, Sanjeev and Tan, Wang-Chiew},
  booktitle={International conference on database theory},
  pages={316--330},
  year={2001}
}

@incollection{castelfranchi1997representatio,
  author           = {Castelfranchi, Cristiano},
  booktitle        = {Human \& Machine Perception: Information Fusion},
  editor           = {Cantoni and Di Ges\'u and Setti and Tegolo},
  pages            = {235-254},
  publisher        = {Plenum Press},
  title            = {Representation and integration of multiple knowledge sources: issue and questions},
  year             = {1997}
}

@article{chen2020concept,
  author           = {Koh, Pang Wei and Nguyen, Thao and Tang, Yew Siang and Mussmann, Stephen and Pierson, Emma and Kim, Been and Liang, Percy},
  journal          = {International Conference on Machine Learning},
  organization     = {PMLR},
  pages            = {5338--5348},
  title            = {Concept bottleneck models},
  year             = {2020}
}

@ARTICLE{chopra2001approximate,
  author={Chopra, S and Parikh, R and Wassermann, R},
  journal={Logic Journal of the IGPL}, 
  title={Approximate belief revision}, 
  year={2001},
  volume={9},
  number={6},
  pages={755-768},
  doi={10.1093/jigpal/9.6.755}}

@article{christiano2017deep,
  author           = {Christiano, Paul F and Leike, Jan and Brown, Tom and Martic, Miljan and Legg, Shane and Amodei, Dario},
  journal          = {Advances in Neural Information Processing Systems},
  title            = {Deep reinforcement learning from human preferences},
  volume           = {30},
  year             = {2017}
}

@article{cohen1990intention,
  author           = {Cohen, Philip R. and Levesque, Hector J.},
  bibsource        = {dblp computer science bibliography, http://dblp.org},
  doi              = {10.1016/0004-3702(90)90055-5},
  journal          = {Artif. Intell.},
  number           = {2-3},
  pages            = {213--261},
  publisher        = {Elsevier},
  title            = {Intention is Choice with Commitment},
  url              = {https://doi.org/10.1016/0004-3702(90)90055-5},
  volume           = {42},
  year             = {1990}
}

@incollection{cross1992conditionals,
  author           = {Cross, Charles and Thomason, Richmond},
  booktitle        = {Belief Revision},
  editor           = {Peter G\"{a}rdenfors},
  number           = {29},
  pages            = {247-275},
  publisher        = {Cambridge University Press},
  series           = {Cambridge Tracts in Theoretical Computer Science},
  title            = {Conditionals and Knowledge-Base Update},
  year             = {1992}
}

@inproceedings{cruz2009agreement,
  author           = {Cruz, Isabel F and Antonelli, Flavio Palandri and Stroe, Cosmin},
  booktitle        = {Proceedings of the VLDB Endowment},  
  pages            = {1586--1589},
  publisher        = {VLDB Endowment},
  title            = {AgreementMaker: Efficient matching for large real-world schemas and ontologies},
  volume           = {2},
  year             = {2009}
}

@inproceedings{dalal1988investigations,
  address          = {St. Paul},
  author           = {Dalal, Mukesh},
  booktitle        = {Seventh National Conference on Artificial Intelligence, (AAAI-88)},
  pages            = {475--479},
  title            = {Investigations Into a Theory of Knowledge Base Revision: Preliminary Report},
  year             = {1988}
}

@article{darwiche1997logic,
  author           = {Darwiche, Adnan and Pearl, Judea},
  doi              = {10.1016/S0004-3702(96)00038-0},
  journal          = {Artificial Intelligence},
  number           = {1-2},
  pages            = {1--29},
  publisher        = {Elsevier},
  title            = {On the logic of iterated belief revision},
  volume           = {89},
  year             = {1997}
}

@book{darwiche2009modeling,
  address          = {Cambridge, UK},
  author           = {Darwiche, Adnan},
  isbn             = {978-0521884389},
  publisher        = {Cambridge University Press},
  title            = {Modeling and reasoning with {B}ayesian networks},
  year             = {2009}
}

@book{davis1982knowledge,
  author           = {Davis, R and Lenat, {DB}},
  title            = {{Knowledge-Based} Systems in Artificial Intelligence: 2 Case Studies},
  year             = {1982},
  publisher        = {McGraw-Hill, Inc.}
}

@article{dean1989model,
  author           = {Dean, Thomas and Kanazawa, Keiji},
  doi              = {10.1111/j.1467-8640.1989.tb00324.x},
  journal          = {Computational Intelligence},
  number           = {3},
  pages            = {142--150},
  publisher        = {Wiley},
  title            = {A model for reasoning about persistence and causation},
  volume           = {5},
  year             = {1989}
}

@article{dean1995mapreduce,
  author           = {Dean, Jeffrey and Ghemawat, Sanjay},
  journal          = {Communications of the ACM},
  number           = {1},
  pages            = {107--113},
  publisher        = {ACM},
  title            = {MapReduce: Simplified data processing on large clusters},
  volume           = {51},
  year             = {2008}
}

@article{dekleer1986assumption,
  author           = {De Kleer, Johan},
  journal          = {Artificial intelligence},
  number           = {2},
  pages            = {127--162},
  publisher        = {Elsevier},
  title            = {An assumption-based TMS},
  volume           = {28},
  year             = {1986}
}

@article{dekleer1987diagnosing,
  title={Diagnosing multiple faults},
  author={De Kleer, Johan and Williams, Brian C},
  journal={Artificial Intelligence},
  volume={32},
  number={1},
  pages={97--130},
  year={1987},
  publisher={Elsevier},
  doi={10.1016/0004-3702(87)90050-X}
}

@article{delgrande2013approach,
  address          = {New York, NY, USA},
  articleno        = {14},
  author           = {Delgrande, James and Schaub, Torsten and Tompits, Hans and Woltran, Stefan},
  doi              = {10.1145/2480759.2480766},
  issn             = {1529-3785},
  issue_date       = {June 2013},
  journal          = {ACM Trans. Comput. Logic},
  month            = {June},
  number           = {2},
  numpages         = {46},
  publisher        = {Association for Computing Machinery},
  title            = {A Model-Theoretic Approach to Belief Change in Answer Set Programming},
  url              = {https://doi.org/10.1145/2480759.2480766},
  volume           = {14},
  year             = {2013}
}

@book{ditmarsch2008dynamic,
  author           = {Van Ditmarsch, Hans and van Der Hoek, Wiebe and Kooi, Barteld},
  publisher        = {Springer},
  series           = {Synthese Library},
  title            = {Dynamic Epistemic Logic},
  volume           = {337},
  year             = {2008}
}

@mastersthesis{doyle1978truth,
  author           = {Doyle, Jon},
  title            = {Truth Maintenance Systems for Problem Solving},
  school           = {Massachusetts Institute of Technology},
  type             = {Master's thesis},
  address          = {Cambridge, MA},
  year             = {1978},
  month            = {August},
  note             = {Department of Electrical Engineering and Computer Science}
}

@incollection{doyle1979truth,
  author           = {Doyle, Jon},
  booktitle        = {Readings in Artificial Intelligence},
  doi              = {https://doi.org/10.1016/B978-0-934613-03-3.50039-8},
  editor           = {Bonnie Lynn Webber and Nils J. Nilsson},
  isbn             = {978-0-934613-03-3},
  journal          = {Artificial Intelligence},
  pages            = {496-516},
  publisher        = {Morgan Kaufmann},
  title            = {A Truth Maintenance System},
  url              = {https://www.sciencedirect.com/science/article/pii/B9780934613033500398},
  volume           = {12},
  year             = {1979}
}

@techreport{doyle1980model,
  author            = {Doyle, Jon},
  title             = {A Model for Deliberation, Action, And Introspection},
  year              = {1980},
  institution       = {Massachusetts Institute of Technology},
  address           = {USA}
}

@article{doyle1980selected,
  author           = {Doyle, Jon and London, Philip},
  title            = {A selected descriptor-indexed bibliography to the literature on belief revision},
  year             = {1980},
  issue_date       = {April 1980},
  publisher        = {Association for Computing Machinery},
  address          = {New York, NY, USA},
  number           = {71},
  volume           = {71},
  issn             = {0163-5719},
  url              = {https://doi.org/10.1145/1056441.1056442},
  doi              = {10.1145/1056441.1056442},
  journal          = {ACM SIGART Bulletin},
  month            = apr,
  pages            = {7–22},
  numpages         = {16}
}

@inproceedings{doyle1988knowledge,
  author           = {Doyle, Jon},
  booktitle        = {Proceedings of the 1988 Workshop on Computational Models of Scientific Discovery and Theory Formation},
  pages            = {1--15},
  title            = {Knowledge, representation, and rational self-government},
  year             = {1988}
}

@article{dragoni1997belief,
  address          = {USA},
  author           = {Dragoni, Aldo Franco},
  doi              = {10.1017/S026988899700204X},
  issn             = {0269-8889},
  issue_date       = {June 1997},
  journal          = {The Knowledge Engineering Review},
  month            = {June},
  number           = {2},
  numpages         = {33},
  pages            = {147--179},
  publisher        = {Cambridge University Press},
  title            = {Belief Revision: From Theory to Practice},
  url              = {https://doi.org/10.1017/S026988899700204X},
  volume           = {12},
  year             = {1994}
}

@article{dragoni2003distributed,
  author           = {Dragoni, Aldo Franco and Giorgini, Paolo},
  journal          = {Autonomous agents and multi-agent systems},
  number           = {2},
  pages            = {115--143},
  publisher        = {Springer},
  title            = {Distributed belief revision},
  volume           = {6},
  year             = {2003}
}

@article{dubois1993belief,
  author           = {Dubois, Didier and Prade, Henri},
  journal          = {Proceedings of the 13th International Joint Conference on Artificial Intelligence},
  pages            = {620--625},
  publisher        = {Morgan Kaufmann},
  title            = {Belief revision and updates in numerical formalisms: an overview, with new results for the possibilistic framework},
  year             = {1993}
}

@article{eiter1992complexity,
  author           = {Eiter, Thomas and Gottlob, Georg},
  journal          = {Artificial Intelligence},
  number           = {2--3},
  pages            = {227--270},
  title            = {On the Complexity of Propositional Knowledge Base Revision, Updates, and Counterfactuals},
  volume           = {57},
  year             = {1992}
}

@incollection{eiter2009answer,
  author           = {Eiter, Thomas and Ianni, Giovambattista and Krennwallner, Thomas},
  booktitle        = {Reasoning Web International Summer School},
  pages            = {40--110},
  publisher        = {Springer},
  title            = {Answer set programming: A primer},
  year             = {2009}
}

@article{endsley2015autonomous,
  author           = {Endsley, Mica R},
  journal          = {Air University Press},
  title            = {Autonomous horizons: System autonomy in the air force},
  year             = {2015}
}

@inproceedings{fagin1983semantics,
  author           = {Fagin, Ronald and Ullman, Jeffrey and Vardi, Moshe},
  booktitle        = {Proceedings of Second ACM SIGACT-SIGMOD Symposium on
Principles of Database Systems},
  pages            = {352--365},
  title            = {On the Semantics of Updates in Databases},
  year             = {1983}
}

@article{falappa2002belief,
  author           = {Falappa, Marcelo and Kern-Isberner, Gabriele and Simari, Guillermo R.},
  journal          = {Artificial Intelligence},
  pages            = {1- 28},
  title            = {Explanations, belief revision and defeasible reasoning},
  volume           = {141},
  year             = {2002}
}

@inproceedings{falappa2006logic,
  author           = {Falappa, Marcelo and Ferm\'e, Eduardo and Kern-Isberner, Gabriele},
  booktitle        = {Proceedings 17th European Conference on Artificial Intelligence, ECAI06},
  editor           = {Gerhard Brewka and Silvia Coradeschi and Anna Perini and Paolo Traverso},
  pages            = {402-406},
  title            = {On the Logic of Theory Change: Relations between Incision and Selection Functions},
  year             = {2006}
}

@article{falappa2013stratified,
  author           = {Falappa, Marcelo Alejandro and Garc{\'\i}a, Alejandro Javier and Kern-Isberner, Gabriele and Simari, Guillermo Ricardo},
  journal          = {Journal of Philosophical Logic},
  number           = {1},
  pages            = {161--193},
  publisher        = {Springer},
  title            = {Stratified Belief Bases Revision with Argumentative Inference},
  volume           = {42},
  year             = {2013}
}

@article{ferme1992actualizaci,
  author           = {Ferm\'{e}, Eduardo},
  journal          = {Iberamia 92},
  pages            = {419--436},
  title            = {Actualizaci\'{o}n de Bases de Conocimiento usando Teor\'{i}as de Cambio de Creencia},
  year             = {1992}
}

@article{ferme1998logic,
  author           = {Ferm\'{e}, Eduardo},
  journal          = {Journal of Logic, Language and Information},
  pages            = {127--137},
  title            = {On the logic of Theory Change: Contraction Without Recovery},
  volume           = {7},
  year             = {1998}
}

@article{ferme1998semi,
  author           = {Ferm\'{e}, Eduardo and Rodr\'{\i}guez, Ricardo},
  journal          = {Notre Dome Journal of Formal Logic},
  number           = {3},
  pages            = {332--345},
  title            = {Semi-Contraction: Axioms and Construction},
  volume           = {39},
  year             = {1998}
}

@article{ferme1999selective,
  author           = {Ferm\'{e}, Eduardo and Hansson, Sven Ove},
  journal          = {Studia Logica},
  pages            = {331--342},
  title            = {Selective Revision},
  volume           = {63:3},
  year             = {1999}
}

@article{ferme2000irrevocable,
  author           = {Ferm\'{e}, Eduardo},
  journal          = {Logic Journal of the IGPL},
  number           = {5},
  pages            = {645--652},
  title            = {Irrevocable Belief Revision and Epistemic Entrenchment},
  volume           = {8},
  year             = {2000}
}

@incollection{ferme2001shielded,
  author           = {Ferm\'{e}, Eduardo and Hansson, Sven Ove},
  booktitle        = {Frontiers in Belief Revision},
  editor           = {H.Rott and M-A Williams},
  pages            = {85-107},
  publisher        = {Kluwer Academic Publishers},
  series           = {Applied Logic Series},
  title            = {Shielded Contraction},
  year             = {2001}
}

@article{ferme2003credibility,
  author           = {Ferm\'{e}, Eduardo and Mikalef, Juan and Taboada, Jorge},
  journal          = {Journal of Logic and Computation},
  pages            = {99--110},
  title            = {Credibility-limited Functions for Belief Bases},
  volume           = {13:1},
  year             = {2003}
}

@article{ferme2008axiomatic,
  author           = {Ferm\'e, Eduardo and Krevneris, Mart\'in and Reis, Maur\'icio},
  journal          = {Journal of Logic and Computation},
  number           = {5},
  pages            = {739--753},
  title            = {An Axiomatic Characterization of Ensconcement-Based Contraction},
  volume           = {18},
  year             = {2008}
}

@book{ferme2018belief,
  author           = {Ferm\'{e}, Eduardo and Hansson, Sven Ove},
  publisher        = {Springer},
  title            = {Belief Change: Introduction and Overview},
  year             = {2018}
}

@book{FGR25,
  author           = {Ferm\'e, Eduardo and Garapa , Marco and Reis, Maur\'icio D. L.},
  isbn             = {978-3-031-97021-4},
  publisher        = {Springer Nature},
  series           = {Artificial Intelligence: Foundations, Theory, and Algorithms Series},
  title            = {Non-prioritized belief change},
  year             = {2025}
}

@article{fikes1971strips,
  author           = {Fikes, Richard E and Nilsson, Nils J},
  journal          = {Artificial intelligence},
  title            = {{STRIPS:} A new approach to the application of theorem proving to problem solving},
  year             = {1971},
  volume           = {2},
  number           = {3-4},
  pages            = {189--208},
  publisher        = {Elsevier}
}

@article{flouris2005handling,
  title={Handling ontology change: Survey and proposal for a future research direction},
  author={Flouris, Giorgos and Plexousakis, Dimitris},
  journal={Institute of Computer Science, Forth. Greece, Technical Report TR-362 FORTH-ICS},
  year={2005}
}

@inproceedings{flouris2006inconsistencies,
  title={Inconsistencies, negations and changes in ontologies},
  author={Flouris, Giorgos and Huang, Zhisheng and Pan, Jeff Z and Plexousakis, Dimitris and Wache, Holger},
  booktitle={AAAI},
  volume={21},
  number={2},
  pages={1295--1300},
  year={2006}
}

@inproceedings{flouris2006bridging,
  title={Bridging ontology evolution and belief change},
  author={Flouris, Giorgos and Plexousakis, Dimitris},
  booktitle={Hellenic Conference on Artificial Intelligence},
  pages={486--489},
  year={2006},
  organization={Springer}
}

@article{flouris2006thesis,
author = {Flouris, Giorgos},
title = {On belief change in ontology evolution: Thesis},
year = {2006},
issue_date = {December 2006},
publisher = {IOS Press},
address = {NLD},
volume = {19},
number = {4},
issn = {0921-7126},
journal = {AI Commun.},
month = dec,
pages = {395–397},
numpages = {3}
}

@inproceedings{konstantinidis2007rdf,
  title={On RDF/S ontology evolution},
  author={Konstantinidis, George and Flouris, Giorgos and Antoniou, Grigoris and Christophides, Vassilis},
  booktitle={Workshop on Ontologies-based techniques for DataBases and Information Systems},
  pages={21--42},
  year={2007},
  organization={Springer}
}

@article{flouris2008ontology,
  author           = {Flouris, Giorgos and Plexousakis, Dimitris and Antoniou, Grigoris},
  booktitle        = {International Semantic Web Conference},
  doi              = {10.1017/S0269888908001367},
  journal          = {The Knowledge Engineering Review},
  number           = {2},
  pages            = {117--152},
  publisher        = {Cambridge University Press},
  title            = {Ontology change: classification and survey},
  volume           = {23},
  year             = {2008}
}

@article{flouris2013formal,
  title={Formal foundations for RDF/S KB evolution},
  author={Flouris, Giorgos and Konstantinidis, George and Antoniou, Grigoris and Christophides, Vassilis},
  journal={Knowledge and information systems},
  volume={35},
  number={1},
  pages={153--191},
  year={2013},
  publisher={Springer}
}

@book{forbus1993building,
  address          = {Cambridge, MA},
  author           = {Forbus, Kenneth D. and De Kleer, Johan},
  booktitle        = {Artificial Intelligence Series},
  isbn             = {978-0262061575},
  publisher        = {MIT Press},
  title            = {Building problem solvers},
  year             = {1993}
}

@article{fuhrmann1991theory,
  author           = {Fuhrmann, Andr\'{e}},
  journal          = {Journal of Philosophical Logic},
  pages            = {175--203},
  title            = {Theory Contraction through Base Contraction},
  volume           = {20},
  year             = {1991}
}

@article{friedman1999belief,
 ISSN = {09258531, 15729583},
 URL = {http://www.jstor.org/stable/40180151},
 author = {Friedman, Nir and Halpern, Joseph Y.},
 journal = {Journal of Logic, Language, and Information},
 number = {4},
 pages = {401--420},
 publisher = {Springer},
 title = {Belief Revision: A Critique},
 urldate = {2026-01-18},
 volume = {8},
 year = {1999}
}

@book{fuhrmann1988relevant,
  title={Relevant logics, modal logics and theory change},
  author={Fuhrmann, Andr{\'e}},
  year={1988},
  publisher={The Australian National University (Australia)}
}

@article{fuhrmann1991,
  author    = {Fuhrmann, Andr{\'e}},
  title     = {Theory contraction through base contraction},
  journal   = {Journal of Philosophical Logic},
  year      = {1991},
  volume    = {20},
  number    = {2},
  pages     = {175--203},
  doi       = {10.1007/BF00284974},
  url       = {https://doi.org/10.1007/BF00284974},
  issn      = {1573-0433}
}

@book{fuhrmann1996essay,
  address          = {Stanford},
  author           = {Fuhrmann, Andr\'{e}},
  publisher        = {CSLI Publications},
  series           = {Studies in Logic, Language and Information},
  title            = {An Essay on Contraction},
  year             = {1996}
}

@article{garcez2019neural,
  author           = {Garcez, Artur d'Avila and Lamb, Luis C and Gabbay, Dov M},
  journal          = {arXiv preprint arXiv:1905.06088},
  title            = {Neural-symbolic computing: An effective methodology for principled integration of machine learning and reasoning},
  year             = {2019}
}

@inbook{gardenfordsrott95,
  address          = {USA},
  author           = {G\"{a}rdenfors, Peter and Rott, Hans},
  booktitle        = {Handbook of Logic in Artificial Intelligence and Logic Programming (Vol. 4): Epistemic and Temporal Reasoning},
  isbn             = {0198537913},  
  pages            = {514–534},
  publisher        = {Oxford University Press, Inc.},
  chapter          = {5},
  title            = {Belief revision},
  year             = {1995}
}

@inproceedings{gardenfors1982rules,
  author           = {G{\"a}rdenfors, Peter},
  booktitle        = {Philosophical Essays dedicated to {L}ennart {\.A}qvist on his fiftieth birthday},
  editor           = {Tom Pauli},
  number           = {34},
  pages            = {88--101},
  series           = {Philosophical Studies},
  title            = {Rules for Rational Changes of Belief},
  year             = {1982}
}

@book{gardenfors1988knowledge,
  address          = {Cambridge},
  author           = {G{\"a}rdenfors, Peter},
  publisher        = {The MIT Press},
  title            = {Knowledge in Flux: Modeling the Dynamics of Epistemic States},
  year             = {1988}
}

@inproceedings{gardenfors1988revisions,
  address          = {Los Altos},
  author           = {G\"{a}rdenfors, Peter and Makinson, David},
  booktitle        = {Proceedings of the Second Conference on Theoretical Aspects of Reasoning About Knowledge},
  editor           = {Moshe Y. Vardi},
  journal          = {Proceedings of TARK},
  pages            = {83--95},
  publisher        = {Morgan Kaufmann},
  title            = {Revisions of Knowledge Systems using Epistemic Entrenchment},
  year             = {1988}
}

@InProceedings{gardenfors1991nonmonotonic,
author="G{\"a}rdenfors, Peter",
editor="van Eijck, J.",
title="Belief revision and nonmonotonic logic: Two sides of the same coin?",
booktitle="Logics in AI",
year="1991",
publisher="Springer Berlin Heidelberg",
address="Berlin, Heidelberg",
pages="52--54",
isbn="978-3-540-46982-7"
}

@inproceedings{genesereth1981dart,
author = {Genesereth, Michael and Bennett, James S. and Hollander, Clifford R.},
title = {DART: Expert systems for automated computer fault diagnosis},
year = {1981},
isbn = {0897910494},
publisher = {Association for Computing Machinery},
address = {New York, NY, USA},
url = {https://doi.org/10.1145/800175.809887},
doi = {10.1145/800175.809887},
booktitle = {Proceedings of the ACM '81 Conference},
pages = {246},
series = {ACM '81}
}

@article{goldzsmidt1993maximum,
author = {Goldzsmidt, M. and Morris, P. and Pearl, J.},
title = {A Maximum Entropy Approach to Nonmonotonic Reasoning},
year = {1993},
issue_date = {March 1993},
publisher = {IEEE Computer Society},
address = {USA},
volume = {15},
number = {3},
issn = {0162-8828},
url = {https://doi.org/10.1109/34.204904},
doi = {10.1109/34.204904},
month = mar,
pages = {220–232},
numpages = {13}
}

@article{grove1988modelings,
  author           = {Grove, Adam},
  journal          = {Journal of Philosophical Logic},
  number           = {2},
  pages            = {157--170},
  publisher        = {Springer},
  title            = {Two Modelings for Theory Change},
  volume           = {17},
  year             = {1988}
}

@inproceedings{guarino2009ontoclean,
  author           = {Guarino, Nicola and Welty, Christopher A},
  booktitle        = {Communications of the ACM},  
  pages            = {61--65},
  publisher        = {ACM},
  title            = {OntoClean: Evaluating ontological decisions with OntoClean},
  volume           = {45},
  year             = {2002}
}

@article{gunning2019xai,
  author           = {Gunning, David and Stefik, Mark and Choi, Jaesik and Miller, Timothy and Stumpf, Simone and Yang, Guang-Zhong},
  journal          = {Science Robotics},
  number           = {37},
  pages            = {eaay7120},
  publisher        = {American Association for the Advancement of Science},
  title            = {XAI—Explainable artificial intelligence},
  volume           = {4},
  year             = {2019}
}

@article{halpern2005causes,
  author           = {Halpern, Joseph Y and Pearl, Judea},
  doi              = {10.1093/bjps/axi147},
  journal          = {The British Journal for the Philosophy of Science},
  number           = {4},
  pages            = {843--887},
  publisher        = {Oxford University Press},
  title            = {Causes and explanations: A structural-model approach. Part I: Causes},
  volume           = {56},
  year             = {2005}
}

@article{hamilton2022neural,
  author           = {Hamilton, William L and Ying, Rex and Leskovec, Jure},
  journal          = {Advances in Neural Information Processing Systems},
  title            = {Inductive representation learning on large graphs},
  volume           = {30},
  year             = {2017}
}

@article{hansson1989new,
  title={New operators for theory change},
  author={Hansson, Svenove},
  journal={Theoria},
  volume={55},
  number={2},
  pages={114--132},
  year={1989},
  publisher={Wiley Online Library}
}

@phdthesis{hansson1991thesis,
  title={Belief Base Dynamics},
  author={Hansson, Sven Ove},
  year={1991},
  school={Uppsala University, Department of Philosophy},
  address={Uppsala, Sweden},
  type={PhD thesis}
}

@article{hansson1991belief,
  author           = {Hansson, Sven Ove},
  journal          = {Studia logica},
  issn             = {00393215, 15728730},
  url              = {http://www.jstor.org/stable/20015577},
  number           = {2},
  pages            = {251--260},
  publisher        = {Springer},
  title            = {Belief contraction without recovery},
  volume           = {50},
  year             = {1991}
}

@article{hansson1992defense,
  author           = {Hansson, Sven Ove},
  journal          = {Synthese},
  pages            = {239--245},
  title            = {In defense of Base Contraction},
  volume           = {91},
  year             = {1992}
}

@article{hansson1992defenseb,
  author           = {Hansson, Sven Ove},
  journal          = {The Journal of Philosophy},
  pages            = {522--540},
  title            = {In defense of the {R}amsey test},
  volume           = {89},
  year             = {1992}
}

@article{hansson1993theory,
  author           = {Hansson, Sven Ove},
  journal          = {Journal of Symbolic Logic},
  pages            = {602--625},
  title            = {Theory Contraction and Base Contraction Unified},
  volume           = {58},
  year             = {1993}
}

@article{hansson1994kernel,
  author           = {Hansson, Sven Ove},
  doi              = {10.2307/2275912},
  journal          = {Journal of Symbolic Logic},
  number           = {3},
  pages            = {845--859},
  publisher        = {Cambridge University Press},
  title            = {Kernel Contraction},
  volume           = {59},
  year             = {1994}
}

@incollection{hansson1994taking,
  author           = {Hansson, Sven Ove},
  booktitle        = {Logic and Philosophy of Science in Uppsala},
  editor           = {Prawitz and Westerst{\r{a}}hl},
  pages            = {13--28},
  publisher        = {Kluwer Academic Publishers, Dordrecht},
  title            = {Taking belief bases seriously},
  year             = {1994}
}

@incollection{hansson1996hidden,
  author    = {Hansson, Sven Ove},
  editor    = {Fuhrmann, Andr{\'e} and Rott, Hans},
  title     = {Hidden Structures of Belief},
  booktitle = {Logic, Action and Information: Essays on Logic in Philosophy and Artificial Intelligence},
  publisher = {De Gruyter},
  address   = {Berlin, Boston},
  year      = {1996},
  pages     = {79--100},
  doi       = {10.1515/9783110880014.79}
}

@article{hansson1996knowledge,
  author           = {Hansson, Sven Ove},
  journal          = {Artificial Intelligence},
  pages            = {215--235},
  title            = {Knowledge Level Analysis of Belief Base Operation},
  volume           = {82},
  year             = {1996}
}

@article{hansson1996test,
  author           = {Hansson, Sven Ove},
  journal          = {Artificial Intelligence},
  pages            = {341--352},
  title            = {A test battery for rational database updating},
  volume           = {82},
  year             = {1996}
}

@article{hansson1997semi,
  author           = {Hansson, Sven Ove},
  journal          = {Journal of Applied Non-Classical Logic},
  number           = {1-2},
  pages            = {151--175},
  title            = {Semi-Revision},
  volume           = {7},
  year             = {1997}
}

@article{hansson1997what,
  author           = {Hansson, Sven Ove},
  journal          = {Theoria},
  note             = {Editor's Introduction.},
  pages            = {1--13},
  title            = {What's new isn't always best},
  volume           = {63},
  year             = {1997}
}

@article{hansson1999recovery,
  author           = {Hansson, Sven Ove},
  doi              = {10.1023/A:1008316915066},
  issn             = {0925-8531},
  journal          = {Journal of Logic, Language and Information},
  language         = {English},
  number           = {4},
  pages            = {421-428},
  publisher        = {Kluwer Academic Publishers},
  title            = {Recovery and Epistemic Residue},
  volume           = {8},
  year             = {1999}
}

@article{hansson1999survey,
  author           = {Hansson, Sven Ove},
  journal          = {Erkenntnis},
  pages            = {413--427},
  title            = {A survey of Non-Prioritized Belief Revision},
  volume           = {50},
  year             = {1999}
}

@book{hansson1999textbook,
  address          = {Dordrecht},
  author           = {Hansson, Sven Ove},
  isbn             = {978-0-7923-5327-8},
  publisher        = {Kluwer Academic Publishers},
  series           = {Applied Logic Series},
  title            = {A Textbook of Belief Dynamics. Theory Change and Database Updating.},
  year             = {1999}
}

@article{hansson2001credibility,
  author           = {Hansson, Sven Ove and Ferm\'{e}, Eduardo and Cantwell, John and Falappa, Marcelo},
  journal          = {Journal of Symbolic Logic},
  number           = {4},
  pages            = {1581-1596},
  title            = {Credibility-Limited Revision},
  volume           = {66},
  year             = {2001}
}

@article{hansson2002local,
  author           = {Hansson, Sven Ove and Wassermann, Renata},
  journal          = {Studia Logica},
  pages            = {49-76},
  title            = {Local Change},
  volume           = {70 (1)},
  year             = {2002}
}

@article{hansson2013outcome,
  author           = {Hansson, Sven Ove},
  journal          = {The Review of Symbolic Logic},
  number           = {2},
  pages            = {183--204},
  publisher        = {Cambridge University Press},
  title            = {Outcome level analysis of belief contraction},
  volume           = {6},
  year             = {2013}
}

@article{harman1986change,
  address          = {Cambridge, MA},
  author           = {Harman, Gilbert},
  doi              = {10.1207/s15516709cog1001_4},
  isbn             = {978-0-262-08153-5},
  journal          = {Cognitive Science},
  number           = {1},
  pages            = {63--81},
  publisher        = {Wiley},
  title            = {Change in View: Principles of Reasoning},
  volume           = {10},
  year             = {1986}
}

@article{harper1976rational,
  author           = {Harper, William L.},
  booktitle        = {PSA: Proceedings of the Biennial Meeting of the Philosophy of Science Association, Volume Two: Symposia and Invited Papers},
  editor           = {The University of Chicago Press},
  journal          = {PSA: Proceedings of the Biennial Meeting of the Philosophy of Science Association},
  pages            = {462--494},
  publisher        = {University of Chicago Press},
  title            = {Rational Conceptual Change},
  volume           = {1976},
  year             = {1976}
}

@techreport{hewitt1972planner,
  author      = {Hewitt, Carl},
  title       = {Description and theoretical analysis (using schemata) of {PLANNER}: a language for proving theorems and manipulating models in a robot},
  institution = {Massachusetts Institute of Technology (MIT), Artificial Intelligence Lab},
  year        = {1972},
  number      = {AI-TR-258},
  address     = {Cambridge, Mass.},
  month       = {April},
  note        = {Ph.D. Thesis}
}

@article{hollingsworth2022towards,
  author           = {Hollingsworth, Andrew},
  doi              = {10.1515/nzsth-2022-0011},
  issue            = {2},
  journal          = {Neue Zeitschrift für Systematische Theologie und Religionsphilosophie},
  pages            = {207 - 228},
  publication_type = {article},
  title            = {Towards a Doctrinal Pragmaticism: Charles S. Peirce and the Nature of Doctrine},
  volume           = {64},
  year             = {2022}
}

@book{horridge2011justification,
  title={Justification based explanation in ontologies},
  author={Horridge, Matthew},
  year={2011},
  publisher={The University of Manchester (United Kingdom)}
}

@inproceedings{hunter2022brl,
  author           = {Hunter, Aaron and Boyarinov, Konstantin},
  booktitle        = {ICAART (3)},
  pages            = {753--756},
  title            = {BRL: A Toolkit for Learning How an Agent Performs Belief Revision.},
  year             = {2022}
}

@article{james1896will,
  author           = {James, William},
  journal          = {The New World},
  note             = {Early articulation of ideas later expanded in the book},
  number           = {1},
  pages            = {327--347},
  publisher        = {University of Chicago Press},
  title            = {The Will to Believe},
  volume           = {5},
  year             = {1896}
}

@book{jeffrey1983logic,
  address          = {Chicago},
  author           = {Jeffrey, Richard C},
  edition          = {2nd},
  isbn             = {978-0-226-39582-6},
  publisher        = {University of Chicago Press},
  title            = {The Logic of Decision},
  year             = {1983}
}

@article{jimenez2011logmap,
  author           = {Jim{\'e}nez-Ruiz, Ernesto and Grau, Bernardo Cuenca},
  journal          = {International Semantic Web Conference},
  pages            = {273--288},
  publisher        = {Springer},
  title            = {LogMap: Logic-based and scalable ontology matching},
  year             = {2011}
}

@inproceedings{katsuno1991difference,
  address          = {San Francisco, CA, USA},
  author           = {Katsuno, Hirofumi and Mendelzon, Alberto O.},
  booktitle        = {Proceedings of the Second International Conference on Principles of Knowledge Representation and Reasoning},
  isbn             = {1558601651},
  location         = {Cambridge, MA, USA},
  numpages         = {8},
  pages            = {387–394},
  publisher        = {Morgan Kaufmann Publishers Inc.},
  series           = {KR'91},
  title            = {On the difference between updating a knowledge base and revising it},
  year             = {1991}
}

@article{kalyanpur2005debugging,
title = {Debugging unsatisfiable classes in OWL ontologies},
journal = {Journal of Web Semantics},
volume = {3},
number = {4},
pages = {268-293},
year = {2005},
note = {World Wide Web Conference 2005------Semantic Web Track},
issn = {1570-8268},
doi = {https://doi.org/10.1016/j.websem.2005.09.005},
url = {https://www.sciencedirect.com/science/article/pii/S1570826805000260},
author = {Kalyanpur, Aditya and Parsia, Bijan and Sirin, Evren and Hendler, James}
}

@inproceedings{kalyanpur2006repairing,
  title={Repairing unsatisfiable concepts in OWL ontologies},
  author={Kalyanpur, Aditya and Parsia, Bijan and Sirin, Evren and Cuenca-Grau, Bernardo},
  booktitle={European Semantic Web Conference},
  pages={170--184},
  year={2006},
  organization={Springer}
}

@article{katsuno1991propositional,
  author           = {Katsuno, Hirofumi and Mendelzon, Alberto O.},
  doi              = {10.1016/0004-3702(91)90069-V},
  journal          = {Artificial Intelligence},
  number           = {3},
  pages            = {263--294},
  publisher        = {Elsevier},
  title            = {Propositional Knowledge Base Revision and Minimal Change},
  volume           = {52},
  year             = {1991}
}

@article{kautz2020third,
  author           = {Kautz, Henry},
  journal          = {AI Magazine},
  number           = {1},
  pages            = {93--104},
  title            = {The third AI summer: AAAI Robert S. Engelmore memorial award lecture},
  volume           = {43},
  year             = {2022}
}

@article{klein2004common,
  author           = {Klein, Gary and Feltovich, Paul J and Bradshaw, Jeffrey M and Woods, David D},
  journal          = {Organizational Simulation},
  pages            = {139--184},
  publisher        = {Wiley Online Library},
  title            = {Common ground and coordination in joint activity},
  volume           = {53},
  year             = {2005}
}

@article{konieczny2000framework,
  author           = {Konieczny, S{\'e}bastien and Pino P{\'e}rez, Ram{\'o}n},
  journal          = {Journal of Applied Non-Classical Logics},
  number           = {3-4},
  pages            = {339-367},
  title            = {A framework for iterated revision},
  volume           = {10},
  year             = {2000}
}

@article{konieczny2002merging,
  author           = {Konieczny, S{\'e}bastien and Pino P{\'e}rez, Ram{\'o}n},
  doi              = {10.1093/logcom/12.5.773},
  journal          = {Journal of Logic and Computation},
  number           = {5},
  pages            = {773--808},
  publisher        = {Oxford University Press},
  title            = {Merging information under constraints: a logical framework},
  volume           = {12},
  year             = {2002}
}

@inproceedings{konieczny2008improvement,
  author           = {Konieczny, S{\'e}bastien and Pino P{\'e}rez, Ram{\'o}n},
  booktitle        = {Eleventh International Conference on
Principles of Knowledge Representation and Reasoning},
  journal          = {KR},
  pages            = {177-187},
  title            = {Improvement Operators},
  volume           = {8},
  year             = {2008}
}

@article{konieczny2011logic,
  author           = {Konieczny, S{\'e}bastien and Pino P{\'e}rez, Ram{\'o}n},
  journal          = {Journal of Philosophical Logic},
  number           = {2},
  pages            = {239--270},
  publisher        = {Springer},
  title            = {Logic based merging},
  volume           = {40},
  year             = {2011}
}

@inproceedings{KR2024-9,
  author           = {Baader, Franz and Wassermann, Renata},
  booktitle        = {{Proceedings of the 21st International Conference on Principles of Knowledge Representation and Reasoning}},
  doi              = {10.24963/kr.2024/9},
  month            = {8},
  pages            = {94--105},
  title            = {{Contractions Based on Optimal Repairs}},
  url              = {https://doi.org/10.24963/kr.2024/9},
  year             = {2024}
}

@book{krantz1971foundations,
title = {Foundations of Measurement, Volume I: Additive and Polynomial Representations},
author = {Krantz, David H. and Luce, R. Duncan and Suppes, Patrick and Tversky, Amos},
year = {1971},
publisher = {Academic Press},
address = {New York},
note = {Reprinted by Dover Publications, 2006},
}

@article{kraus1990nonmonotonic,
  author           = {Kraus, Sarit and Lehmann, Daniel and Magidor, Menachem},
  journal          = {Artificial Intelligence},
  number           = {1},
  pages            = {167-207},
  title            = {Nonmonotonic Reasoning, Preferential Models and Cumulative Logics},
  volume           = {44},
  year             = {1990}
}

@article{lecun2015deep,
  author           = {LeCun, Yann and Bengio, Yoshua and Hinton, Geoffrey},
  journal          = {Nature},
  number           = {7553},
  pages            = {436--444},
  publisher        = {Nature Publishing Group},
  title            = {Deep learning},
  volume           = {521},
  year             = {2015}
}

@inproceedings{lehmann1995belief,
  author           = {Lehmann, Daniel J.},
  booktitle        = {Proceedings of the Fourteenth International Joint Conference on Artificial Intelligence (IJCAI'95)},
  pages            = {1534-1540},
  title            = {Belief Revision, Revised},
  year             = {1995}
}

@inbook{levi1978subjunctives,
	title        = {Subjunctives, Dispositions and Chances},
	author       = {Levi, Isaac},
	year         = 1978,
	booktitle    = {Dispositions},
	publisher    = {Springer Netherlands},
	address      = {Dordrecht},
	pages        = {303--335},
	doi          = {10.1007/978-94-017-1282-8_18},
	isbn         = {978-94-017-1282-8},
	url          = {https://doi.org/10.1007/978-94-017-1282-8\%5F18},
	editor       = {Tuomela, Raimo}
}

@book{levi1991fixation,
  address          = {Cambridge},
  author           = {Levi, Isaac},
  publisher        = {Cambridge University Press},
  title            = {The fixation of belief and its undoing: changing beliefs through inquiry},
  year             = {1991}
}

@unpublished{levi1997contraction,
  author           = {Levi, Isaac},
  note             = {(manuscript, fifth version)},
  title            = {Contraction and Informational Value},
  year             = {1997}
}

@book{levi2004mild,
  address          = {Cambridge},
  author           = {Levi, Isaac},
  publisher        = {Oxford University Press},
  title            = {Mild Contraction: Evaluating Loss of Information Due to Loss of Belief},
  year             = {2004}
}

@inproceedings{liberatore1996complexity,
  author           = {Liberatore, Paolo and Schaerf, Marco and others},
  booktitle        = {AAAI/IAAI, Vol. 1},
  pages            = {556--561},
  title            = {The complexity of model checking for belief revision and update},
  year             = {1996}
}

@article{liberatore2023mixed,
  author           = {Liberatore, Paolo},
  journal          = {ACM Transactions on Computational Logic},
  number           = {3},
  pages            = {1--49},
  publisher        = {ACM New York, NY},
  title            = {Mixed iterated revisions: rationale, algorithms, and complexity},
  volume           = {24},
  year             = {2023}
}

@article{locatello2020object,
  author           = {Locatello, Francesco and Weissenborn, Dirk and Unterthiner, Thomas and Mahendran, Aravindh and Heigold, Georg and Uszkoreit, Jakob and Dosovitskiy, Alexey and Kipf, Thomas},
  journal          = {Advances in Neural Information Processing Systems},
  pages            = {11525--11538},
  title            = {Object-centric learning with slot attention},
  volume           = {33},
  year             = {2020}
}

@article{makinson1985give,
  author           = {Makinson, David},
  doi              = {10.1007/BF00485895},
  journal          = {Synthese},
  number           = {3},
  pages            = {347--363},
  title            = {How to Give It Up: A Survey of Some Formal Aspects of the Logic of Theory Change},
  volume           = {62},
  year             = {1985}
}

@article{makinson1997screened,
  author           = {Makinson, David},
  journal          = {Theoria},
  pages            = {14--23},
  title            = {Screened revision},
  volume           = {63},
  year             = {1997}
}

@article{makinson2002ways,
  author           = {Makinson, David},
  journal          = {Journal of Logic and Computation},
  pages            = {3--13},
  title            = {Ways of doing logic: what was different about AGM 1985?},
  volume           = {13},
  year             = {2002}
}

@article{makinson2009propositional,
  author           = {Makinson, David},
  journal          = {Journal of Applied Logic},
  number           = {4},
  pages            = {377-387},
  title            = {Propositional relevance through letter-sharing},
  volume           = {7},
  year             = {2009}
}

@article{makinson2011conditional,
  author           = {Makinson, David},
  journal          = {Journal of Philosophical Logic},
  number           = {2},
  pages            = {121--153},
  publisher        = {Springer},
  title            = {Conditional probability in the light of qualitative belief change},
  volume           = {40},
  year             = {2011}
}

@article{manhaeve2018deepproblog,
  author           = {Manhaeve, Robin and Dumancic, Sebastijan and Kimmig, Angelika and Demeester, Thomas and De Raedt, Luc},
  journal          = {Advances in Neural Information Processing Systems},
  title            = {DeepProbLog: Neural probabilistic logic programming},
  volume           = {31},
  year             = {2018}
}

@article{marcus2020next,
  author           = {Marcus, Gary},
  journal          = {arXiv preprint arXiv:2002.06177},
  title            = {The next decade in AI: four steps towards robust artificial intelligence},
  year             = {2020}
}

@phdthesis{marsh1994formalising,
  author           = {Marsh, Stephen Paul},
  school           = {University of Stirling},
  title            = {Formalising trust as a computational concept},
  year             = {1994}
}

@article{martins1988model,
  author           = {Martins, Jo{\~a}o and Shapiro, Stuart},
  journal          = {Artificial Intelligence},
  pages            = {25--79},
  title            = {A model for Belief Revision},
  volume           = {35},
  year             = {1988}
}

@inproceedings{matos2022repairing,
  title={Repairing ontologies via kernel pseudo-contraction},
  author={Matos, Vin{\'\i}cius Bitencourt and Wassermann, Renata},
  booktitle={CEUR Workshop Proceedings},
  volume={3197},
  pages={16--26},
  year={2022}
}

@article{mcallester1982reasoning,
  author           = {McAllester, David A},
  journal          = {Version One, Artificial Intelligence Laboratory, AIM-667, MIT, Cambridge, MA},
  title            = {Reasoning utility package user's manual},
  year             = {1982}
}

@article{mccarthy1980circumscription,
  author           = {McCarthy, John},
  doi              = {10.1016/0004-3702(80)90011-9},
  issn             = {0004-3702},
  journal          = {Artificial Intelligence},
  number           = {1},
  pages            = {27--39},
  title            = {Circumscription—A Form of Non-Monotonic Reasoning},
  volume           = {13},
  year             = {1980}
}

@incollection{mccarthy1981some,
  title            = {Some philosophical problems from the standpoint of artificial intelligence},
  author           = {McCarthy, John and Hayes, Patrick J},
  booktitle        = {Readings in artificial intelligence},
  pages            = {431--450},
  year             = {1981},
  publisher        = {Elsevier}
}

@techreport{mcdermott1972conniver,
  title = {The {CONNIVER} Reference Manual},
  author = {McDermott, D. and Sussman, G. J.},
  year = {1972},
  institution = {Massachusetts Institute of Technology, AI Lab},
  number = {Memo No. 259},
  address = {Cambridge, MA}
}

@article{meyer2002systematic,
  author           = {Meyer, Thomas and Heidema, Johannes and Labuschagne, Willem and Leenen, Louise},
  journal          = {Journal of Philosophical Logic},
  pages            = {415-443},
  title            = {Systematic Withdrawal},
  volume           = {31:5},
  year             = {2002}
}

@article{miller2019explanation,
  author           = {Miller, Tim},
  journal          = {Artificial Intelligence},
  pages            = {1--38},
  publisher        = {Elsevier},
  title            = {Explanation in artificial intelligence: Insights from the social sciences},
  volume           = {267},
  year             = {2019}
}

@book{mitchell1997machine,
  address          = {New York},
  author           = {Mitchell, Tom M},
  publisher        = {McGraw-Hill},
  title            = {Machine Learning},
  year             = {1997}
}

@inproceedings{nebel1989knowledge,
  author           = {Nebel, Bernhard},
  booktitle        = {Proceedings of the 1st
International Conference of Principles of Knowledge Representation and Rea-
soning},
  pages            = {301--311},
  publisher        = {Morgan Kaufmann},
  title            = {A knowledge level analysis of belief revision},
  year             = {1989}
}

@article{nebel1994base,
  author           = {Nebel, Bernhard},
  journal          = {ECAI},
  pages            = {341--345},
  title            = {Base revision operations and schemes: Semantics, representation and complexity},
  volume           = {94},
  year             = {1994}
}

@incollection{nebel1998hard,
  author           = {Nebel, Bernhard},
  booktitle        = {Handbook of Defeasible Reasoning and Uncertainty Management Systems},
  chapter          = {3},
  doi              = {10.1007/978-94-011-5054-5_3},
  editor           = {Dubois, Didier and Prade, Henri},
  journal          = {Handbook of Defeasible Reasoning and Uncertainty Management Systems},
  pages            = {77--145},
  publisher        = {Kluwer Academic Publishers},
  title            = {How hard is it to revise a belief base?},
  volume           = {3},
  year             = {1998}
}

@article{newell1982knowledge,
  author           = {Newell, Allen},
  journal          = {Artificial intelligence},
  number           = {1},
  pages            = {87--127},
  publisher        = {Elsevier},
  title            = {The knowledge level},
  volume           = {18},
  year             = {1982}
}

@inproceedings{niederee1991multiple,
  address          = {Berlin},
  author           = {Nieder\'ee, Reinhard},
  booktitle        = {The Logic of Theory Change},
  editor           = {Fuhrmann and Morreau},
  pages            = {322--334},
  publisher        = {Springer-Verlag},
  title            = {Multiple contraction: {A} further case against {G}\"ardenfors'
principle of recovery},
  year             = {1991}
}

@phdthesis{olsson1997coherence,
  author           = {Olsson, Erik},
  journal          = {Studies in Epistemology and Belief Revision},
  school           = {Department of Philosophy. Uppsala University},
  title            = {Coherence},
  year             = {1997}
}

@article{ouyang2022training,
  author           = {Ouyang, Long and Wu, Jeffrey and Jiang, Xu and Almeida, Diogo and Wainwright, Carroll and Mishkin, Pamela and Zhang, Chong and Agarwal, Sandhini and Slama, Katarina and Ray, Alex and others},
  journal          = {Advances in Neural Information Processing Systems},
  pages            = {27730--27744},
  title            = {Training language models to follow instructions with human feedback},
  volume           = {35},
  year             = {2022}
}

@inproceedings{paglieri2004argumentation,
  author           = {Paglieri, Fabio and Castelfranchi, Cristiano},
  booktitle        = {CMNA IV: 4th workshop on Computational Models of Natural Argument},
  pages            = {5-12},
  publisher        = {Valencia: ECAI 2004},
  title            = {Argumentation and Data-oriented Belief Revision: On the Two-sided Nature of Epistemic Change},
  year             = {2004}
}

@phdthesis{paglieri2006belief,
  author           = {Paglieri, Fabio},
  school           = {Universita degli Studi di Siena},
  title            = {Belief dynamics: From formal models to cognitive architectures, and back again},
  year             = {2006}
}

@incollection{paglieri2006toulmin,
  author           = {Paglieri, Fabio and Castelfranchi, Cristiano},
  booktitle        = {Arguing on the Toulmin model},
  editor           = {D. Hitchcock, B. Verheij},
  pages            = {359-377},
  publisher        = {Berlin, Springer},
  title            = {The Toulmin Test: Framing argumentation within belief revision theories},
  year             = {2006}
}

@incollection{parikh1999beliefs,
  author           = {Parikh, Rohit},
  booktitle        = {Logic, Language, and Computation},
  journal          = {Logic, Language and Computation},
  pages            = {266-268},
  publisher        = {Springer-Verlag},
  series           = {CSLI Lecture Notes },
  title            = { Beliefs, belief revision, and splitting languages},
  volume           = {96-2},
  year             = {1999}
}

@article{pearl1986fusion,
  author           = {Pearl, Judea},
  journal          = {Artificial Intelligence},
  number           = {3},
  pages            = {241--288},
  publisher        = {Elsevier},
  title            = {Fusion, propagation, and structuring in belief networks},
  volume           = {29},
  year             = {1986}
}

@inproceedings{pearl1990causal,
  address          = {NLD},
  author           = {Verma, Thomas and Pearl, Judea},
  booktitle        = {Proceedings of the Fourth Annual Conference on Uncertainty in Artificial Intelligence},
  isbn             = {0444886508},
  numpages         = {10},
  pages            = {69–78},
  publisher        = {North-Holland Publishing Co.},
  series           = {UAI '88},
  title            = {Causal networks: semantics and expressiveness},
  year             = {1990}
}

@article{pearl1990system,
  author           = {Pearl, Judea},
  journal          = {Proceedings of the 3rd Conference on Theoretical Aspects of Reasoning about Knowledge},
  pages            = {121--135},
  publisher        = {Morgan Kaufmann},
  title            = {System Z: a natural ordering of defaults with tractable applications to nonmonotonic reasoning},
  year             = {1990}
}

@book{pearl2009causality,
  address          = {Cambridge},
  author           = {Pearl, Judea},
  doi              = {10.1017/CBO9780511803161},
  edition          = {2nd},
  isbn             = {978-0-521-89560-6},
  publisher        = {Cambridge University Press},
  title            = {Causality: Models, Reasoning and Inference},
  year             = {2009}
}

@book{pearl2018book,
  author           = {Pearl, Judea and Mackenzie, Dana},
  publisher        = {Basic Books},
  title            = {The book of why: the new science of cause and effect},
  year             = {2018}
}

@article{peirce1877fixation,
  author           = {Peirce, Charles Sanders},
  journal          = {Popular Science Monthly},
  number           = {1},
  pages            = {1--15},
  title            = {The Fixation of Belief},
  volume           = {12},
  year             = {1877}
}

@article{pollock1987defeasible,
  author           = {Pollock, John L},
  doi              = {10.1207/s15516709cog1104_4},
  journal          = {Cognitive Science},
  number           = {4},
  pages            = {481--518},
  publisher        = {Wiley},
  title            = {Defeasible reasoning},
  volume           = {11},
  year             = {1987}
}

@article{poveda2010oops,
  author           = {Poveda-Villal{\'o}n, Mar{\'\i}a and Su{\'a}rez-Figueroa, Mari Carmen and G{\'o}mez-P{\'e}rez, Asunci{\'o}n},
  journal          = {International Journal on Semantic Web and Information Systems (IJSWIS)},  
  number           = {2},
  pages            = {7--34},
  title            = {OOPS! (OntOlogy Pitfall Scanner!): An on-line tool for ontology evaluation},
  volume           = {10},
  year             = {2014}
}

@inproceedings{rao1995agents,
  address          = {Menlo Park, CA},
  author           = {Rao, Anand S and Georgeff, Michael P and others},
  booktitle        = {ICMAS},
  isbn             = {978-0-262-51087-5},
  pages            = {312--319},
  publisher        = {AAAI Press},
  title            = {BDI agents: from theory to practice.},
  volume           = {95},
  year             = {1995}
}

@article{reiter1980logic,
  author           = {Reiter, Raymond},
  journal          = {Artificial Intelligence},
  number           = {1--2},
  pages            = {81--132},
  title            = {A Logic for Default Reasoning},
  volume           = {13},
  year             = {1980}
}

@article{rott1992preferential,
  author           = {Rott, Hans},
  journal          = {Journal of Logic, Language and Information},
  number           = {1},
  pages            = {45--78},
  publisher        = {Springer},
  title            = {Preferential belief change using generalized epistemic entrenchment},
  volume           = {1},
  year             = {1992}
}

@techreport{rott1998just,
  author           = {Rott, Hans},
  title            = {{``Just Because'': Taking Belief Bases Seriously}},
  institution      = {Department of Philosophy, Lund University},
  type             = {Technical Report},
  number           = {LP-1998-13},
  address          = {Lund, Sweden},
  year             = {1998}
}

@incollection{rott1991two,
  title={Two methods of constructing contractions and revisions of knowledge systems},
  author={Rott, Hans},
  booktitle={Journal of Philosophical Logic},
  volume={20},
  number={2},
  pages={149--173},
  year={1991},
  publisher={Springer}
}

@article{rott1999severe,
  author           = {Rott, Hans and Pagnucco, Maurice},
  journal          = {Journal of Philosophical Logic},
  pages            = {501--547},
  title            = {Severe Withdrawal (and Recovery)},
  volume           = {28},
  year             = {1999}
}

@book{rott2001change,
  address          = {Oxford},
  author           = {Rott, Hans},
  isbn             = {978-0-19-825027-8},
  publisher        = {Clarendon Press},
  series           = {Oxford Logic Guides},
  title            = {Change, Choice and Inference: A Study of Belief Revision and Nonmonotonic Reasoning},
  year             = {2001}
}

@incollection{rott2001revealed,
  author           = {Rott, Hans},
  isbn             = {9780198503064},
  title            = {Revealed Preferences: Understanding the Theory of Epistemic Entrenchment},
  booktitle        = {Change, Choice and Inference: A Study of Belief Revision and Nonmonotonic Reasoning},
  publisher        = {Oxford University Press},
  year             = {2001},
  pages            = {223--267},
  month            = {10},    
  doi              = {10.1093/oso/9780198503064.003.0009},
  url              = {https://doi.org/10.1093/oso/9780198503064.003.0009},
  eprint           = {https://academic.oup.com/book/0/chapter/422192581/chapter-pdf/52439740/isbn-9780198503064-book-part-9.pdf},
}

@article{rott2003coherence,
  author           = {Rott, Hans},
  journal          = {Journal of Logic and Computation},
  pages            = {111--145},
  title            = {Coherence and conservatism in the dynamics of belief.
Part {II}: Iterated belief change without dispositional coherence},
  volume           = {13},
  year             = {2003}
}

@incollection{rott2009shifting,
  author           = {Rott, Hans},
  booktitle        = {Towards Mathematical Philosophy},
  editor           = {D. Makinson, J. Malinowski, H. Wansing},
  number           = {28},
  pages            = {269-296},
  publisher        = {Springer Science},
  series           = {Trends in Logic},
  title            = {Shifting Priorities: Simple Representations for Twenty-seven Iterated Theory Change Operators},
  year             = {2009}
}

@book{russell2010artificial,
  author           = {Russell, Stuart J. and Norvig, Peter},
  title            = {Artificial Intelligence: A Modern Approach},
  edition          = {3rd},
  year             = {2010},
  publisher        = {Prentice Hall},
  address          = {Upper Saddle River, NJ},
  isbn             = {978-0-13-604259-4}
}

@book{russell2019human,
  author           = {Russell, Stuart J.},
  publisher        = {Viking},
  title            = {Human compatible: Artificial intelligence and the problem of control},
  isbn             = {9780525558613},
  lccn             ={2019029688},
  url              = {https://books.google.pt/books?id=8vm0DwAAQBAJ},
  year             = {2019}
}

@article{sabater2005review,
  author           = {Sabater, Jordi and Sierra, Carles},
  journal          = {Artificial Intelligence Review},
  number           = {1},
  pages            = {33--60},
  publisher        = {Springer},
  title            = {Review on computational trust and reputation models},
  volume           = {24},
  year             = {2005}
}

@inproceedings{satoh1988nonmonotonic,
  author           = {Satoh, Ken},
  booktitle        = {Proceedings of the International Conference on Fifth Generation Computer Systems (FGCS'88)},
  pages            = {455-462},
  title            = {Nonmonotonic Reasoning by Minimal Belief Revision},
  year             = {1988}
}

@article{schaerf1995tractable,
  author           = {Schaerf, Marco and Cadoli, Marco},
  journal          = {Artificial Intelligence},
  number           = {2},
  pages            = {249--310},
  publisher        = {Elsevier},
  title            = {Tractable reasoning via approximation},
  volume           = {74},
  year             = {1995}
}

@techreport{segerberg1998irrevocable,
  author           = {Segerberg, Krister},
  institution      = {Dep. of Philosophy, Uppsala University},
  number           = {6},
  title            = {Irrevocable Belief Revision in Dynamic Doxastic Logic},
  type             = {Uppsala Prints and Preprints in Philosophy},
  year             = {1998}
}

@article{shearer2008hermit,
  author           = {Shearer, Rob and Motik, Boris and Horrocks, Ian},
  journal          = {OWLED},
  pages            = {91},
  title            = {HermiT: A highly-efficient OWL reasoner},
  volume           = {432},
  year             = {2008}
}

@article{sirin2007pellet,
  author           = {Sirin, Evren and Parsia, Bijan and Grau, Bernardo Cuenca and Kalyanpur, Aditya and Katz, Yarden},
  journal          = {Journal of Web Semantics},
  number           = {2},
  pages            = {51--53},
  publisher        = {Elsevier},
  title            = {Pellet: A practical OWL-DL reasoner},
  volume           = {5},
  year             = {2007}
}

@phdthesis{souza2024bridging,
  author           = {Souza, Davy Alves de},
  school           = {Universidade de S{\~a}o Paulo},
  title            = {Bridging belief revision and ontology repair: moving closer to optimal repairs},
  year             = {2024}
}

@article{spirtes1993causation,
  address          = {New York},
  author           = {Spirtes, Peter and Glymour, Clark and Scheines, Richard},
  doi              = {10.1007/978-1-4612-2748-9},
  isbn             = {978-0-387-97979-3},
  journal          = {Lecture Notes in Statistics},
  publisher        = {Springer-Verlag},
  title            = {Causation, prediction, and search},
  volume           = {81},
  year             = {1993}
}

@incollection{spohn1988ordinal,
  address          = {Dordrecht},
  author           = {Spohn, Wolfgang},
  booktitle        = {Causation in Decision, Belief Change and Statistics},
  editor           = {Harper, W. and Skyrms, B.},
  pages            = {105--134},
  publisher        = {D. Reidel},
  title            = {Ordinal Conditional Functions: A Dynamic Theory of
Epistemic States},
  volume           = {volume 2},
  year             = {1988}
}

@article{stallman1977forward,
  author           = {Stallman, Richard M and Sussman, Gerald J},
  journal          = {Artificial intelligence},
  number           = {2},
  pages            = {135--196},
  publisher        = {Elsevier},
  title            = {Forward reasoning and dependency-directed backtracking in a system for computer-aided circuit analysis},
  volume           = {9},
  year             = {1977}
}

@article{stone2000multiagent,
  author           = {Stone, Peter and Veloso, Manuela},
  journal          = {Autonomous Robots},
  number           = {3},
  pages            = {345--383},
  publisher        = {Springer},
  title            = {Multiagent systems: A survey from a machine learning perspective},
  volume           = {8},
  year             = {2000}
}

@InCollection{strasser2024,
	author       =	{Strasser, Christian and Antonelli, G. Aldo},
	title        =	{{Non-monotonic Logic}},
	booktitle    =	{The {Stanford} Encyclopedia of Philosophy},
	editor       =	{Edward N. Zalta and Uri Nodelman},
	howpublished =	{\url{https://plato.stanford.edu/archives/win2024/entries/logic-nonmonotonic/}},
	year         =	{2024},
	edition      =	{{W}inter 2024},
	publisher    =	{Metaphysics Research Lab, Stanford University}
}

@article{tampitsikas2019theoretical,
  author           = {Robnik-{\v{S}}ikonja, Marko and Kononenko, Igor},
  journal          = {Machine Learning},
  number           = {1-2},
  pages            = {23--69},
  publisher        = {Springer},
  title            = {Theoretical and empirical analysis of ReliefF and RReliefF},
  volume           = {53},
  year             = {2003}
}

@article{thimm2014tweety,
  title={Tweety: A comprehensive collection of Java libraries for logical aspects of artificial intelligence and knowledge representation},
  author={Thimm, Matthias},
  journal={Proceedings of the 14th International Conference on Principles of Knowledge Representation and Reasoning},
  pages={528--537},
  year={2014}
}

@book{toulmin2003uses,
  author           = {Toulmin, Stephen E},
  publisher        = {Cambridge university press},
  title            = {The uses of argument},
  year             = {2003}
}

@incollection{vanharmelen2008,
  author           = {Peppas, Pavlos},
  booktitle        = {Handbook of Knowledge Representation},
  doi              = {https://doi.org/10.1016/S1574-6526(07)03008-8},
  editor           = {Frank {van Harmelen} and Vladimir Lifschitz and Bruce Porter},
  issn             = {1574-6526},
  pages            = {317-359},
  publisher        = {Elsevier},
  series           = {Foundations of Artificial Intelligence},
  title            = {Chapter 8 Belief Revision},
  url              = {https://www.sciencedirect.com/science/article/pii/S1574652607030088},
  volume           = {3},
  year             = {2008}
}

@article{wang2019human,
  author           = {Wang, Dakuo and Weisz, Justin D and Muller, Michael and Ram, Parikshit and Geyer, Werner and Dugan, Casey and Tausczik, Yla and Samulowitz, Horst and Gray, Alexander},
  journal          = {Proceedings of the 2019 CHI Conference on Human Factors in Computing Systems},
  pages            = {1--12},
  title            = {Human-AI collaboration in data science: Exploring data scientists' perceptions of automated feature engineering},
  year             = {2019}
}

@article{wassermann1999resource,
  author           = {Wassermann, Renata},
  issn             = {01650106, 15728420},
  journal          = {Erkenntnis (1975-)},
  number           = {2/3},
  pages            = {429--446},
  publisher        = {Springer},
  title            = {Resource Bounded Belief Revision},
  url              = {http://www.jstor.org/stable/20012927},
  urldate          = {2025-10-29},
  volume           = {50},
  year             = {1999}
}

@article{williams1986doing,
  author           = {Williams, Brian C.},
  journal          = {Proceedings of the Fifth National Conference on Artificial Intelligence (AAAI-86)},
  pages            = {105--112},
  publisher        = {AAAI Press},
  title            = {Doing time: Putting qualitative reasoning on firmer ground},
  year             = {1986}
}

@inproceedings{williams1994transmutations,
  author           = {Williams, Mary-Anne},
  booktitle        = {Principles of Knowledge Representation and Reasoning},
  organization     = {Elsevier},
  pages            = {619--629},
  title            = {Transmutations of knowledge systems},
  year             = {1994}
}

@article{williams1996towards,
  author           = {Williams, Mary-Anne},
  journal          = {KR},
  pages            = {412--420},
  title            = {Towards a practical approach to belief revision: Reason-based change},
  volume           = {96},
  year             = {1996}
}

@inproceedings{williams1997anytime,
	title        = {Anytime belief revision},
	author       = {Williams, Mary-Anne},
	year         = 1997,
	booktitle    = {Proceedings of the 15th International Joint Conference on Artifical Intelligence - Volume 1},
	location     = {Nagoya, Japan},
	publisher    = {Morgan Kaufmann Publishers Inc.},
	address      = {San Francisco, CA, USA},
	series       = {IJCAI'97},
	pages        = {74–79},
	isbn         = 15558604804,
	numpages     = 6
}

@inproceedings{williams1997applications,
	title        = {Applications of belief revision},
	author       = {Williams, Mary-Anne},
	year         = 1998,
	booktitle    = {Transactions and Change in Logic Databases},
	publisher    = {Springer Berlin Heidelberg},
	address      = {Berlin, Heidelberg},
	pages        = {287--316},
	isbn         = {978-3-540-49449-2},
	editor       = {Freitag, Burkhard and Decker, Hendrik and Kifer, Michael and Voronkov, Andrei}
}

@inproceedings{winslett1988reasoning,
  author           = {Winslett, Marianne},
  booktitle        = {AAAI},
  pages            = {89-93},
  title            = {Reasoning about Action Using a Possible Models Approach},
  year             = {1988}
}

@book{winslett1991updating,
  address          = {USA},
  author           = {Winslett, Marianne},
  isbn             = {0521373719},
  publisher        = {Cambridge University Press},
  title            = {Updating logical databases},
  year             = {1991}
}

@article{zaharia2016apache,
  author           = {Zaharia, Matei and Xin, Reynold S and Wendell, Patrick and Das, Tathagata and Armbrust, Michael and Dave, Ankur and Meng, Xiangrui and Rosen, Josh and Venkataraman, Shivaram and Franklin, Michael J and others},
  journal          = {Communications of the ACM},
  number           = {11},
  pages            = {56--65},
  publisher        = {ACM},
  title            = {Apache Spark: a unified engine for big data processing},
  volume           = {59},
  year             = {2016}
}

@article{zhang2001infinitary,
  author           = {Zhang, Dongmo and Foo, Norman},
  journal          = {Journal of Philosophical Logic},
  number           = {6},
  pages            = {525--570},
  publisher        = {Springer},
  title            = {Infinitary belief revision},
  volume           = {30},
  year             = {2001}
}

@article{zhang2010logic,
  author           = {Zhang, Dongmo},
  journal          = {Artificial Intelligence},
  pages            = {1307-1322},
  title            = {A Logic-Based Axiomatic Model of Bargaining},
  volume           = {174},
  year             = {2010}
}

@incollection{zimmerman2022pragmatism,
  author           = {Zimmerman, Aaron},
  doi              = {10.4324/9781315149592-32},
  pages            = {226 - 238},
  booktitle        = {The Routledge Companion to Pragmatism},
  publisher        = {Routledge eBooks},
  title            = {Pragmatism in the Philosophy of Mind},
  year             = {2022}
}

\appendix
\section{Theoretical Development Tracking Methodology} \label{app1}

Our theoretical development tracking methodology employed a systematic snowballing approach anchored in Doyle and London's 1980 bibliography. Starting with the pre-AGM papers catalogued in their taxonomy, we identified forward citations that led to the emergence of the AGM framework, particularly tracing how early works on consistency maintenance, belief dependencies, and rational choice influenced Alchourrón, Gärdenfors, and Makinson's foundational 1985 paper. From this AGM cornerstone, we applied forward snowballing to capture major theoretical extensions, using citation analysis to identify papers with high impact (>100 citations) that explicitly built upon AGM foundations. We prioritized works that: (1) extended core AGM constructs (epistemic entrenchment, kernel methods); (2) addressed AGM limitations (iterated revision, belief bases); (3) provided representation theorems or complexity analyses; and (4) demonstrated clear lineage from pre-AGM computational insights. This approach ensured systematic coverage of the theoretical evolution while maintaining traceability from Doyle and London's empirical starting point to contemporary AGM-based research.

\subsection{Detailed Methodology Explanation}

\subsubsection{Phase 1: Pre-AGM Foundation Analysis}

The theoretical development tracking began with a comprehensive analysis of Doyle and London's 1980 bibliography, which contained approximately 250 papers representing the pre-AGM computational landscape. We systematically categorized these papers according to their original eight-point taxonomy, identifying those that addressed:

\begin{itemize}
\item \textbf{Consistency maintenance mechanisms} (Truth Maintenance Systems, dependency networks)
\item \textbf{Belief revision procedures} (backtracking, assumption management, conflict resolution)
\item \textbf{Rational choice and preference modeling} (decision theory applications, utility-based approaches)
\item \textbf{Knowledge representation frameworks} (semantic networks, frame systems, production rules)
\end{itemize}

From this foundation, we employed \textbf{backward citation analysis} to identify the theoretical and philosophical roots that influenced these computational approaches, tracing connections to works in epistemology, philosophy of science, and formal logic that would later inform AGM theory.

\subsubsection{Phase 2: AGM Emergence Tracking}

Using \textbf{forward citation analysis} from the pre-AGM papers, we traced the emergence of the AGM framework by identifying papers that:

\begin{enumerate}
\item \textbf{Cited multiple pre-AGM computational works} while developing formal theoretical frameworks
\item \textbf{Addressed theoretical unification} of disparate computational approaches under common principles
\item \textbf{Introduced normative postulates} for rational belief change operations
\item \textbf{Provided mathematical formalizations} of concepts that appeared algorithmically in pre-AGM work
\end{enumerate}

This process revealed how Alchourrón, Gärdenfors, and Makinson's 1985 foundational paper synthesized insights from:
\begin{itemize}
\item Doyle's Truth Maintenance Systems (dependency tracking and conflict resolution)
\item Decision theory and rational choice literature (preference structures and consistency requirements)
\item Formal logic and model theory (semantic approaches to belief change)
\item Philosophy of science literature (theory change and scientific rationality)
\end{itemize}

\subsubsection{Phase 3: Post-AGM Extension Mapping}

From the AGM foundation, we applied \textbf{systematic forward snowballing} using the following selection criteria:

\textbf{Citation Impact Threshold:} Papers with $>$ 100 citations that explicitly reference AGM foundational work, indicating significant influence on the field's development.

\textbf{Theoretical Extension Categories:}
\begin{itemize}
\item \textbf{Constructive extensions} that operationalized AGM postulates through specific construction methods
\item \textbf{Limitation addresses} that identified and resolved problems with the original AGM framework
\item \textbf{Representation theorems} that connected abstract postulates to concrete computational procedures
\item \textbf{Complexity analyses} that characterized the computational properties of AGM operations
\end{itemize}

\textbf{Lineage Verification:} Each selected paper was verified to maintain a clear intellectual lineage either from:
\begin{itemize}
\item Pre-AGM computational insights (showing how algorithmic innovations were formalized)
\item AGM theoretical constructs (showing systematic extension or refinement)
\item Cross-connections between computational and theoretical development
\end{itemize}

\subsubsection{Phase 4: Systematic Coverage Validation}

To ensure comprehensive coverage, we employed \textbf{cross-reference validation}:

\begin{enumerate}
\item \textbf{Survey paper analysis:} We examined major belief revision surveys (Gärdenfors 1988, Hansson 1999, Fermé \& Hansson 2018) to identify any significant theoretical developments our snowballing approach might have missed.

\item \textbf{Venue-based sampling:} We systematically reviewed proceedings from major AI conferences (IJCAI, AAAI, KR) and journals (AIJ, JLC, JPL) from 1985--2024 to capture developments that might not have strong citation links to our core papers.

\item \textbf{Paradigm boundary checking:} We verified that our selection captured the major theoretical paradigms within AGM-adjacent belief revision, including:
\begin{itemize}
\item Classical AGM theory and its direct extensions
\item Belief base approaches and their relationship to belief set methods
\item Iterated revision and dynamic belief change
\item Computational complexity and approximation theory
\item Multi-agent and social choice extensions
\end{itemize}
\end{enumerate}

\subsubsection{Phase 5: Temporal and Conceptual Organization}

The selected papers were organized both chronologically and conceptually to trace:

\textbf{Temporal Development:} How theoretical insights emerged, matured, and influenced subsequent work over the 1985--2024 period.

\textbf{Conceptual Lineage:} How specific pre-AGM computational innovations (e.g., dependency tracking in TMS) evolved into formal theoretical constructs (e.g., epistemic entrenchment orderings).

\textbf{Cross-Paradigm Connections:} How developments in different areas (complexity theory, preference learning, multi-agent systems) intersected with and influenced core AGM theory.

\subsubsection{Quality Assurance and Validation}

Our selection process included several quality assurance measures:

\textbf{Peer Review Integration:} We prioritized papers published in top-tier venues with rigorous peer review processes, ensuring theoretical quality and methodological soundness.

\textbf{Replication and Extension Analysis:} We traced how key theoretical results were replicated, extended, or challenged by subsequent work, ensuring our selection captured both stable contributions and evolving understanding.

\textbf{Implementation Connection:} We maintained explicit attention to papers that bridged theory and implementation, ensuring our theoretical tracking remained connected to computational practice.

\textbf{Expert Validation:} Our selection was cross-checked against reading lists from major belief revision courses and recommendations from established researchers in the field.

This systematic approach ensured that our theoretical development tracking maintained both historical accuracy and comprehensive coverage while preserving the connection between pre-AGM computational insights and post-AGM theoretical developments that is central to our analysis.
\end{document}